\documentclass[10pt,twocolumn,letterpaper,table]{article}

\usepackage[pagenumbers]{wacv} %

\usepackage{graphicx}
\usepackage{amsmath}
\usepackage{amssymb}
\usepackage{booktabs}

\usepackage[]{tabularx}
\usepackage{multirow}
\usepackage[10pt]{moresize}
\usepackage{pgfplots}
\pgfplotsset{compat=1.18}

\usepackage[accsupp]{axessibility} %

\usepackage{soul}
\usepackage{xcolor}

\usepackage{hhline}

\usepackage{chngcntr}

\usepackage{tikz}
\usetikzlibrary{spy,arrows}

\usepackage[framemethod=tikz]{mdframed}

\usepackage{etoolbox}   %
\providetoggle{arxiv}
\settoggle{arxiv}{false}

\def\psnr{PSNR\xspace}
\def\lpips{LPIPS\xspace}
\def\ssim{SSIM\xspace}
\def\dssim{DSSIM\xspace}

\def\psnr{PSNR\xspace}
\def\lpips{LPIPS\xspace}
\def\ssim{SSIM\xspace}
\def\flip{\FLIP\xspace}
\def\psnrArrow{PSNR~$\uparrow$\xspace}

\def\dssimArrow{\dssim~$\downarrow$\xspace}
\def\lpipsArrow{\lpips~$\downarrow$\xspace}
\def\flipArrow{\flip~$\downarrow$\xspace}
\def\gt{GT\xspace}
\def\tslong{Torrance-Sparrow\xspace}
\def\ts{TS\xspace}
\def\rplong{Realistic Phong\xspace}
\def\rp{RP\xspace}
\def\disney{Disney\xspace}
\def\fmbrdflong{Fresnel Microfacet BRDF\xspace}
\def\fmbrdf{FMBRDF\xspace}
\def\merl{MERL\xspace}
\def\diligent{DiLiGenT-MV\xspace}
\def\lpipsc{\lpips~\cite{zhang2018unreasonable} \xspace}
\def\ssimc{\ssim~\cite{wang2004image} \xspace}

\def\tslongc{\tslong~\cite{torrance1967theory}\xspace}
\def\tsc{\ts~\cite{torrance1967theory}\xspace}
\def\rplongc{\rplong~\cite{lafortune1994using}\xspace}
\def\rpc{\rp~\cite{lafortune1994using}\xspace}
\def\disneyc{\disney~\cite{burley2012physically}\xspace}
\def\fmbrdflongc{\fmbrdflong~\cite{ichikawa2023fresnel}\xspace}
\def\fmbrdfc{\fmbrdf~\cite{ichikawa2023fresnel}\xspace}
\def\merlc{\merl~\cite{matusik2003MERL}\xspace}
\def\diligentc{\diligent~\cite{Li2020DiLiGentMVDataset}\xspace}

\definecolor{greenValid}{HTML}{33cc33}%

\definecolor{best}{HTML}{66bb6a}%
\definecolor{secondBest}{HTML}{a5d6a7}%
\definecolor{thirdBest}{HTML}{e8f5e9}%
\definecolor{valid}{HTML}{66bb6a}%

\definecolor{cellParamBased}{HTML}{FFF9C4}    %
\definecolor{celPurelyNeural}{HTML}{E1F5FE}

\definecolor{cellParamBasedDarker}{HTML}{FFF9C4}

\definecolor{mycolor1}{rgb}{0.00000,0.44700,0.74100}%
\definecolor{mycolor2}{rgb}{0.85000,0.32500,0.09800}%
\definecolor{mycolor3}{rgb}{0.92900,0.69400,0.12500}%
\definecolor{mycolor4}{rgb}{0.49400,0.18400,0.55600}%
\definecolor{mycolor5}{rgb}{0.46600,0.67400,0.18800}%
\definecolor{mycolor6}{rgb}{0.30100,0.74500,0.93300}%
\definecolor{mycolor7}{rgb}{0.63500,0.07800,0.18400}%

\DeclareRobustCommand{\hlBest}[1]{{\sethlcolor{best}\hl{#1}}}
\DeclareRobustCommand{\hlSecondBest}[1]{{\sethlcolor{secondBest}\hl{#1}}}
\DeclareRobustCommand{\hlThirdBest}[1]{{\sethlcolor{thirdBest}\hl{#1}}}

\newcommand{\mysquare}[1][black]{\small\textcolor{#1}{\ensuremath\blacksquare}}

\newcommand{\brdf}{f}
\newcommand{\point}{x}
\newcommand{\light}{l}
\newcommand{\view}{v}
\newcommand*\Sphere{\mathbb{S}}
\newcommand*\Hemi{\mathbb{H}}
\newcommand{\LightIntIn}{L_i}
\newcommand{\LightIntOut}{L_o}

\newcommand{\FLIP}{\protect\reflectbox{F}LIP\xspace}

\usepackage[breaklinks,colorlinks]{hyperref}

\usepackage[capitalize]{cleveref}
\crefname{section}{Sec.}{Secs.}
\Crefname{section}{Section}{Sections}
\Crefname{table}{Table}{Tables}
\crefname{table}{Tab.}{Tabs.}

\begin{document}
\settoggle{arxiv}{true}

\input{includes/paperInfo}

\title{\paperTitle}
\maketitle
\thispagestyle{empty}

\begin{abstract}

The bidirectional reflectance distribution function (BRDF) is an essential tool to capture the complex interaction of light and matter. 
Recently, several works have employed neural methods for BRDF modeling, following various strategies, ranging from utilizing existing parametric models to purely neural parametrizations. 
While all methods yield impressive results, a comprehensive comparison of the different approaches is missing in the literature.
In this work, we present a thorough evaluation of several approaches, including results for qualitative and quantitative reconstruction quality and an analysis of reciprocity and energy conservation. 
Moreover, we propose two extensions that can be added to existing approaches: A novel additive combination strategy for neural BRDFs that split the reflectance into a diffuse and a specular part, and an input mapping that ensures reciprocity exactly by construction, while previous approaches only ensure it by soft constraints.
   
\end{abstract}

\section{Introduction}
The physical principles underlying the interaction of light with matter are diverse and complex. The electromagnetic waves are reflected and refracted at material interfaces, and they are scattered and absorbed within materials. These principles need to be taken into account to obtain realistic rendering results. The standard approach to describe an opaque surface is the \emph{bidirectional reflectance distribution function} (BRDF), which for a given light direction describes the amount of light reflected in a specific view direction.

Realistic BRDF modelling is a long-standing research question and countless methods have been proposed. A popular class of approaches aims to explicitly replicate the physical behavior. Some works are mainly phenomenological \cite{Blinn77,Phong75}, others, particularly the microfacet approaches, carefully model the physical principles \cite{CookTorranceT82,trowbridge1975average,walter2007microfacet,smith1967geometrical,schlick1994inexpensive}. While these works rely on a thorough analysis of the underlying physics, they only include a subset of the governing phenomena, limiting them to the modeling choices.

\input{figures/teaser}

Recently, neural fields have become 
popular for BRDF modeling, allowing for a continuous and resolution-free representation. While several works ultimately rely on parametric physical models, for which the neural field is used to predict the parameters \cite{bi2020neuralReflectanceFields,srinivasan2021nerv,Boss2021NERD,Zhang22IRON,Deschaintre2018SingleImageSVBRDFCaptureDeepNN,Henzler2021GenerativeModellingBRDFFlashIms,Guo2020MaterialGAN,Zhang21PhySG,Zhang2022ModellingIndirIlluminationInvRendering,Brahimi24SuperVol,brahimi24SparseViewsNearLight}, other approaches use neural networks to directly predict the value of the BRDF \cite{sztrajman2021neural,Fan2022NeuralLayeredBRDF,Zhang2021NeRFactor,Sarkar23LitNerf}. Although all neural BRDFs yield impressive results, a thorough comparison between the different approaches is lacking in the literature.

\paragraph{Contribution}

In this work, we 
address this shortcoming and 
perform an exhaustive comparison of different approaches for neural BRDF modeling. In contrast to 
the setting in 
most previous works, we deliberately estimate the reflectance for \emph{given} geometry and \emph{calibrated} light to avoid cross-influences from the joint estimation of shape, material and lighting conditions. This ensures that the capabilities of the approaches can be compared as unobscured as possible. %
The results suggest advantages for purely neural methods over approaches based on parametric models in particular for highly specular materials.

Moreover, we propose two extensions that can be added to existing purely neural approaches: A novel input mapping that ensures reciprocity of the BRDF by construction as well as an enhancement for split-based methods that aims at representing the physical behavior more faithfully. 

We summarize our main contributions as follows:
\begin{enumerate}
    \item We perform an extensive comparison of different neural BRDF approaches, including neural fields for parameters of different physical models and direct BRDF prediction by the neural network.
    \item We analyze the energy and reciprocity constraint for all methods. The results show that neural approaches do not seem to learn reciprocity from data. We propose a modification to ensure reciprocity by construction.
    \item We introduce a novel splitting scheme for the diffuse and specular part of additive neural BRDF models and show that it improves existing methods.
\end{enumerate}

Please see \href{https://florianhofherr.github.io/neural-brdfs}{florianhofherr.github.io/neural-brdfs} and the supplementary material for additional architecture and evaluation details as well as more experiments.

\section{Related Work}
BRDF modeling has been studied extensively, and various approaches have been proposed. Comprehensive overviews can be found in \cite{montes2012overviewBRDFModels} and \cite{Guarnera2016BRDFRepresentationAcquisition}. 

\paragraph{Phenomenological and Physically-Based BRDF Models}
Classical works explicitly model the interaction between light and matter by parametric equations, often employing a split into a diffuse component like the Lambertian reflection \cite{Lamber1760} and a specular component. The Phong and the Blinn-Phong models \cite{Phong75,Blinn77} are popular phenomenological approaches for the specular part, despite their physical incorrectness. Several modifications have been proposed, making them, \eg, energy conserving and anisotropic \cite{lafortune1994using,Ashikhmin00AnAnisotropicPhongBRDFModel}.

Most physically-based models for the specular part are microfacet models, introduced by Torrance and Sparrow \cite{torrance1967theory}. They use statistically distributed small mirror-like microfacets in combination with the Fresnel reflection law.
Several extensions to the microfacet model have been presented
\cite{CookTorranceT82,trowbridge1975average,walter2007microfacet,smith1967geometrical,schlick1994inexpensive,Holzschuch17ATwoScaleMicrofacetReflectanceModel}, among those, the popular Disney BRDF \cite{burley2012physically} and the Oren-Nayar model that generalizes diffuse reflection by Lambertian microfacets \cite{Oren94OrenNayar}.
Recently,  Ichikawa \etal addressed shortcomings in the latter model and proposed an approach based on Fresnel reflectance and transmission that also models polarimetric behavior \cite{ichikawa2023fresnel}.

\paragraph{Neural Approaches Based on Parametric Models}
Several neural reflectance methods rely on the extensive research on parametric and physical BRDF models. The idea is to compute the parameters of those models with spatial MLPs for a given input location and to compute the reflectance using the model with the predicted parameters.

The well-established microfacet BRDF \cite{torrance1967theory, walter2007microfacet,CookTorranceT82} has been successfully applied to neural scene reconstruction, where a spatial neural network jointly parametrizes the scene geometry and the parameters for the reflection model \cite{bi2020neuralReflectanceFields,srinivasan2021nerv,Boss2021NERD,Zhang22IRON}. Trained on multi-view images, this allows for novel view synthesis and other downstream tasks like re-lighting or the extraction of 3D assets. Other works employ the microfacet model for material estimation from a single flash image based on U-Nets \cite{Deschaintre2018SingleImageSVBRDFCaptureDeepNN,Henzler2021GenerativeModellingBRDFFlashIms} and in combination with Generative Adversarial Networks to create a generative reflectance model for material reconstruction \cite{Guo2020MaterialGAN}.

The Disney BRDF \cite{burley2012physically} is another parametric model that has been used successfully to design neural reflectance methods. Zhang \etal combine it with a neural SDF for joint estimation of scene geometry and appearance \cite{Zhang21PhySG}. Also, it is used in the scene model proposed by Zhang \etal which includes indirect illumination \cite{Zhang2022ModellingIndirIlluminationInvRendering}. Brahimi \etal employ it for scene reconstruction based on volumetric rendering \cite{Brahimi24SuperVol} and for uncalibrated point-light photometric stereo \cite{brahimi24SparseViewsNearLight}.

\paragraph{Purely Neural Models}
In contrast, purely neural methods do not depend on parametric models but use MLPs to predict the BRDF values directly. 
Two main approaches can be distinguished: While some works use a single MLP to compute the complete BRDF value, others rely on an additive split into a diffuse and a specular part.

Sztrajman \etal \cite{sztrajman2021neural} use a single neural network to represent a non-spatially-varying BRDF, which they train on the MERL data \cite{matusik2003MERL}. By using a variational-autoencoder, they compress their neural BRDFs into a latent space.
Hu \etal \cite{hu2020deepbrdf} employ an autoencoder to compress the MERL BRDFs directly.
Zhang \etal \cite{Zhang2021NeRFactor} use a neural BRDF based on an additive split with a specular component pre-trained on the MERL data \cite{matusik2003MERL} for neural scene reconstruction under unknown lighting conditions. They demonstrate a slight advantage of their purely neural approach over a microfacet BRDF-based neural model.
Sarka \etal \cite{Sarkar23LitNerf} employ several MLPs to compute an additive split as well as terms for indirect illumination to obtain highly realistic reconstructions of human heads. 
Among the other methods,
they are the only ones who aim to ensure the reciprocity of the BRDF.
Zheng \etal \cite{Zeng23RelightingNeRFsWithShadowAndHighlightHints} use a single MLP to model the reflectance for volume rendering based scene reconstruction. In contrast to previous methods, they do not model effects like interreflection and Lambert's cosine law explicitly but have the neural network learn them, which they facilitate by additional \emph{light transport hints} predicted by separate networks.

\paragraph{Neural Methods in Computer Graphics}

Neural approaches have also been explored in computer graphics to model highly complex materials.
Fan \etal \cite{Fan2022NeuralLayeredBRDF} propose learned operations in a latent space to model the layering of multiple materials.
Rainer \etal \cite{rainer2019neuralBTFCompression,rainer2020unifiedNeuralEncodingOfBTFs} and Rodriguez-Pardo \etal\cite{rodriguez2023neubtf} employ autoencoders to compress Bidirectional Texture Functions (BTFs) into a latent space, enabling fast test-time encodings.
Kuznetsov \etal extend an MLP decoder for BTFs by a generalization of classic mipmap pyramids and curvature awareness \cite{Kuznetsov21NeuMIP,kuznetsov2022renderingNeuralMaterialsOnCurvedSurfaces}.
Tg \etal have investigated neural approaches to model the subsurface scattering in translucent objects \cite{tg2024neuralSSS,TG23NeuralBSSRDF}.

\section{Background: BRDF and Rendering}
\label{sec:background}

The \emph{bidirectional reflectance distribution function} (BRDF) describes how a surface reflects incoming light. More precisely, the BRDF $\brdf(\point, \light, \view)$ describes the ratio of the radiance reflected in the viewing direction $\view\in\Sphere^2$ to the irradiance incident from the light direction $\light\in\Sphere^2$ at the position $\point$. 
In this work, we focus on \emph{spatially varying} BRDF (SVBRDF) where $\brdf$ depends on $\point$. For simplicity, however, we use BRDF and SVBRDF interchangeably.

The \emph{rendering equation} 
integrates the reflections of the incident irradiances $\LightIntIn\geq 0$ over the upper hemisphere $\Hemi$ centered around the surface normal to obtain the total radiance $\LightIntOut(\point, \view)$ at position $\point$ in the viewing direction $\view$,
\begin{equation}\label{eq:rendering_eq}
    \LightIntOut(\point, \view) = \int_{\Hemi} \brdf(\point, \light, \view) \LightIntIn(\point, \light) \cos\theta_\light \,\mathrm{d}\light.
\end{equation}
A plausible and physically realistic BRDF needs to fulfill three properties:
\emph{positivity} (\cref{eq:brdf_positivity}), \emph{Helmholtz reciprocity} (\cref{eq:brdf_reciprocity}), and \emph{energy conservation} (\cref{eq:brdf_energy_conservation}),
\begin{align}
    \brdf(\point, \light, \view)\geq 0,\qquad &\forall \point,\, \forall \light,\, \forall \view, \label{eq:brdf_positivity} \\
    \brdf(\point, \light, \view) = \brdf(\point, \view, \light),\qquad &\forall \point,\, \forall \light,\, \forall \view, \label{eq:brdf_reciprocity} \\
    \int_{\Hemi} \brdf(\point, \light, \view)\cos{\theta_{\view}} \,\mathrm{d}\view \le 1,\qquad &\forall \point,\, \forall \light. \label{eq:brdf_energy_conservation}
\end{align}
The first two properties are fulfilled by most state-of-the-art BRDF models~\cite{lafortune1994using,torrance1967theory,burley2012physically}. However, only very few models fulfill energy conservation by construction~\cite{lafortune1994using}. While for special algorithms like (bidirectional) path-tracing, reciprocity and energy conservation ensure convergence, in most cases, it is sufficient to fulfill them only approximately without noticeable artifacts \cite{akenine2019realTimeRendering}.

Mainly two physical processes are responsible for the reflection of light. %
These are often modeled as two separate terms within the BRDF: Surface reflection and subsurface scattering, often called specular and diffuse term.
In the first case, %
light is directly mirrored at the %
surface
and creates a %
view-dependent %
reflection lobe.
In the second case, %
light enters the %
surface and is scattered and partially absorbed %
until a fraction of light is re-emitted. 
While in general, %
subsurface scattering is not purely uniform, \ie, not Lambertian, but depending on the viewing direction~\cite{Oren94OrenNayar}, most models still assume a uniform diffuse reflection.
Light %
being reflected at the surface is %
absent for subsurface scattering~\cite{akenine2019realTimeRendering}, \ie, one can disjointly split the %
light used for surface reflection and for subsurface scattering.

A common special case are isotropic BRDFs which we consider in this paper.
In this case the reflectance at a point does not change if the object is rotated around 
the normal, or in other words, if the relative angle between the light and the view direction remains the same. 
In that case, three angles are sufficient for the BRDF parameterization.

\section{Neural BRDFs}
\begin{figure*}
    \centering
    \includegraphics[scale=1.0]{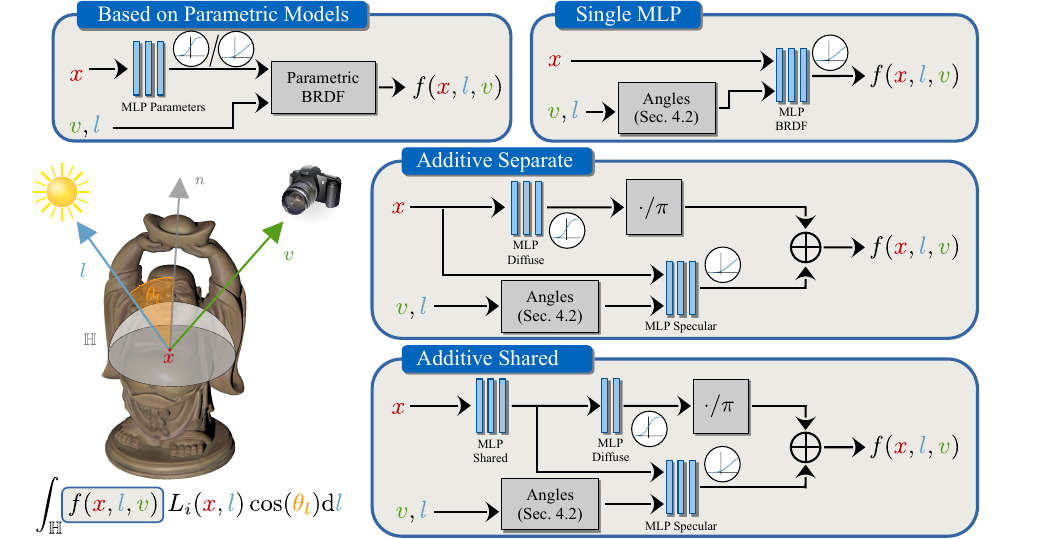}
    \caption{
    Overview of the neural models to compute the BRDF $f(x, l, v)$. Approaches based on a parametric model employ an MLP to predict the parameters for the respective model. A sigmoid or softplus output nonlinearity is used, depending on the range of the parameters. Methods based on a single MLP compute the BRDF value directly from the position and the view and light direction. Additive models split the reflection into a diffuse and a specular component. While the \emph{separate} architecture uses independent MLPs for both, the \emph{shared} architecture uses a common MLP with two separate heads.
    For all \emph{purely} neural approaches, the Rusinkiewicz angles \cite{rusinkiewicz1998new} are used to parametrize the directions. Moreover, we employ an intrinsic approach \cite{KoestlerIntrinsicNeuralFields22} to encode the position directly on the mesh as well as positional encoding for the angles. See \cref{sec:angleAndEncoding} for details on the angles and the encoding.
    }
    \label{fig:architectures}
\end{figure*}
This work aims to compare different neural BRDF modeling approaches in a unified manner. We deliberately choose a setting with \emph{given} geometry and \emph{calibrated} lights to avoid confounding effects due to joint estimation of geometry, reflectance, and light. This ensures that the results capture the capabilities of the models as clearly as possible. We use meshes to represent the geometry since they allow for accurate normals. We choose to reconstruct the BRDF from posed images rather than from BRDF measurements since this is the more practical 
and widely used setting.

In \cref{sec:approaches2NeuralBRDF}, we describe the neural BRDF models considered in this work. \cref{sec:angleAndEncoding} describes the angular parametrization of the directions and the input encodings used for the neural networks. Moreover, we propose two extensions for existing approaches to ensure reciprocity by construction and to enhance architectures that use an additive split of diffuse and specular parts in \cref{sec:reciMapping} and \ref{sec:enhancingAddSplit}.

\subsection{Approaches to Neural BRDF Modelling}
\label{sec:approaches2NeuralBRDF}
Since several ideas of previous methods are similar in spirit but implemented with slightly different architectures, we did our best to tune one architecture for each idea to yield optimal results for our data. Note that we focus on the actual BRDF representation; therefore, we do not include parts of the models concerned with the geometry estimation, like \eg additional regularizers.

\subsubsection{Neural BRDFs Based on Parametric Models}
This class of neural BRDFs is based on physically-based BRDF models to represent the reflective behavior and is used in several previous works \cite{bi2020neuralReflectanceFields,srinivasan2021nerv,Boss2021NERD,Zhang22IRON,Deschaintre2018SingleImageSVBRDFCaptureDeepNN,Henzler2021GenerativeModellingBRDFFlashIms,Guo2020MaterialGAN,Zhang21PhySG,Zhang2022ModellingIndirIlluminationInvRendering,brahimi24SparseViewsNearLight,Brahimi24SuperVol}. A neural network predicts the (spatially varying) parameters of the reflection model, which are then combined with the viewing and light direction to compute the BRDF values using the analytical formula of the parametric model. See \cref{fig:architectures} for a visualization.

In this work, we evaluate parametric neural BRDFs based on the \tslong (\ts) model~\cite{torrance1967theory} 
and the isotropic variant of
the \disney BRDF \cite{burley2012physically} as state-of-the-art parametric models. For reference, we also evaluate the energy-preserving variant of the Phong BRDF (\rp) \cite{lafortune1994using}. We refer to the appendix for more details on the models.

Finally, we compare against the recently proposed \fmbrdf model \cite{ichikawa2023fresnel}, which addresses shortcomings in the Oren-Nayar model. 
Since the normalization estimation for this model's normal distribution is computationally intractable for the spatially varying data, we estimate one single normalization per object. At the same time, all the other parameters depend on the position.
Note that for the semi-synthetic MERL-based data, this makes the estimation less complex since the material is uniform over the mesh.

We use an MLP with 6 layers of width 128 and a skip connection to the third layer to predict the parameters for all parametric models. We use ReLU nonlinearities between the layers and a sigmoid or a softplus output nonlinearity, depending on the range of the respective parameter. For the Disney and the Microfacet model, we found that the convergence is more stable when the output for the roughness is scaled by 0.5 before the last sigmoid function.

\subsubsection{Purely Neural BRDFs}
In contrast to the models in the previous section, purely neural models do not rely on a parametric model but directly predict the BRDF value using neural networks.

\paragraph{Single MLP}
Several previous works employ a single MLP to predict the BRDF value from the position $x$ and the light and view direction $l$ and $v$ directly \cite{sztrajman2021neural,Fan2022NeuralLayeredBRDF,Zeng23RelightingNeRFsWithShadowAndHighlightHints}. The adapted architecture for our experiments is shown in \cref{fig:architectures}. We map the directions to the Rusinkiewicz angles as described in \cref{sec:angleAndEncoding} and employ a softplus output nonlinearity to ensure the positivity of the resulting BRDF value. Again, we use a 6-layer MLP of width 128 with ReLU activations and an input skip to layer 3.

\paragraph{Additive Split}
Other works predict diffuse and specular reflections, which are additively combined into the final BRDF value. NeRFactor \cite{Zhang2021NeRFactor} uses two separate MLPs, which we adapt as \emph{Additive Separate}, see \cref{fig:architectures}. We remove their albedo clamping and use a 3D RGB specular part instead of the proposed scalar term since both yielded worse results. See the appendix for details and the experimental evaluation. We use a per-object specular part instead of the pre-trained module to facilitate a fair comparison between all methods without the need for pre-training.
For both MLPs, we use 4 layers of width 128 and a skip connection to the second layer, following the original work. For the diffuse MLP, we use a sigmoid output and divide by $\pi$ to obtain the Lambertian diffuse term. For the specular MLP, a softplus output ensures positivity.

Instead of two separate MLPs, LitNeRF \cite{Sarkar23LitNerf} employs a shared spatial MLP that extracts common features that are then used by a diffuse and specular head. We adopt this architecture as \emph{Additive Shared}, see \cref{fig:architectures}. We use 5 layers for the shared ReLU network of width 128 with a skip connection to the third layer, a single layer of width 128 for the diffuse MLP and two layers of width 128 for the specular output. Again, a sigmoid and a softplus output are used for the diffuse and the specular head, respectively. 
Both additive architectures use the angle parametrization introduced in the next section.

\subsection{Angle Parametrization and Input Encodings}
\label{sec:angleAndEncoding}
Following previous work \cite{Zhang2021NeRFactor, sztrajman2021neural}, we map the view and the light direction to the Rusinkiewicz angles \cite{rusinkiewicz1998new} before feeding them into the network. The most important advantage of this reparametrization is that the specular peaks align with the coordinate axes.  For an isotropic BRDF, the Rusinkiewicz angles read $(\theta_h,\theta_d,\phi_d)\in[0,\frac{\pi}{2}]^2\times[0, 2\pi]$. Please see the appendix for a more detailed discussion of the angle parameterizations and an experimental comparison to the common view-light angles.

Previous work has shown that vanilla MLPs have difficulties representing high-frequency data. Therefore, applying some encoding function to the input is common practice.
Since we parametrize the BRDF directly on the mesh, we encode the spatial position with the intrinsic encoding for neural fields on manifolds proposed by Koestler \etal \cite{KoestlerIntrinsicNeuralFields22}, which has been shown to be advantageous compared to common extrinsic encodings. We use positional encoding \cite{Mildenhall20Nerf} for the Rusinkiewicz angles. For more details on the encodings, we refer to the appendix.

\subsection{A Novel Mapping to Ensure Reciprocity}
\label{sec:reciMapping}
While the neural BRDFs based on parametric models fulfill the reciprocity constraint in \cref{eq:brdf_reciprocity} by construction, LitNeRF \cite{Sarkar23LitNerf} is the only purely neural approach that aims at fulfilling this constraint. During training, they randomly swap view and light direction to force the model to treat them interchangeably. However, this is a soft constraint, and it is not guaranteed that the reciprocity is fulfilled.
In contrast, we propose a mapping of the Rusinkiewicz angles that ensures that 
\cref{eq:brdf_reciprocity}
is fulfilled exactly by construction.

For an isotropic BRDF and the Rusinkiewicz angles, the reciprocity condition reads
\begin{equation}
    \brdf(\point,\theta_h,\theta_d,\phi_d) = \brdf(\point, \theta_h,\theta_d,\phi_d + \pi).
\end{equation}
We exploit this and map the Rusinkiewicz angles to
\begin{equation}
    \Theta = [\theta_h,\theta_d,\phi_{d,\pi},\phi_{d,\pi}+\pi], \label{eq:rusinkiewicz_reciprocity_mapping}
\end{equation}
where $\phi_{d,\pi}$ is short notation for $\phi_d$ modulo $\pi$ and $\text{range}(\Theta) = [0,\frac{\pi}{2}]^2\times\left[0, \pi\right]\times[\pi,2\pi]$. We now have
\begin{equation}
    \Theta(\theta_h,\theta_d,\phi_d) = \Theta(\theta_h,\theta_d,\phi_d + \pi).
\end{equation}
Hence, 
if we use $\Theta$ as input to the downstream blocks, the reciprocity constraint in \cref{eq:brdf_reciprocity} is fulfilled by construction.

\subsection{Enhancing the Additive Split}
\label{sec:enhancingAddSplit}
The existing purely neural models based on an additive split do not consider that light reflected at the surface cannot be used for subsurface scattering.
We propose a novel extension to additive split approaches that can model this phenomenon. Besides the specular reflection $f_\text{s}(x, l, v)$, we use the specular MLP to predict a weight $\xi(x, l, v)\in [0, 1]^3$, which we use to reduce the diffuse part $f_\text{d}(x)$ of the BRDF channel-wise. The resulting additive split reads
\begin{equation}
    f(x, l, v) = (1 - \xi(x, l, v)) \circ f_\text{d}(x) + f_\text{s}(x, l, v).
\end{equation}

Since the weight $\xi$ makes the diffuse summand view-dependent, which increases the ambiguity, 
we add two regularizers. We use an L1 loss between the diffuse part and the reference image to encourage the model to represent as much as possible with the diffuse part, and moreover, we use an L1 regularizer on the specular part to encourage sparsity. Please see the appendix for more details.
\section{Methodology and Datasets}
We estimate the BRDFs from a sparse set of HDR images from multiple viewpoints, each taken under changing directional lighting conditions. We assume mesh, camera poses, and light calibration to be known. This controlled setting ensures that the strengths of the reflection models can be assessed without confounding effects from estimating other quantities as well.

\subsection{Rendering}
For a single directional light, the irradiance $\LightIntIn$ is independent of the position $\point$, and the light direction $\light$ is constant for each view. In that case, the integral rendering equation in \cref{eq:rendering_eq}, reduces to a single evaluation and the rendering $\LightIntOut(\point, \view)$ for the pixel corresponding to $x$ and $v$ now reads
\begin{equation}
    \label{eq:rendering_single_dir_light}
    \LightIntOut(\point, \view) = \brdf(\point, \light, \view) \LightIntIn \mathbb{I}_s(\point, \light) \cos\theta_\light.
\end{equation}
The indicator function $\mathbb{I}_s(\point, \light)$ is used as a masking term to account for cast-shadows. Its value is computed by casting a secondary ray from the first intersection point $\point$ on the mesh in direction $\light$ and checking if the ray intersects the object. We apply a shadow bias to avoid self-shadow aliasing \cite{akenine2019realTimeRendering}. The intersection points $\point$ are computed by standard ray mesh intersection, and we compute the normal on the mesh by barycentric interpolation of the vertex normals.

\subsection{Loss and Training}
Since we are working with HDR images, which have a much larger dynamic range than sRGB images, a standard $L_2$ loss would be dominated by the bright regions, suppressing information from darker regions. To avoid this, 
we follow the analysis by Mildenhall \etal \cite{Mildenhall22NerfInTheDark} and
use a gamma correction function $\gamma: [0, 1]\mapsto [0,1]$ to construct a tone-mapped MSE-based loss term
\begin{equation}\label{eq:loss}
    \mathcal{L} = \frac{1}{N}\sum_{i=1}^N (\gamma(L_o(x, v)) - \gamma(L_{GT}(x, v))^2,
\end{equation}
where $L_{GT}(x, v)$ is the ground truth color of the pixel corresponding to $x$ and $v$ in linear color space. 

We train the models using the Adam optimizer \cite{Kingma14Adam} with a batch size of $N_{b}=2^{15}$ color values. For more details on the loss and the training, see the appendix.

\subsection{Datasets}
\label{sec:datasets}
We use the DiLiGenT-MV real-world multi-view dataset with calibrated lighting~\cite{Li2020DiLiGentMVDataset}. It consists of HDR images of $5$ objects with spatially varying, complex reflectance behaviors, taken from $20$ views, each captured under $96$ calibrated directional lights. We use the included ground truth meshes. %

Additionally, we create a semi-synthetic dataset based on real BRDF measurements to evaluate and analyze the performance in a more controlled setting. We render HDR images with 25 different \emph{uniform} MERL BRDFs \cite{matusik2003MERL} on 9 common 3D test meshes \cite{jacobson2020common}.
We add Gaussian noise with 
$\sigma = 10^{-3}$ to the images.
Besides the rendering, we also store the raw 
reflectance values to enable the direct evaluation of the BRDF prediction.
We train all models on $10$ views with $30$ lights each and use $12$ lights for each of the $10$ remaining views as the unseen test set for the evaluation.

\section{Experiments}
\begin{figure*}[t]  %
  \centering  %
  \footnotesize
  \newcommand{\mywidthc}{0.02\textwidth}  %
  \newcommand{\mywidthx}{0.11\textwidth}  %
  \newcommand{\mywidthw}{0.015\textwidth}  %
  \newcommand{\myheightx}{0.12\textwidth}  %
  \newcommand{\mywidtht}{0.035\textwidth}  %
  \newcolumntype{C}{ >{\centering\arraybackslash} m{\mywidthc} } %
  \newcolumntype{X}{ >{\centering\arraybackslash} m{\mywidthx} } %
  \newcolumntype{W}{ >{\centering\arraybackslash} m{\mywidthw} } %
  \newcolumntype{T}{ >{\centering\arraybackslash} m{\mywidtht} } %

  \newcommand{\heightcolorbar}{0.10\textwidth}  %
  \newcommand{\xposOne}{-0.95}
  \newcommand{\yposOne}{0.35}
  \newcommand{\xposTwo}{0.35}
  \newcommand{\yposTwo}{0.8}

  \newcommand{\fontsizePSNR}{\ssmall}
  
  \setlength\tabcolsep{0pt} %

  \setlength{\extrarowheight}{1.25pt}
  
  \def\arraystretch{0.8} %
  \begin{tabular}{TXXXXXXXWX}

\multirow{2}{*}{\rotatebox{90}{\parbox{3cm}{\centering \merlc \\ grease covered steel}}} &
\begin{tikzpicture}
\draw(0,0) node[inner sep=0] {\includegraphics[width=\mywidthx, height=\myheightx, keepaspectratio]{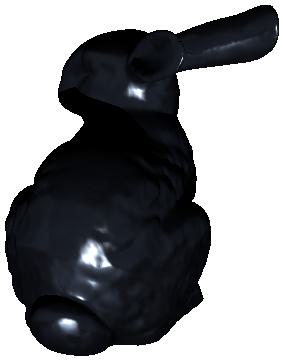}};
\node[align=left] at (\xposOne, \yposOne) {\fontsizePSNR PSNR: \\\fontsizePSNR 33.52};
\end{tikzpicture} &
\begin{tikzpicture}
\draw(0,0) node[inner sep=0] {\includegraphics[width=\mywidthx, height=\myheightx, keepaspectratio]{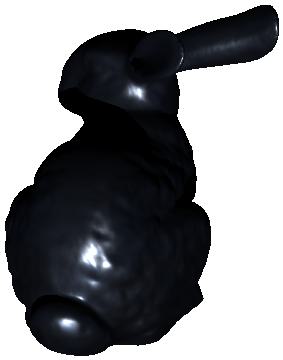}};
\node[align=left] at (\xposOne, \yposOne) {\fontsizePSNR PSNR: \\\fontsizePSNR 34.30};
\end{tikzpicture} &
\begin{tikzpicture}
\draw(0,0) node[inner sep=0] {\includegraphics[width=\mywidthx, height=\myheightx, keepaspectratio]{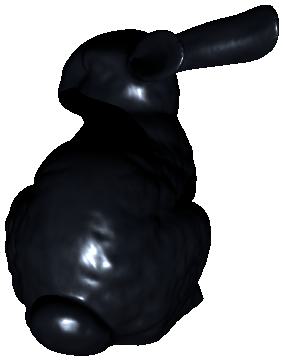}};
\node[align=left] at (\xposOne, \yposOne) {\fontsizePSNR PSNR: \\\fontsizePSNR 34.20};
\end{tikzpicture} &
\begin{tikzpicture}
\draw(0,0) node[inner sep=0] {\includegraphics[width=\mywidthx, height=\myheightx, keepaspectratio]{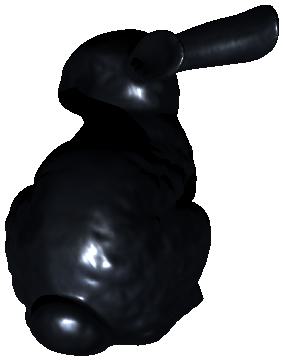}};
\node[align=left] at (\xposOne, \yposOne) {\fontsizePSNR PSNR: \\\fontsizePSNR 34.17};
\end{tikzpicture} &
\begin{tikzpicture}
\draw(0,0) node[inner sep=0] {\includegraphics[width=\mywidthx, height=\myheightx, keepaspectratio]{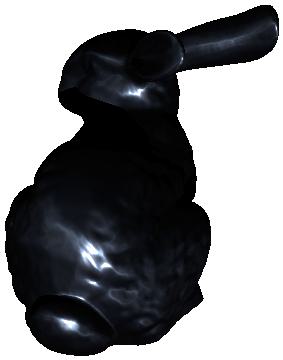}};
\node[align=left] at (\xposOne, \yposOne) {\fontsizePSNR PSNR: \\\fontsizePSNR 47.65};
\end{tikzpicture} &
\begin{tikzpicture}
\draw(0,0) node[inner sep=0] {\includegraphics[width=\mywidthx, height=\myheightx, keepaspectratio]{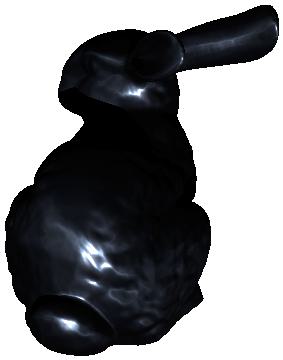}};
\node[align=left] at (\xposOne, \yposOne) {\fontsizePSNR PSNR: \\\fontsizePSNR 47.62};
\end{tikzpicture} &
\begin{tikzpicture}
\draw(0,0) node[inner sep=0] {\includegraphics[width=\mywidthx, height=\myheightx, keepaspectratio]{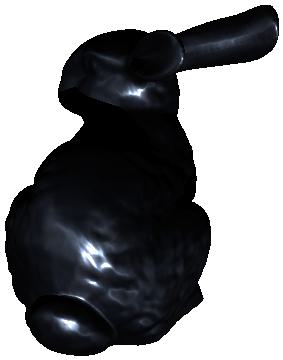}};
\node[align=left] at (\xposOne, \yposOne) {\fontsizePSNR PSNR: \\\fontsizePSNR 46.78};
\end{tikzpicture} &
& %
\includegraphics[width=\mywidthx, height=\myheightx, keepaspectratio]{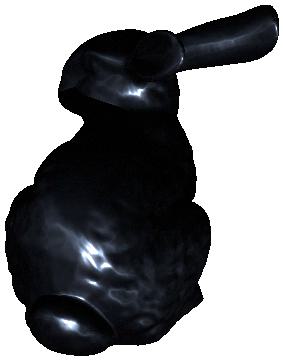}\\
 & \includegraphics[width=\mywidthx, height=\myheightx, keepaspectratio]{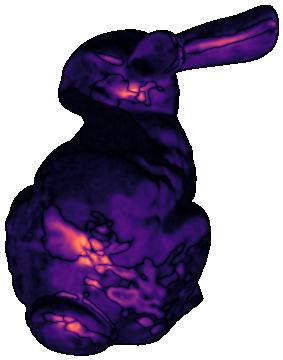} &
\includegraphics[width=\mywidthx, height=\myheightx, keepaspectratio]{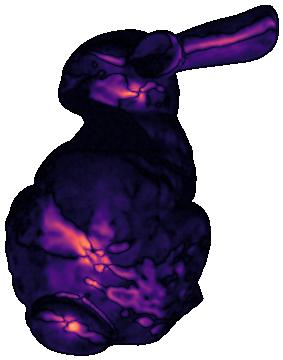} &
\includegraphics[width=\mywidthx, height=\myheightx, keepaspectratio]{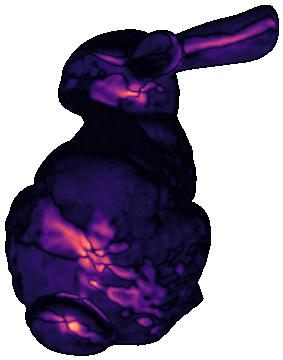} &
\includegraphics[width=\mywidthx, height=\myheightx, keepaspectratio]{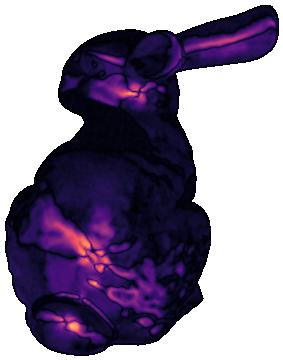} &
\includegraphics[width=\mywidthx, height=\myheightx, keepaspectratio]{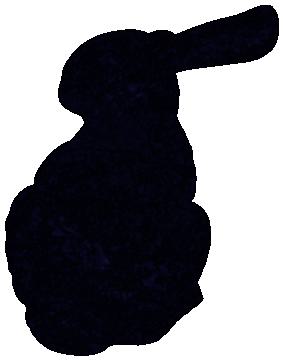} &
\includegraphics[width=\mywidthx, height=\myheightx, keepaspectratio]{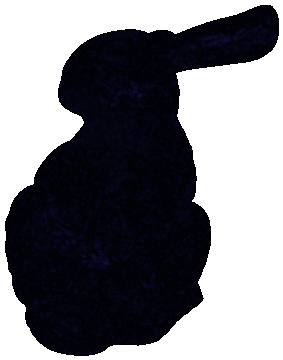} &
\includegraphics[width=\mywidthx, height=\myheightx, keepaspectratio]{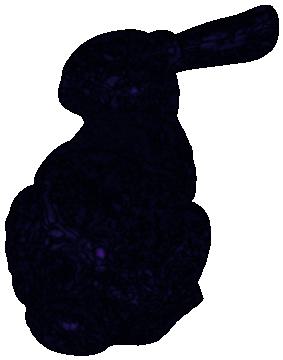} &
& %
\includegraphics[width=\mywidthx, height=\heightcolorbar, keepaspectratio]{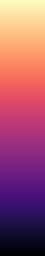} \\
\hline
\multirow{2}{*}{\rotatebox{90}{\parbox{3cm}{\centering \diligentc \\ cow}}} &
\begin{tikzpicture}
\draw(0,0) node[inner sep=0] {\includegraphics[width=\mywidthx, height=\myheightx, keepaspectratio]{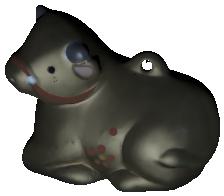}};
\node[align=left] at (\xposTwo, \yposTwo) {\fontsizePSNR PSNR: 41.43};
\end{tikzpicture} &
\begin{tikzpicture}
\draw(0,0) node[inner sep=0] {\includegraphics[width=\mywidthx, height=\myheightx, keepaspectratio]{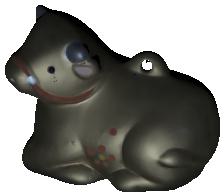}};
\node[align=left] at (\xposTwo, \yposTwo) {\fontsizePSNR PSNR: 41.76};
\end{tikzpicture} &
\begin{tikzpicture}
\draw(0,0) node[inner sep=0] {\includegraphics[width=\mywidthx, height=\myheightx, keepaspectratio]{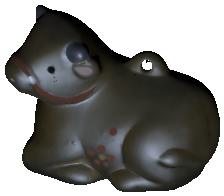}};
\node[align=left] at (\xposTwo, \yposTwo) {\fontsizePSNR PSNR: 39.61};
\end{tikzpicture} &
\begin{tikzpicture}
\draw(0,0) node[inner sep=0] {\includegraphics[width=\mywidthx, height=\myheightx, keepaspectratio]{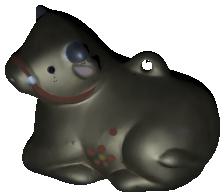}};
\node[align=left] at (\xposTwo, \yposTwo) {\fontsizePSNR PSNR: 41.80};
\end{tikzpicture} &
\begin{tikzpicture}
\draw(0,0) node[inner sep=0] {\includegraphics[width=\mywidthx, height=\myheightx, keepaspectratio]{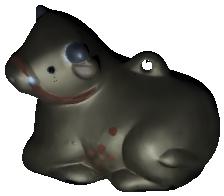}};
\node[align=left] at (\xposTwo, \yposTwo) {\fontsizePSNR PSNR: 41.39};
\end{tikzpicture} &
\begin{tikzpicture}
\draw(0,0) node[inner sep=0] {\includegraphics[width=\mywidthx, height=\myheightx, keepaspectratio]{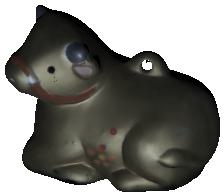}};
\node[align=left] at (\xposTwo, \yposTwo) {\fontsizePSNR PSNR: 41.52};
\end{tikzpicture} &
\begin{tikzpicture}
\draw(0,0) node[inner sep=0] {\includegraphics[width=\mywidthx, height=\myheightx, keepaspectratio]{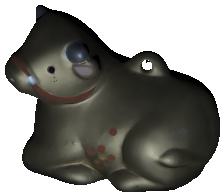}};
\node[align=left] at (\xposTwo, \yposTwo) {\fontsizePSNR PSNR: 42.09};
\end{tikzpicture} &
& %
\includegraphics[width=\mywidthx, height=\myheightx, keepaspectratio]{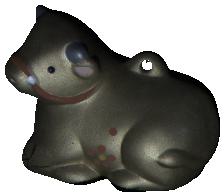}\\
 & \includegraphics[width=\mywidthx, height=\myheightx, keepaspectratio]{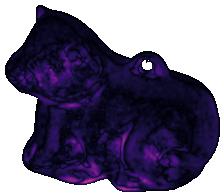} &
\includegraphics[width=\mywidthx, height=\myheightx, keepaspectratio]{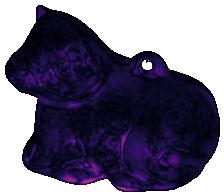} &
\includegraphics[width=\mywidthx, height=\myheightx, keepaspectratio]{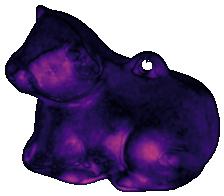} &
\includegraphics[width=\mywidthx, height=\myheightx, keepaspectratio]{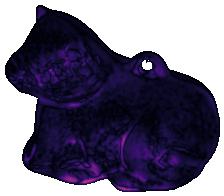} &
\includegraphics[width=\mywidthx, height=\myheightx, keepaspectratio]{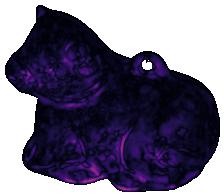} &
\includegraphics[width=\mywidthx, height=\myheightx, keepaspectratio]{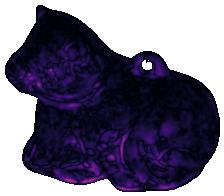} &
\includegraphics[width=\mywidthx, height=\myheightx, keepaspectratio]{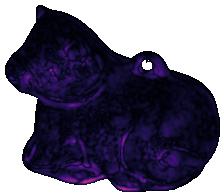} &
& %
\includegraphics[width=\mywidthx, height=\heightcolorbar, keepaspectratio]{figures/renderings/qualitative/colorbar_magma.jpg} \\
\hline\\[-0.2cm]
 & \cellcolor{cellParamBased}\rpc		%
 & \cellcolor{cellParamBased}\tsc		%
 & \cellcolor{cellParamBased}\fmbrdfc		%
 & \cellcolor{cellParamBased}\disneyc		%
 & \cellcolor{celPurelyNeural}Single MLP		%
 & \cellcolor{celPurelyNeural}Add Sep		%
 & \cellcolor{celPurelyNeural}Add Shared		%
& \cellcolor{celPurelyNeural} %
& \gt \\
  \end{tabular}
  \caption{Qualitative evaluation of the reconstruction for the MERL grease covered steel BRDF \cite{matusik2003MERL} uniformly rendered on the bunny mesh from \cite{jacobson2020common} and the cow object from the real-world \diligent data \cite{Li2020DiLiGentMVDataset}. Shown are renderings in sRGB space with the corresponding PSNR values and the \FLIP error maps for the sRGB renderings.
  Only the purely neural approaches (\mysquare[celPurelyNeural]) are able to reconstruct the intricate reflection patterns for the grease covered steal faithfully. For the cow, the results for the parametric models (\mysquare[cellParamBased]) are much more on par, and all models show difficulties in similar areas -- in particular in recesses, which suggests unmodeled interreflections as a potential reason.
  }
\label{fig:evaluation_renderings}
\end{figure*}
In this section, we present a thorough evaluation of the neural BRDF approaches. 
First, we conduct a qualitative and quantitative comparison on the semi-synthetic and real-world datasets described in \cref{sec:datasets}. Subsequently, we analyze several aspects of the models, including energy conservation and the approaches to reciprocity, in more depth.
We refer to the appendix for more information on the models and the datasets, as well as several additional experiments.

\paragraph{Evaluation Metrics}

We transform the renderings of the test set from linear to sRGB space and evaluate  \psnr, \lpipsc and DSSIM, where the latter is based on \ssimc by $\text{DSSIM} = (1-\text{SSIM})/2$. Moreover, we report the HDR version of the \FLIP metric \cite{Andersson2021HDRFLIP,Andersson2020FLIP}. For each metric, we average the results of the respective dataset.

Following \cite{Lavoue21PerceptualQualityOfBRDFApproximations}, we 
report the root mean squared error between the cubic root of the predicted and the \gt BRDF values for the synthetic experiments (RMSE$^{\sqrt[3]{}}$). As analyzed in \cite{Lavoue21PerceptualQualityOfBRDFApproximations}, we discard values close to the horizon from the data. Moreover, we exclude values from saturated pixels. 
This BRDF-space metric correlates well with perceptual similarity. 
We refer to the supplement for more details.

\subsection{Comparison of the BRDF Models}
\label{sec:comparison_brdf_models}

\begin{table*}[t]  %
\centering  %
  \footnotesize
    \begin{tabular}{l||lllll|llll}
& \multicolumn{5}{c|}{\merlc} & \multicolumn{4}{c}{\diligentc} \\

 & RMSE$^{\sqrt[3]{}}$$\downarrow$ & \psnrArrow & \dssimArrow & \lpipsArrow & \flipArrow & \psnrArrow & \dssimArrow & \lpipsArrow & \flipArrow \\ \hline \hline\cellcolor{cellParamBased}\rplongc 		%
&  1.632 &  44.05 & 0.779 & 1.656 & 6.025 & 41.59 & 0.775 & 1.771 &  3.119 \\
\cellcolor{cellParamBased}\tslongc 		%
&  1.216 &  46.06 & 0.620 & 1.174 & 5.717 & 41.83 & 0.757 & 1.763 &  3.054 \\
\cellcolor{cellParamBased}\fmbrdflongc 		%
&  1.167 &  46.40 & 0.654 & 1.310 & 5.876 & 41.42 & 0.752 & 1.781 &  3.228 \\
\cellcolor{cellParamBased}\disneyc 		%
&  1.218 &  46.46 & 0.616 & 1.158 & 5.735 & \hlThirdBest{{41.89}} &  0.750 & 1.788 &  3.056 \\
\hline
\cellcolor{celPurelyNeural}Single MLP 		%
&  \hlSecondBest{{0.554}} &  \hlSecondBest{{52.35}} &  \hlBest{\textbf{0.411}} &  \hlBest{\textbf{0.565}} &  \hlSecondBest{{4.795}} &  41.64 & 0.789 & 1.765 &  2.962 \\
\cellcolor{celPurelyNeural}Additive Separate 		%
&  \hlThirdBest{{0.562}} &  \hlThirdBest{{52.28}} &  \hlThirdBest{{0.413}} &  \hlThirdBest{{0.571}} &  \hlThirdBest{{4.802}} &  41.77 & 0.758 & 1.702 &  2.924 \\
\cellcolor{celPurelyNeural}Additive Shared 		%
&  0.665 &  51.40 & 0.428 & 0.621 & 5.075 & \hlSecondBest{{42.35}} &  \hlSecondBest{{0.714}} &  \hlThirdBest{{1.671}} &  \hlSecondBest{{2.856}} \\
\hline \hline 
\cellcolor{celPurelyNeural}Additive Separate (enhanced) 		%
&  \hlBest{\textbf{0.547}} &  \hlBest{\textbf{52.39}} &  \hlSecondBest{{0.412}} &  \hlSecondBest{{0.565}} &  \hlBest{\textbf{4.794}} &  41.88 & \hlThirdBest{{0.744}} &  \hlSecondBest{{1.663}} &  \hlThirdBest{{2.883}} \\
\cellcolor{celPurelyNeural}Additive Shared (enhanced) 		%
&  0.579 &  51.95 & 0.419 & 0.580 & 4.836 & \hlBest{\textbf{42.38}} &  \hlBest{\textbf{0.709}} &  \hlBest{\textbf{1.659}} &  \hlBest{\textbf{2.827}} \\

\end{tabular}
  
  \caption{
  Quantitative comparison of the BRDF models on the MERL-based semi-synthetic dataset \cite{matusik2003MERL} and the real-world data \cite{Li2020DiLiGentMVDataset}.
  RMSE$^{\sqrt[3]{}}$ is the RMSE of the cubic root of the BRDF values. \psnr, DSSIM, \lpips are computed for the sRGB renderings. All quantities are first averaged over one object and then averaged over all objects in the respective dataset. RMSE$^{\sqrt[3]{}}$, DSSIM, LPIPS and \FLIP are scaled by 100.
  While the purely neural approaches (\mysquare[celPurelyNeural]) show superior results over the parametric models (\mysquare[cellParamBased]) on the semi-synthetic data, the difference is much smaller on the real-world data. We attribute this to more materials with complex reflections in the MERL BRDFs and potentially unmodelled noise patterns and interreflections in the DiLiGenT-MV dataset.
  }
\label{tab:quantitative}
\end{table*}

We show a qualitative comparison of the models in \cref{fig:evaluation_renderings} and report the metrics in \cref{tab:quantitative}.
For the semi-synthetic examples based on the MERL data \cite{matusik2003MERL}, the results show a significant advantage of the purely neural approaches, outperforming the methods based on parametric models by a large margin on all metrics. The examples in the top rows of \cref{fig:evaluation_renderings} and \cref{fig:teaser} confirm that only the purely neural approaches can faithfully capture the complex reflection patterns of highly specular materials. For the real-world data, the difference between the models is much less significant. One reason is the absence of highly specular materials in the DiLiGenT-MV dataset. Another potential reason are interreflections in the real-world data, which we do not model. Finally, there might be noise in the real-world data not captured by the simple noise model used to create the semi-synthetic data. Indeed, we see a higher influence of noise on the purely neural models: If we remove the noise from the semi-synthetic data, the performance gains for those models are much higher than for the parametric models.

Among the parameter-based models, the Disney BRDF \cite{burley2012physically} performs best, which is expected since it is the most sophisticated model. For the FMBRDF \cite{ichikawa2023fresnel}, we observe a ``glow'' for the cow from the real-world data (see \cref{fig:evaluation_renderings}). We hypothesize that this originates from the scalar specular term since a similar effect is observed for the scalar additive architecture, as shown in the appendix. This effect does not appear for other objects.

For the purely neural approaches, we see a different behavior between the 
two datasets, 
with the ranking order being reversed. The reason seems to be the number of layers that the view and the light direction are fed through. While it is only 2 layers for the additive shared architecture, there are 4 for additive separate and 6 for the single MLP architectures. We investigate this further in the next section.

Our enhancement for the additive split (\cf \cref{sec:enhancingAddSplit}) shows slight improvements for both additive approaches. This confirms that the additional freedom can help the model to adapt better to the underlying BRDF functions.

\subsection{Analysis of the BRDF Models}
\label{sec:analysis_brdf_models}

\paragraph{Number of Layers for View/Light Directions}
\newcommand\stringNLayersDirs{Number layers view/light direction}
\newcommand\stringReci{Reciprocity Mapping}
\newcommand{\spaceBetweenExps}{0mm}
\newcommand{\spaceAfterMetrics}{0mm}

\begin{table*}[t]  %
\centering  %
  \footnotesize
  \vspace{-0.15cm}
    \begin{tabular}{l||ccccc|cccc}
& \multicolumn{5}{c|}{\merlc} & \multicolumn{4}{c}{\diligentc} \\

 & $\Delta$RMSE$^{\sqrt[3]{}}$ & $\Delta$\psnr & $\Delta$\dssim & $\Delta$\lpips & $\Delta$\flip & $\Delta$\psnr & $\Delta$\dssim & $\Delta$\lpips & $\Delta$\flip \\ \hline \hline
Single MLP$^\ast$ (NOL dirs $6\rightarrow 3$) 		%
&   \textcolor{red}{+0.01} &   \textcolor{red}{-0.15} &   \textcolor{red}{+0.01} &   \textcolor{red}{+0.01} &   \textcolor{red}{+0.02} &  \textcolor{greenValid}{+0.51} &  \textcolor{greenValid}{-0.06} &  \textcolor{greenValid}{-0.10}  &  \textcolor{greenValid}{-0.09} \\
Additive Sep.$^\ast$ (NOL dirs $4\rightarrow 2$) 		%
&   \textcolor{red}{+0.02} &   \textcolor{red}{-0.09} &   \textcolor{red}{+0.01} &   \textcolor{red}{+0.01} &   \textcolor{red}{+0.02} &  \textcolor{greenValid}{+0.26} &  \textcolor{greenValid}{-0.01} &  \textcolor{red}{+0.02}  &  \textcolor{red}{+0.01} \\
Additive Shared$^\ast$ (NOL dirs $2\rightarrow 4$) 		%
&   \textcolor{greenValid}{-0.09} &   \textcolor{greenValid}{+0.71} &   \textcolor{greenValid}{-0.01} &   \textcolor{greenValid}{-0.05} &   \textcolor{greenValid}{-0.25} &  \textcolor{red}{-0.37} &  \textcolor{red}{+0.01} &  \textcolor{red}{+0.01}  &  \textcolor{red}{+0.02} \\

\end{tabular}
\vspace{-0.15cm}
  
  \caption{
  Quantitative changes for varying the number of layers (NOL) for the directions. Shown is the difference to the results in \cref{tab:quantitative}; RMSE$^{\sqrt[3]{}}$, DSSIM, LPIPS and \FLIP are scaled by 100. The experiments confirm that reducing the NOL for the directions tends to increase the reconstruction quality for the real-world data while simultaneously decreasing it slightly for the semi-synthetic data. Increasing the NOL has the opposite effect. We hypothesize, that more layers for the directions enable the model to learn the more complex reflection patterns in the MERL data, while simultaneously making the model less robust to noise that might be contained in the real-world data.
  \vspace{-0.15cm}
  }
\vspace{-0.1cm}
\label{tab:changes_quantitative_NOL}
\end{table*}
To confirm that, indeed, the difference in the performance of the purely neural models for the real-world and semi-synthetic data originates from the number of layers (NOL) for the directions, we perform experiments where we vary this parameter for each of the purely neural models. 

The results are shown in \cref{tab:changes_quantitative_NOL}, where we indicate the change in the NOL for the directions per model. Please see the appendix for the detailed architecture changes. The numbers confirm the trend that reducing the NOL decreases the reconstruction quality for the MERL examples while increasing the quality for the DiLiGenT-MV data while increasing the NOL has the opposite effect. This may suggest that more layers for the directions are beneficial for complex reflective patterns as contained in the MERL data, but at the same time, make the model less robust to potentially noisy measurements of real-world data.

\paragraph{Reciprocity Mapping}
\definecolor{Gray}{gray}{0.85}

\begin{table}[t]  %
  \centering  %
  \footnotesize  %

\def\arraystretch{1.1}
\begin{tabular}{l||c|c}

& \merlc & \diligentc \\
\hline \hline

 Single MLP 		%
& $2.93\cdot 10^{-2}$ & 2.52 \\
 Add Sep 		%
& $2.62\cdot 10^{-2}$ & 1.94 \\
 Add Shared 		%
& $4.70\cdot 10^{-3}$ & $6.78\cdot 10^{-1}$ \\
\hline & \\ [-9pt]
 Single MLP (rnd. in-swap) 		%
& $5.42\cdot 10^{-4}$ & $2.21\cdot 10^{-2}$ \\
 Add Sep (rnd. in-swap) 		%
& $4.87\cdot 10^{-4}$ & $1.61\cdot 10^{-2}$ \\
 Add Shared (rnd. in-swap) 		%
& $1.60\cdot 10^{-4}$ & $1.11\cdot 10^{-3}$ \\

\end{tabular}
\vspace{-0.1cm}

\caption{
RMSE of the violation of the reciprocity constraint \cref{eq:brdf_reciprocity}. The error is significant without any reciprocity strategy, in particular for the real-world data. While the random input swap strategy in \cite{Sarkar23LitNerf} reduces the error, the constraint is still violated. In contrast, our mapping ensures reciprocity exactly by construction (and therefore is not shown here). The RMSE is \emph{not} scaled here.
}
\vspace{-0.1cm}
\label{tab:reciprocity}
\end{table}

To evaluate the random input swap training strategy used in LitNerf \cite{Sarkar23LitNerf} as well as our input mapping proposed in \cref{sec:reciMapping}, we quantify how much the symmetry constraint is violated by computing the RMSE between the BRDF values with changed positions, i.e. for the pairs $\brdf(\point, \light, \view)$ and $\brdf(\point, \view, \light)$. We report the results for the random input swap and the models without any reciprocity strategy in \cref{tab:reciprocity}. Note that our approach fulfills the constraint exactly by construction and is therefore not included in the table. We see, that without any reciprocity strategy, the constraint is violated significantly, particularly for the real-world data. While the random input swap does reduce this error by several orders of magnitude, the constraint is still violated, which might cause problems for particular algorithms. In contrast, \cref{eq:brdf_reciprocity} is fulfilled by construction by our approach.
The influence on the reconstruction quality is analyzed in the appendix. Apart from the single MLP architecture, for which the random input swap performs slightly better on the real-world data, both approaches have a similar effect, and we obtain only slightly worse results as a trade-off for the ensured reciprocity.

\paragraph{Energy Conservation}
\definecolor{Gray}{gray}{0.85}

\begin{table}[t]  %
  \centering  %
  \footnotesize  %

\def\arraystretch{1.1}
\begin{tabular}{l||c!{\color{Gray}\vrule}c|c!{\color{Gray}\vrule}c}     %

& \multicolumn{2}{c|}{\merlc} & \multicolumn{2}{c}{\diligentc} \\

& \scriptsize $\%> 1$ & \scriptsize $\mu_{.5}> 1$ & \scriptsize $\%> 1$ & \scriptsize $\mu_{.5}> 1$ \\[2pt] \hline \hline

\cellcolor{cellParamBased}\rpc		%
&  \textbf{none} &  \textcolor{green}{\textbf{all valid}} &  \textbf{none} &  \textcolor{green}{\textbf{all valid}}  \\
\cellcolor{cellParamBased}\tsc		%
&  0.02\% &  \textcolor{red}{1.1} & 0.91\% &  \textcolor{red}{1.1} \\
\cellcolor{cellParamBased}\fmbrdfc		%
&  0.01\% &  \textcolor{red}{1.1} & 1.29\% &  \textcolor{red}{1.1} \\
\cellcolor{cellParamBased}\disneyc		%
&  0.05\% &  \textcolor{red}{1.1} & 1.64\% &  \textcolor{red}{1.1} \\
\hline
\cellcolor{celPurelyNeural}Single MLP		%
&  0.05\% &  \textcolor{red}{1.5} & 5.17\% &  \textcolor{red}{3.7} \\
\cellcolor{celPurelyNeural}Add Sep		%
&  0.02\% &  \textcolor{red}{1.2} & 8.01\% &  \textcolor{red}{2.4} \\
\cellcolor{celPurelyNeural}Add Shared		%
&  0.01\% &  \textcolor{red}{1.2} & 4.43\% &  \textcolor{red}{1.5} \\

\end{tabular}
\vspace{-0.1cm}

  \caption{
  Analysis of energy conservation (\cref{eq:brdf_energy_conservation}) on the MERL-based synthetic dataset \cite{matusik2003MERL} and the real-world data from \diligentc. Shown are the percentage of the 50k sampled point-light pairs violating the energy conservation ($\%> 1$) and the median of the energies larger than 1 ($\mu_{.5}> 1$),
  each averaged over all experiments of the respective dataset. 
  }
\vspace{-0.1cm}
\label{tab:energy}
\end{table}
To analyze how well the models fulfill the energy conservation in \cref{eq:brdf_energy_conservation}, we analyze $50k$ randomly sampled point-light pairs for each object from the test data. We approximate the energy integral in \cref{eq:brdf_energy_conservation} by Monte-Carlo (MC) integration, where we use cosine-weighted hemisphere sampling and draw $20k$ view direction samples for each integral to ensure convergence. A more detailed discussion can be found in the appendix.

The results in \cref{tab:energy} show that for the semi-synthetic data, all approaches fulfill energy conservation almost everywhere. For the real-world data, we observe more significant violations; in particular for the purely neural methods, which might again indicate measurement noise. The fully separated additive approach seems to be particularly disadvantageous for energy conservation.

\section{Conclusion}
In this work, we have presented an exhaustive comparison of different approaches to neural BRDF modeling. The results show that while purely neural approaches have advantages for materials with complex reflective patterns, the performance of methods based on parametric models on real-world data with less complex reflection patterns is comparable.
We found signs that purely neural methods cannot learn reciprocity and energy conservation from data, particularly for real-world images, and we have analyzed approaches to ensure reciprocity, including a newly proposed input mapping that ensures reciprocity by construction.
Finally, we have presented an extension to models based on additive splits that aims to capture the physical behavior better and shows improvements for existing approaches.

{
\small
\paragraph{Acknowledgements}
This work was supported by the ERC Advanced Grant SIMULACRON.
}

\newpage
{\small
\bibliographystyle{ieee_fullname}
\bibliography{references}
}

\newpage
\appendix
{
    \twocolumn[
    \centering
    \Large
    \textbf{\paperTitle}\\
    \vspace{0.5em}Supplementary Material \\
    \vspace{2.0em}
    ]
}

\counterwithin{figure}{section}
\counterwithin{table}{section}

This supplementary material gives additional information on several aspects of our work. In \cref{sec:supp:add_details_models}, we present more details on the neural BRDF modeling, including a discussion of the angles and an overview of the parametric models. In \cref{sec:supp:training_and_eval}, we share more insight on the training and the evaluation, including additional details on the loss formulation and the regularizers for the additive enhancement. Finally, we present several additional experiments in \cref{sec:supp:add_experimental_results}, among them additional renderings of the objects and an analysis of the angle parametrization.
\section{Additional Details on the Models}
\label{sec:supp:add_details_models}

In this section, we give additional details on the neural BRDF modeling. \cref{sec:supp:angle_definitions} reviews the Rusinkiewicz angles used to parametrize the view and light direction, \cref{sec:supp:intrinsic_neural_encoding} gives an overview of the intrinsic encoding used to parametrize the neural BRDFs directly on the meshes. Finally, \cref{sec:supp:parametric_models} reviews the parametric models used to construct the parametric neural BRDFs.

\subsection{Angle Definitions}
\label{sec:supp:angle_definitions}

\begin{figure*}[t]  %
  \centering  %
  \small  %
  \newcommand{\mywidthc}{0.02\textwidth}  %
  \newcommand{\mywidthx}{0.45\textwidth}  %
  \newcolumntype{C}{ >{\centering\arraybackslash} m{\mywidthc} } %
  \newcolumntype{X}{ >{\centering\arraybackslash} m{\mywidthx} } %
  \setlength\tabcolsep{1pt} %
  \def\arraystretch{1} %
  \begin{tabular}{XX}
    \includegraphics[width=\mywidthx]{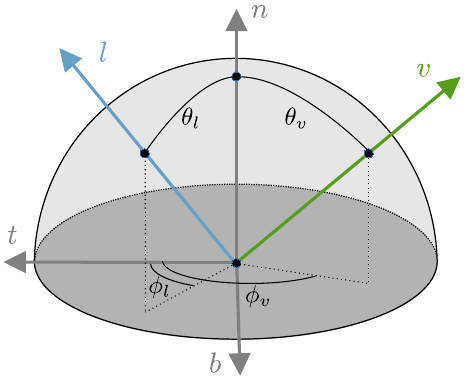} &
    \includegraphics[width=\mywidthx]{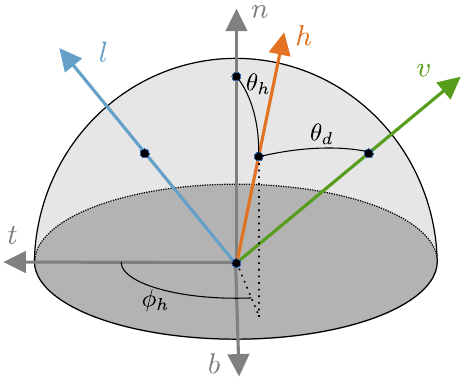}\\
    View-Light Angles & Rusinkiewicz Angles
  \end{tabular}
  \caption{Visualization of the view-light and the Rusinkiewicz angles on the left and on the right, respectively. Shown are view and light direction, $\view$ and $\light$ in a local coordinate system given by surface normal $n$, a surface tangent $t$ (which is arbitrary in the case of isotropic BRDFs) and the binormal $b=n\times t$. The view-light angles are simply the polar angles of $\view$ and $\light$, while the Rusinkiewicz angles are given in terms of the half angle $h=\frac{\view + \light}{\|\view + \light\|}$ and a ``difference'' vector $d$ which is the light direction $\light$ in a coordinate system with $h$ as the normal.}
\label{fig:angles}
\end{figure*}

While the BRDF is often defined in terms of the view and the light unit directions, $\view$ and $\light$, respectively, those two quantities are elements on the (curved) $\Sphere^2$ manifold, which makes them impractical to work with. Therefore, a parametrization in terms of the angles of the vectors is usually employed. The most obvious choice is to use the polar angles $(\theta_\view, \phi_\view)$ and $(\theta_\light, \phi_\light)$ of the two vectors directly, see \cref{fig:angles} on the left. We refer to this parametrization as \emph{view-light angles}. For an isotropic BRDF, the absolute orientation in the tangential plane is irrelevant, and only the relative orientation of the vectors with respect to each other is important. In this case the triplet $(\theta_\view, \theta_\light, \phi_\view - \phi_\light)$ is sufficient to parameterize the BRDF.

However, as noted by Rusinkiewicz, the view-light angles are suboptimal to parameterize a BRDF \cite{rusinkiewicz1998new}. They propose a novel parameterization that aligns important features of the BRDF, like the specular peak, with the coordinate axes. The representation is based on the half-vector

\begin{equation}\label{eq:half_angle}
    h = \frac{\view + \light}{\|\view + \light\|}.
\end{equation}

and a ``difference'' vector $d$, which is obtained by rotating the light direction into a coordinate system, where the half vector $h$ coincides with the normal, see \cref{fig:angles} on the right for an overview. We denote the surface normal by $n$ and use an arbitrary tangential vector $t$ and the binormal $b=n\times t$ to define a local coordinate system. The difference vector $d$ is computed as 
\begin{equation}
    d=\mathrm{rot}_{b,-\theta_h}\mathrm{rot}_{n, -\phi_h}l,
\end{equation}
where $\mathrm{rot}_{x, \alpha}y$ means the rotation of vector $y$ around the axis $x$ by the angle $\alpha$. The full Rusinkiewicz angles read $(\theta_h, \phi_h, \theta_d, \phi_d)$. Isotropic BRDFs do not depend on $\phi_h$; therefore, the triplet $(\theta_h, \theta_d, \phi_d)$ is sufficient for the parametrization in this case. See \cite{rusinkiewicz1998new} for more details.

\subsection{Neural Intrinsic Encoding}
\label{sec:supp:intrinsic_neural_encoding}

For the positional encoding on the manifold, we use the neural intrinsic encoding described by Koestler \etal \cite{KoestlerIntrinsicNeuralFields22}, who propose to use a subset of the eigenfunctions of the  Laplace-Beltrami operator (LBO)\footnote{We refer to \cite{Rustamov07Laplace-BeltramiEigenfunctions} for more details on the LBO.} for the encoding. Since the LBO is a generalization of the Laplace operator on an Euclidean domain, this form of encoding can be seen as an adaption of positional encoding from the Euclidean domain to manifolds like meshes. Given a closed, compact manifold $\mathcal{M}\subset\mathbb{R}^n$, we denote the eigenfunctions of the LBO on $\mathcal{M}$ by $\zeta_j$. For the encoding, we consider a subset of indices $\mathcal{I}\subset \mathbb{N}$ and, given a point $\point$ on the mesh, use the encoding
\begin{equation}
    \Theta_{LBO}(\point) = (\zeta_j(\point))_{j\in\mathcal{I}}.
\end{equation}
In the discrete setting of a mesh, the values of the LBO eigenfunctions are computed on the vertices. We use barycentric interpolation to obtain the encoding for an arbitrary position on the mesh.
In practice, it is not necessary to use a connected index set. We follow \cite{KoestlerIntrinsicNeuralFields22} and use several connected blocks of eigenfunctions for the encoding; see \cref{sec:supp_lbo_params} for the concrete numbers.

\subsection{Parametric BRDF Models}
\label{sec:supp:parametric_models}

In the following section, we give a short overview of the parametric models used in this work. We refer to the respective works of more details.

\paragraph{Realistic Phong Model}
We use the model described in \cite{lafortune1994using}. To ensure energy conservation, we predict a combined value $k_\mathrm{full}\in[0,1]^3$ for the reflectivities as well as a split percentage $\zeta\in[0,1]^3$ and compute the individual reflectivities as $k_d = \zeta\circ k_\mathrm{full}$ and $k_s = (1-\zeta)\circ k_\mathrm{full}$.
To ensure the correct range, we use a sigmoid nonlinearity for $k_\mathrm{full}$ and $\zeta$. The specular exponent $n\geq1$ is computed using a softplus output function with an additive offset of 1.

\paragraph{Torrance-Sparrow Model}
For the specular part, we use the basic form of the microfacet model described in \cite{torrance1967theory}, which reads (omitting the dependence on the position)
\begin{equation}
    f_\mathrm{spec}(\light, \view) = \frac{D(h)F(\view, h)G(\light, \view, h)}{4\langle n, l\rangle\langle n, v\rangle},
\end{equation}
where $h$ is the half vector described in \cref{eq:half_angle}, $D$ is the normal distribution function (NDF), $F$ is the Fresnel term, $G$ is the geometric shadowing term and $n\in\Sphere^2$ is the surface normal in the local coordinate system. By $\langle n, l\rangle$ we denote the scalar product between $n$ and $l$.

We use the (isotropic) Trowbridge-Reitz/GGX distribution~\cite{smith1967geometrical, walter2007microfacet, trowbridge1975average} for $D$, which reads 
\begin{equation}
    D(h) = \frac{\alpha^2}{\pi(\langle n, h\rangle^2(\alpha^2 - 1) + 1)^2}
\end{equation}
with the roughness parameter $\alpha>0$.
For $G$, we use smiths method \cite{smith1967geometrical}, which splits the term multiplicatively as
\begin{equation}
    G(\light, \view, h) = \Tilde{G}(\light)\Tilde{G}(\view).
\end{equation}
We use the Trowbridge-Reitz/GGX term~\cite{smith1967geometrical, walter2007microfacet, trowbridge1975average} for $\Tilde{G}$, which for a vector $w\in\mathbb{R}^3$ reads
\begin{equation}
    \Tilde{G}(w) = \frac{2\langle n, w\rangle}{\langle n, w\rangle + \sqrt{\alpha^2 + (1-\alpha^2)\langle n, w\rangle^2}},
\end{equation}
again with the same roughness parameter $\alpha$.
For $F$ we use Schlick's Fresnel term~\cite{schlick1994inexpensive}, which reads
\begin{equation}
    F(\view, h) = F_0 + (1-F_0)(1-\langle\view, h\rangle)^5,
\end{equation}
with the characteristic specular reflectance $F_0\in[0, 1]^3$.

We combine the specular part with a Lambertian diffuse term, which we diminish by $1-F$. The additional factor accounts for the fact that light reflected at the surface is not available for (diffuse) subsurface scattering. Hence, the full BRDF reads
\begin{equation}
     f(\light, \view) = (1- F(\view, h))\frac{\rho_d}{\pi} +  f_\mathrm{spec}(\light, \view).
\end{equation}

The parameters predicted by the neural network are the roughness parameter $r\in[0,1]$ from which we compute $\alpha = r^2$, as well as the diffuse albedo $\rho_d\in[0, 1]^3$ and the characteristic specular reflectance $F_0\in[0, 1]^3$. We use a sigmoid output nonlinearity for all of them to ensure the correct range. Note that we found the convergence to be more stable if we scale the input to the roughness nonlinearity by 0.5.

\paragraph{Fresnel Microfacet BRDF Model}
The Fresnel microfacet BRDF combines an extended microfacet specular term with a generalized radiometric body reflection \cite{ichikawa2023fresnel}. We refer to the paper for an in-depth discussion of the method. One change in the approach for the specular term compared to the Torrance-Sparrow model is that a generalized normal distribution is used for $D$, which involves estimating a corresponding normalization constant during training. While we were able to predict the rest of the parameters in a spatially varying manner, we were unable to do so for the parameters of this generalized normal distribution since the estimation of the normalization makes the computation untraceable. Therefore, we estimate a single set of parameters for the distribution, while all the other parameters are estimated dependent on the position $\point$. We want to point out again that for the synthetic data, this makes the estimation less complex since the material is uniform over the mesh.

\paragraph{Disney BRDF}
The Disney BRDF extends the microfacet model by several effects, and we refer to the paper for a detailed derivation and discussion \cite{burley2012physically}. The only adjustment we do compared to the full model is to fix the \emph{anisotropic} parameter equal to 0, effectively making the BRDF isotropic. Since all of the other models are isotropic, this enables a meaningful comparison between the approaches. Note that we found the convergence to be more stable if we scale the input to the roughness nonlinearity by 0.5.

\subsection{Softplus Scaling}
\label{sec:supp:softplusScaling}

While tuning the models, we occasionally observed runs for the additive purely neural models, where the training converged to a wrong local minimum. We found that scaling the softplus output of the specular MLP by a factor of 0.5 can stabilize the training in those cases. See also \cref{sec:supp:failure_cases} for a visualization of the failure case.
\section{Training and Evaluation}
\label{sec:supp:training_and_eval}

This section contains additional details on the training and the evaluation. \cref{sec:supp_loss_formulation} sheds more light on the gamma-corrected loss formulation. \cref{sec:supp:pre_proc_data} contains insight into the preprocessing of the DiLiGenT-MV dataset. \cref{sec:supp_lbo_params} presents the encoding parameters used for the training. \cref{sec:supp:regularizers_enhanced} gives details on the regularizers employed for the enhancement of the additional split. The excluded values for the computation of the RMSE are discussed in \cref{sec:supp:rmse}. Finally, \cref{sec:supp:mc_integration} presents details on the Monte Carlo integration for the experiments on the energy conservation.

\subsection{Loss Formulation}
\label{sec:supp_loss_formulation}

To avoid a dominant influence of the bright regions on the loss, we use a gamma mapping to transform the RGB values from linear to sRGB space, as described in
\iftoggle{arxiv}{\cref{eq:loss}}{Eq. (10)}
the main paper. We repeat the loss formulation here for convenience.
\begin{equation}
    \mathcal{L} = \frac{1}{N}\sum_{i=1}^N (\gamma(L_o(x, v)) - \gamma(L_{GT}(x, v))^2
\end{equation}

We use the following standard formula for the gamma mapping $g:[0,1]\rightarrow[0,1],~c_{\mathrm{lin}}\mapsto c_{\mathrm{sRGB}}$ \cite{akenine2019realTimeRendering}:
\begin{equation}
    g(c_{\mathrm{lin}}) = \begin{cases}
        \frac{323}{25}~c_{\mathrm{lin}} & \mathrm{if}~c_{\mathrm{lin}}\leq0.0031308 \\
        \frac{211}{200}~c_{\mathrm{lin}}^\frac{5}{12} - \frac{11}{200} & \mathrm{else }  
    \end{cases}
\end{equation}

\subsection{Pre-Processing of the Data}
\label{sec:supp:pre_proc_data}

The triangle meshes supplied with the DiLiGenT-MV dataset \cite{Li2020DiLiGentMVDataset} have an unreasonably high number of vertices, which causes the computation of the LBO eigenfunctions to take very long. Therefore, we reduce the number of vertices from roughly $10^6$ to about $10^5$. To stay consistent with the simplified mesh, we compute the normals on the mesh rather than using the normal maps included in the dataset. To speed up the training process, we pre-compute and store the LBO eigenfunctions as well as the ray-mesh intersections and the shadow rays.

\subsection{Encoding Parameters}
\label{sec:supp_lbo_params}

For the eigenfunctions of the LBO we use 6 blocks between the $1^\mathrm{st}$ and the $512^\mathrm{th}$ eigenfunction. We use 64 eigenfunctions for the first block and follow up with 6 evenly spaced blocks of 16 eigenfunctions. For the positional encoding of the angles, we use 3 encoding frequencies.

\subsection{Regularizers for the Enhanced Splitting}
\label{sec:supp:regularizers_enhanced}

While the enhancement described in
\iftoggle{arxiv}{\cref{sec:enhancingAddSplit}}{Sec. 4.4}
in the main text shows improvements for both additive architectures, it also introduces additional ambiguity. The weight $\xi(x, l, v)$ makes the diffuse summand dependent on the directions, and therefore, this term can now potentially represent specular behavior as well. To separate the phenomena, we introduce two additional regularizers. 

First, we employ an L1 loss between the raw diffuse part (without the additional weight $\xi$) and the ground truth. Again, we use the gamma mapping described in \cref{sec:supp_loss_formulation}.
\begin{equation}
    \mathcal{L}_\mathrm{reg,diff} = \frac{1}{N}\sum_{i=1}^N \|\gamma(f_\text{d}(x)) - \gamma(L_{GT}(x, v))\|_1
\end{equation}
The idea is to encourage the model to represent as much of the appearance as possible by the diffuse term. This will be limited to the non-view-dependent parts of the appearance automatically since the raw diffuse part $f_\text{d}(x)$ is not view-dependent.

Second, we employ an L1 regularizer on the specular part.
\begin{equation}
    \mathcal{L}_\mathrm{reg,spec}=\frac{1}{N}\sum_{i=1}^N \|f_\text{s}(x, l, v)\|_1
\end{equation}
The idea for this term is to encourage the model to represent only those components of the appearance that are actually view-dependent by the specular part. We use a weight of $5\cdot 10^{-4}$ for both terms before adding them to the total loss.

Interestingly, the regularizers do not significantly change the reconstruction quality of the enhanced split; the quantitative evaluation with and without them is almost identical. However, they force a more reasonable split between the diffuse and specular parts. Without them, the model tends to predict extreme mixing colors on the different parts, that combined yield the correct color. The regularizers ensure that realistic results for the diffuse albedo and the specularities are obtained.

\subsection{BRDF-Spcae Metric RMSE$^{\sqrt[3]{}}$}
\label{sec:supp:rmse}

For the semi-synthetic dataset, we utilize the availability of ground truth BRDF data to report a BRDF-space metric on the reconstruction quality. We follow the analysis of Lavoué \etal who have 
investigated a wide range of metrics and analyzed the correlation with reconstruction quality perceived by humans \cite{Lavoue21PerceptualQualityOfBRDFApproximations}. Their results show that applying the cubic root to the BRDF values before computing a standard \emph{root mean squared error} (RMSE) between the reconstruction and the ground truth yields a metric that correlates well with human perception. We refer to this metric as RMSE$^{\sqrt[3]{}}$.

Another aspect that they found helpful to increase the correlation is to discard BRDF values for grazing angles above $80^{\circ}$, which we also adopt in our analysis. A further reason to discard these values is the observation of Burley, who reported anomalies and extrapolation in the MERL data in that range of angles \cite{burley2012physically}, which the semi-synthetic data inevitably inherits. Moreover, we reject values for saturated pixels since in this case, the image value was clipped during the data creation, and therefore the ground truth value is not a reliable reference. Note that BRDF-space metrics can only be reported for the semi-synthetic data, since the DiLiGenT-MV dataset does not contain ground truth BRDF values.

\subsection{Monte Carlo Integration}
\label{sec:supp:mc_integration}

To approximate the integral in the energy conservation
(\iftoggle{arxiv}{\cref{eq:brdf_energy_conservation}}{Eq. (4)} in the main text)
we use Monte Carlo integration with samples from the cosine-weighted hemisphere sampling. The approximation of the integral reads
\begin{equation}\label{eq:mc_approx_raw}
    \int_{\Hemi} \brdf(\point, \light, \view)\cos{\theta_{\view}} \,\mathrm{d}\view \approx \frac{1}{N_{MC}}\sum_{i=1}^{N_{MC}}\frac{\brdf(\point, \light, \view_i)\cos{\theta_{\view_i}}}{p(\view_i)},
\end{equation}
where $\point$ and $\light$ are the position and the light direction for which the integral is evaluated. $\view_i\sim p(\view_i)$ are the sampled view directions, which are drawn according to the cosine-weighted distribution on the hemisphere, which reads
\begin{equation}
    p(\view_i) = \frac{1}{\pi}\cos{\theta_{\view_i}}.
\end{equation}
Hence, the approximation in \cref{eq:mc_approx_raw} can be simplified to
\begin{equation}
    \int_{\Sphere^2} \brdf(\point, \light, \view)\cos{\theta_{\view}} \,\mathrm{d}\view \approx \frac{\pi}{N_{MC}}\sum_{i=1}^{N_{MC}}\brdf(\point, \light, \view_i).
\end{equation}

To ensure convergence, we sample $N_{MC}=20k$ view directions for each randomly sampled point-light pair.
\section{Additional Experimental Results}
\label{sec:supp:add_experimental_results}

In this section, we present several additional results. \cref{sec:supp:quantitative_reciprocity} and \cref{sec:supp:arch_changes_view_light_exps} give further results and details for the experiments on the reciprocity approach and the number of layers for the directions from 
\iftoggle{arxiv}{\cref{sec:analysis_brdf_models}}{Sec.~6.2}
in the main text. In \cref{sec:supp:exps_angle_param}, we analyze the influence of the angle parametrization on the reconstruction quality. \cref{sec:supp:changes_nerfactor} justifies our changes to the NeRFactor architecture, which is the basis of our additive separate architecture.
\cref{sec:supp:failure_cases} gives insight to failure cases for the purely neural additive approaches that we observed occasionally.
While \cref{sec:supp:qualitative_diffuse_specular} presents a qualitative analysis of the diffuse and specular parts of the models based on a split of the reflectance
\cref{sec:supp:spat_var_recon_brdf} uses the albedo maps to analyze how well the models can represent the spatially uniform BRDFs of the semi-synthetic dataset.
Finally, \cref{sec:supp:additional_experiments_comp} provides additional qualitative and quantitative results for both datasets, comparing the various neural BRDF approaches.

\subsection{Quantitative Results for the Reciprocity Approaches}
\label{sec:supp:quantitative_reciprocity}

\begin{table*}[t]  %
\centering  %
  \footnotesize
    \begin{tabular}{l||ccccc|cccc}
& \multicolumn{5}{c|}{\merlc} & \multicolumn{4}{c}{\diligentc} \\

 & $\Delta$RMSE$^{\sqrt[3]{}}$ & $\Delta$\psnr & $\Delta$\dssim & $\Delta$\lpips & $\Delta$\flip & $\Delta$\psnr & $\Delta$\dssim & $\Delta$\lpips & $\Delta$\flip \\ \hline \hline
Single MLP (rnd. in-swap) 		%
&   \textcolor{greenValid}{-0.08} &   \textcolor{greenValid}{+0.04} &   \textcolor{red}{+0.02} &   \textcolor{red}{+0.02} &   \textcolor{red}{+0.02} &  \textcolor{greenValid}{+0.06} &  \textcolor{greenValid}{-0.01} &  \textcolor{greenValid}{-0.01}  &  \textcolor{greenValid}{-0.02} \\
Single MLP (ours) 		%
&   \textcolor{greenValid}{-0.08} &   \textcolor{red}{-0.10} &   \textcolor{red}{+0.02} &   \textcolor{red}{+0.01} &   \textcolor{red}{+0.03} &  \textcolor{red}{-0.30} &  \textcolor{red}{+0.03} &  \textcolor{red}{+0.06}  &  \textcolor{red}{+0.07} \\
Additive Separate (rnd. in-swap) 		%
&   \textcolor{greenValid}{-0.07} &   \textcolor{greenValid}{+0.04} &   \textcolor{red}{+0.02} &   \textcolor{red}{+0.01} &   \textcolor{red}{+0.02} &  0.00 &  0.00 &  0.00  &  \textcolor{red}{+0.01} \\
Additive Separate (ours) 		%
&   \textcolor{greenValid}{-0.07} &   \textcolor{red}{-0.05} &   \textcolor{red}{+0.02} &   \textcolor{red}{+0.02} &   \textcolor{red}{+0.04} &  \textcolor{red}{-0.02} &  0.00 &  \textcolor{red}{+0.01}  &  \textcolor{red}{+0.01} \\
Additive Shared (rnd. in-swap) 		%
&   \textcolor{greenValid}{-0.05} &   \textcolor{red}{-0.10} &   \textcolor{red}{+0.04} &   \textcolor{red}{+0.16} &   \textcolor{greenValid}{-0.12} &  \textcolor{red}{-0.15} &  \textcolor{red}{+0.01} &  \textcolor{red}{+0.01}  &  \textcolor{red}{+0.01} \\
Additive Shared (ours) 		%
&   \textcolor{greenValid}{-0.01} &   \textcolor{red}{-0.21} &   \textcolor{red}{+0.06} &   \textcolor{red}{+0.24} &   \textcolor{greenValid}{-0.05} &  \textcolor{red}{-0.13} &  0.00 &  0.00  &  \textcolor{red}{+0.01} \\

\end{tabular}
  
\caption{
Effect of the reciprocity strategies on the reconstruction quality. Shown are the differences to the results in
\iftoggle{arxiv}{\cref{tab:quantitative}}{Tab.~1} in the main text, which reports the reconstruction quality of the architectures without any reciprocity strategy. With a few exceptions, we see that the influence of both strategies on the results is fairly marginal, with a tendency for slightly worse results. RMSE$^{\sqrt[3]{}}$, DSSIM, LPIPS and \FLIP are scaled by 100.
}
\label{tab:changes_quantitative_reci}
\end{table*}

As shown in 
\iftoggle{arxiv}{\cref{sec:analysis_brdf_models}}{Sec.~6.2}
in the main text, the random input swap of view and light direction applied during training, as described in LitNerf \cite{Sarkar23LitNerf}, can reduce the RMSE of the reciprocity constraint. However, this strategy does not provide a guarantee that the reciprocity is actually fulfilled. In contrast, our input mapping, as proposed in
\iftoggle{arxiv}{\cref{sec:reciMapping}}{Sec.~4.3},
ensures that the reciprocity constraint is exactly fulfilled by construction.
In \cref{tab:changes_quantitative_reci}, we analyze the effect of both strategies on the reconstruction quality by comparing against the results without any reciprocity strategy. Overall, we observe only a minimal difference with a minor tendency for slightly worse results as a trade-off for the fulfilled reciprocity.
One exception is the single MLP architecture, where, for the real-world examples, the random input swap seems to have a more noticeable advantage over our strategy.

\subsection{Architecture Changes Experiments: Number of Layers for View/Light Direction}
\label{sec:supp:arch_changes_view_light_exps}

In
\iftoggle{arxiv}{\cref{sec:analysis_brdf_models}}{Sec.~6.2}
in the main text, we analyzed the influence of the number of layers (NOL) for the directions on the reconstruction quality. In the following, we provide detailed architectural changes for the different models.

\begin{itemize}
    \item For the single MLP architecture, we feed the directions in at layer 4 instead of 1, decreasing the NOL for the directions from 6 to 3.
    \item For the additive separate architecture, we reduce the NOL of the specular MLP from 4 to 2. Also, we remove the input skip.
    \item For the additive shared architecture, we reduce the NOL of the shared MLP from 5 to 3 while increasing the NOL of the diffuse and specular MLPs (and therefore NOL for the directions) by 2, respectively.
\end{itemize}

\subsection{Influence of the Angle Parametrization}
\label{sec:supp:exps_angle_param}

\begin{table*}[t]  %
\centering  %
  \footnotesize
    \begin{tabular}{l||ccccc|cccc}
& \multicolumn{5}{c|}{\merlc} & \multicolumn{4}{c}{\diligentc} \\

 & $\Delta$RMSE$^{\sqrt[3]{}}$ & $\Delta$\psnr & $\Delta$\dssim & $\Delta$\lpips & $\Delta$\flip & $\Delta$\psnr & $\Delta$\dssim & $\Delta$\lpips & $\Delta$\flip \\ \hline \hline
Single MLP (view-light) 		%
&   \textcolor{red}{+0.11} &   \textcolor{red}{-0.86} &   \textcolor{red}{+0.05} &   \textcolor{red}{+0.14} &   \textcolor{red}{+0.15} &  \textcolor{red}{-0.10} &  \textcolor{red}{+0.01} &  \textcolor{red}{+0.01}  &  \textcolor{red}{+0.01} \\
Additive Separate (view-light) 		%
&   \textcolor{red}{+0.11} &   \textcolor{red}{-0.85} &   \textcolor{red}{+0.04} &   \textcolor{red}{+0.12} &   \textcolor{red}{+0.16} &  \textcolor{red}{-0.03} &  \textcolor{red}{+0.01} &  \textcolor{greenValid}{-0.02}  &  \textcolor{red}{+0.01} \\
Additive Shared (view-light) 		%
&   \textcolor{red}{+0.17} &   \textcolor{red}{-1.01} &   \textcolor{red}{+0.05} &   \textcolor{red}{+0.09} &   \textcolor{red}{+0.07} &  \textcolor{red}{-0.07} &  \textcolor{red}{+0.01} &  \textcolor{greenValid}{-0.01}  &  0.00 \\

\end{tabular}
  
\caption{
Effect of using the view-light angles instead of the Rusinkiewicz angles as a parametrization of the directions. Shown are the differences to the results in
\iftoggle{arxiv}{\cref{tab:quantitative}}{Tab.~1} in the main text, which reports the reconstruction quality of the architectures with the Rusinkiewicz angles. The results show that overall, the view-light parametrization reduces the reduction quality compared to the Rusinkiewicz parametrization. The difference is more significant for the semi-synthetic MERL-based dataset. The reason is that this data contains more highly specular materials with complex reflective patterns. The alignment of the specular peaks with the coordinate axes provided by the Rusinkiewicz angles seems to provide a significant benefit in this case. RMSE$^{\sqrt[3]{}}$, DSSIM, LPIPS and \FLIP are scaled by 100.
}
\label{tab:changes_quantitative_view_light}
\end{table*}

Following previous work \cite{Zhang2021NeRFactor, sztrajman2021neural}, we use the Rusinkiewicz angles \cite{rusinkiewicz1998new} to parametrize the directions as inputs for the MLPs; see \cref{sec:supp:angle_definitions} for a review and discussion. As noted by Rusinkiewicz, this parametrization aligns the specular peaks better with the coordinate axes, which benefits learning highly specular materials. The results in \cref{tab:changes_quantitative_view_light} confirm that using the view-light angles as parametrization for purely neural BRDF models reduces the reconstruction quality compared to the Rusinkiewicz angles. Moreover, we see that, indeed, this effect is more prominent for the MERL-based semi-synthetic dataset, which contains more highly specular materials.

\subsection{Changes from the NeRFactor Architecture}
\label{sec:supp:changes_nerfactor}

\begin{table*}[t]  %
\centering  %
  \footnotesize
    \begin{tabular}{l||ccccc|cccc}
& \multicolumn{5}{c|}{\merlc} & \multicolumn{4}{c}{\diligentc} \\

 & $\Delta$RMSE$^{\sqrt[3]{}}$ & $\Delta$\psnr & $\Delta$\dssim & $\Delta$\lpips & $\Delta$\flip & $\Delta$\psnr & $\Delta$\dssim & $\Delta$\lpips & $\Delta$\flip \\ \hline \hline
Additive Separate (albedo clamp.) 		%
&   \textcolor{red}{+0.90} &   \textcolor{red}{-2.25} &   \textcolor{red}{+0.54} &   \textcolor{red}{+0.84} &   \textcolor{red}{+0.90} &  \textcolor{greenValid}{+0.05} &  0.00 &  \textcolor{greenValid}{-0.02}  &  \textcolor{red}{+0.01} \\
Additive Separate (scalar spec.) 		%
&   \textcolor{red}{+0.41} &   \textcolor{red}{-3.32} &   \textcolor{red}{+0.10} &   \textcolor{red}{+0.34} &   \textcolor{red}{+0.74} &  \textcolor{red}{-0.54} &  \textcolor{red}{+0.01} &  \textcolor{red}{+0.04}  &  \textcolor{red}{+0.32} \\

\end{tabular}
  
\caption{
Effect of using the albedo clamping and a scalar specular term proposed in NeRFactor  \cite{Zhang2021NeRFactor} for the \emph{additive separate} architecture. Shown are the differences to the results in
\iftoggle{arxiv}{\cref{tab:quantitative}}{Tab.~1} in the main text, which reports the reconstruction quality of the additive separate model with neither of the two. We see, that the albedo clamping reduces the reconstruction quality, in particular for the MERL-based data. The clamping prohibits the model from predicting an albedo close to zero, which is necessary, however, for the metallic materials contained in this dataset. See also \cref{sec:supp:qualitative_diffuse_specular} and in particular in \cref{fig:supp_diff_spec_synth_2}. Similarly, the scalar specular term reduces the reconstruction quality for both datasets. We find, that for certain materials, an RGB specularity is necessary for a faithful reconstruction, see \cref{fig:supp:nerfactor_cow_scalar_spec}. RMSE$^{\sqrt[3]{}}$, DSSIM, LPIPS and \FLIP are scaled by 100.
}
\label{tab:supp:changes_quantitative_nerfactor}
\end{table*}
\begin{figure}
    \centering
    
    \begin{tabular}{cc}
    
     \includegraphics[width=0.38\linewidth]{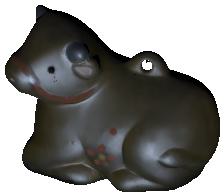}
    &
    \includegraphics[width=0.38\linewidth]{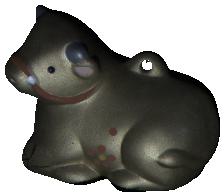} \\
    Scalar Specularity 
    &
    GT
    \end{tabular}    
    
\caption{
    Qualitative BRDF reconstruction for the cow object from the  DiLiGenT-MV dataset \cite{Li2020DiLiGentMVDataset} for the additive separate architecture with a scalar specular term (as suggested in NeRFactor \cite{Zhang2021NeRFactor}). The results show, that a scalar specular term is unable to reconstruct the reflectance of this object and creates a spurious glow. This indicates that for some materials, a specular term with 3 channels is necessary to yield high-quality reconstructions.
}
\label{fig:supp:nerfactor_cow_scalar_spec}
\end{figure}

While our additive separate architecture is based on NeRFactor \cite{Zhang2021NeRFactor}, we made two changes, which improved the results for our data significantly. In this section, we present the comparison to justify these adjustments.

As a first change, we remove the albedo clamping. The original work empirically constrains the diffuse reflection (\ie the albedo) to $[0.03, 0.8]$. Since we work in linear space, we transform these values from sRGB to linear space, which yields the range $[0.0023, 0.6038]$. The results in \cref{tab:supp:changes_quantitative_nerfactor} show that we obtain significantly worse results for the semi-synthetic MERL-based data. The reason lies in the metallic materials contained in this dataset. Metals show almost no subsurface scattering due to the free electrons \cite{akenine2019realTimeRendering}. As demonstrated in \cref{sec:supp:qualitative_diffuse_specular} and visible in particular in \cref{fig:supp_diff_spec_synth_2}, all models replicate this behavior and show almost no diffuse contribution. However, the lower bound on the diffuse part imposed by the albedo clamping prohibits a negligible contribution of the albedo, and we observe a dark gray base color for the diffuse renderings. This leads to the decrease in construction quality. While we see slight improvements with the albedo clamping for non-metal materials and the real-world data, we still decided to remove it due to the significant performance decrease for metallic materials.

As a second change, we use an RGB specular term instead of a scalar one. While the original work assumes, that all color information can be handeled by the albedo network, we find, that for certain materials, a colored specular part is necessary for good reconstructions. The most extreme example we observed is the cow object from the DiLiGenT-MV dataset \cite{Li2020DiLiGentMVDataset} as shown in \cref{fig:supp:nerfactor_cow_scalar_spec}. We can clearly see, that an approach with a scalar specular term yields an inaccurate glow effect that is not observed for models with an RGB specular term (\cf
the rendering for the additive separate architecture in
\iftoggle{arxiv}{\cref{fig:evaluation_renderings}}{Fig.~3}
in the main text and \cref{fig:supp_diff_spec_real} in the appendix).
While the effect is less prominent for other materials, \cref{tab:supp:changes_quantitative_nerfactor} confirms that a scalar specular term instead of an RGB one yields systematically worse results for both datasets. Recall that we observe a similar effect on the cow for the FMBRDF model \cite{ichikawa2023fresnel}, which also employs a scalar specular term; \cf
\iftoggle{arxiv}{\cref{sec:comparison_brdf_models}}{Sec.~6.1}
in the main text.

\subsection{Failure Cases}
\label{sec:supp:failure_cases}

\begin{figure}
    \centering
    
    \begin{tabular}{ccc}
    
     \includegraphics[width=0.31\linewidth]{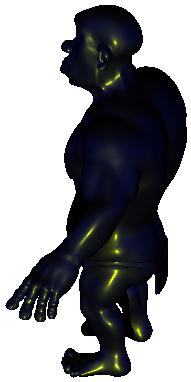}
    &
    \includegraphics[width=0.31\linewidth]{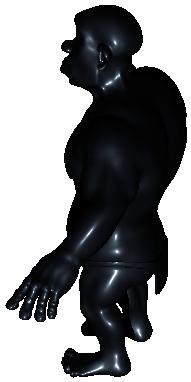}
    &
    \includegraphics[width=0.31\linewidth]{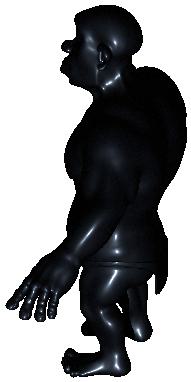} \\
    No Scaling
    &
    Output Scaling 0.5
    &
    GT
    \end{tabular}    
    
\caption{
    Failure case for the additive shared architecture. For additive purely neural methods with softplus output nonlinearity, we observed occasional failures like this for very shiny materials. As described in \cref{sec:supp:softplusScaling}, scaling the output of the softplus function solves this issue.
}
\label{fig:supp:failure_softplus}
\end{figure}

Although all methods show quite stable convergence with the chosen parameters, we observed occasional issues with very shiny materials for the additive purely neural models that employ a softplus output nonlinearity. \cref{fig:supp:failure_softplus} shows a failure example for the additive shared architecture. As described in \cref{sec:supp:softplusScaling}, scaling the output of the softplus function solves this issue.

\subsection{Qualitative Results Diffuse and Specular Split}
\label{sec:supp:qualitative_diffuse_specular}

\begin{figure*}[t]  %
  \centering  %
  \footnotesize
  \newcommand{\mywidthc}{0.02\textwidth}  %
  \newcommand{\mywidthx}{0.10\textwidth}  %
  \newcommand{\mywidthw}{0.008\textwidth}  %
  \newcommand{\myheightx}{0.15\textwidth}  %
  \newcommand{\mywidtht}{0.04\textwidth}  %
  \newcolumntype{C}{ >{\centering\arraybackslash} m{\mywidthc} } %
  \newcolumntype{X}{ >{\centering\arraybackslash} m{\mywidthx} } %
  \newcolumntype{W}{ >{\centering\arraybackslash} m{\mywidthw} } %
  \newcolumntype{T}{ >{\centering\arraybackslash} m{\mywidtht} } %

  \newcommand{\heightcolorbar}{0.10\textwidth}  %
  \newcommand{\xposOne}{-0.95}
  \newcommand{\yposOne}{0.35}
  \newcommand{\xposTwo}{0.35}
  \newcommand{\yposTwo}{0.8}

  \newcommand{\fontsizePSNR}{\ssmall}
  
  \setlength\tabcolsep{0pt} %

  \setlength{\extrarowheight}{1.25pt}
  
  \def\arraystretch{0.8} %
  \begin{tabular}{TTXXXXXXXXWX}

\multirow{2}{*}{\rotatebox{90}{\parbox{3cm}{\centering \diligentc \\ cow}}} 
&
\rotatebox{90}{\centering \emph{diffuse}} 
&
\includegraphics[width=\mywidthx, height=\myheightx, keepaspectratio]{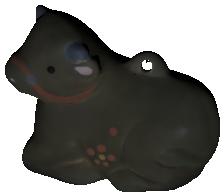}
&
\includegraphics[width=\mywidthx, height=\myheightx, keepaspectratio]{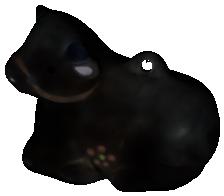}
&
\includegraphics[width=\mywidthx, height=\myheightx, keepaspectratio]{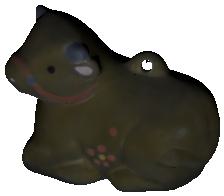}
&
\includegraphics[width=\mywidthx, height=\myheightx, keepaspectratio]{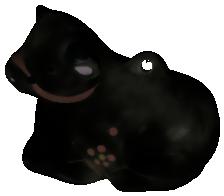}
&
\includegraphics[width=\mywidthx, height=\myheightx, keepaspectratio]{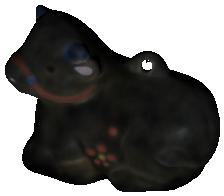}
&
\includegraphics[width=\mywidthx, height=\myheightx, keepaspectratio]{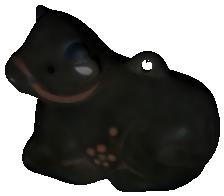}
&
\includegraphics[width=\mywidthx, height=\myheightx, keepaspectratio]{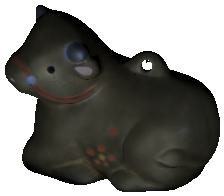}
&
\includegraphics[width=\mywidthx, height=\myheightx, keepaspectratio]{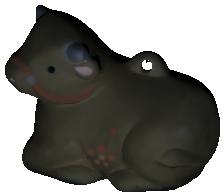}
&
&
\\
&
\rotatebox{90}{\centering \emph{specular}} 
&
\includegraphics[width=\mywidthx, height=\myheightx, keepaspectratio]{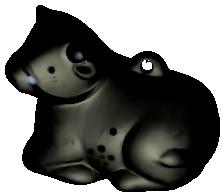}
&
\includegraphics[width=\mywidthx, height=\myheightx, keepaspectratio]{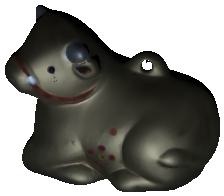}
&
\includegraphics[width=\mywidthx, height=\myheightx, keepaspectratio]{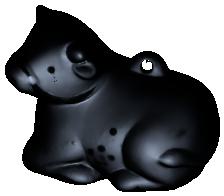}
&
\includegraphics[width=\mywidthx, height=\myheightx, keepaspectratio]{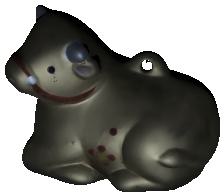}
&
\includegraphics[width=\mywidthx, height=\myheightx, keepaspectratio]{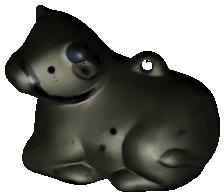}
&
\includegraphics[width=\mywidthx, height=\myheightx, keepaspectratio]{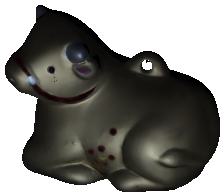}
&
\includegraphics[width=\mywidthx, height=\myheightx, keepaspectratio]{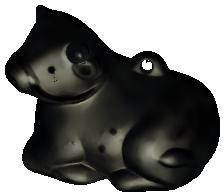}
&
\includegraphics[width=\mywidthx, height=\myheightx, keepaspectratio]{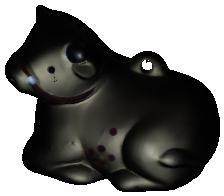}
&
&
\\
&
\rotatebox{90}{\centering \emph{added}} 
&
\includegraphics[width=\mywidthx, height=\myheightx, keepaspectratio]{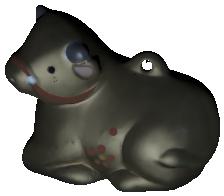}
&
\includegraphics[width=\mywidthx, height=\myheightx, keepaspectratio]{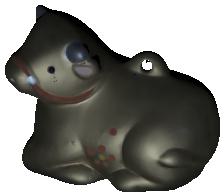}
&
\includegraphics[width=\mywidthx, height=\myheightx, keepaspectratio]{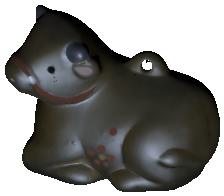}
&
\includegraphics[width=\mywidthx, height=\myheightx, keepaspectratio]{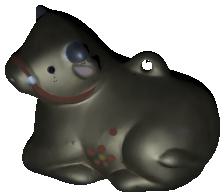}
&
\includegraphics[width=\mywidthx, height=\myheightx, keepaspectratio]{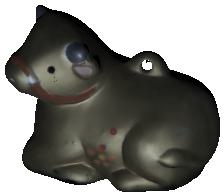}
&
\includegraphics[width=\mywidthx, height=\myheightx, keepaspectratio]{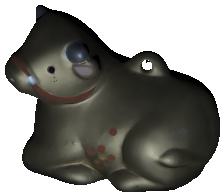}
&
\includegraphics[width=\mywidthx, height=\myheightx, keepaspectratio]{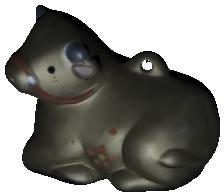}
&
\includegraphics[width=\mywidthx, height=\myheightx, keepaspectratio]{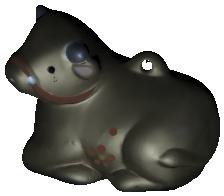}
&
&
\includegraphics[width=\mywidthx, height=\myheightx, keepaspectratio]{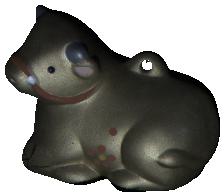}\\
\hline\hline\\
\multirow{2}{*}{\rotatebox{90}{\parbox{3cm}{\centering \diligentc \\ pot2}}} 
&
\rotatebox{90}{\centering \emph{diffuse}} 
&
\includegraphics[width=\mywidthx, height=\myheightx, keepaspectratio]{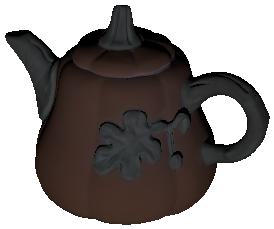}
&
\includegraphics[width=\mywidthx, height=\myheightx, keepaspectratio]{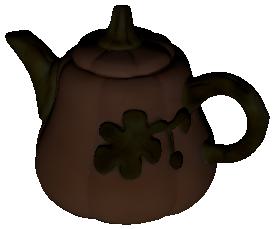}
&
\includegraphics[width=\mywidthx, height=\myheightx, keepaspectratio]{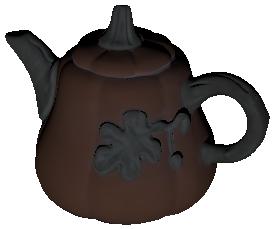}
&
\includegraphics[width=\mywidthx, height=\myheightx, keepaspectratio]{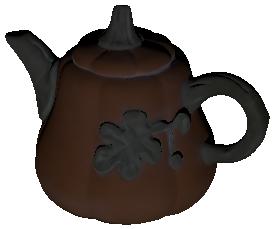}
&
\includegraphics[width=\mywidthx, height=\myheightx, keepaspectratio]{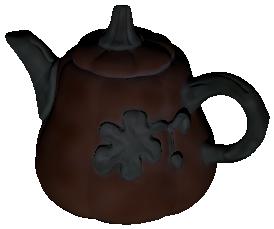}
&
\includegraphics[width=\mywidthx, height=\myheightx, keepaspectratio]{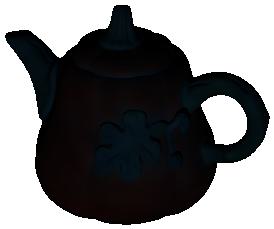}
&
\includegraphics[width=\mywidthx, height=\myheightx, keepaspectratio]{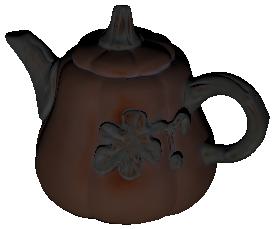}
&
\includegraphics[width=\mywidthx, height=\myheightx, keepaspectratio]{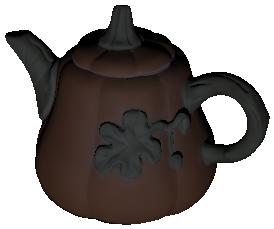}
&
&
\\
&
\rotatebox{90}{\centering \emph{specular}} 
&
\includegraphics[width=\mywidthx, height=\myheightx, keepaspectratio]{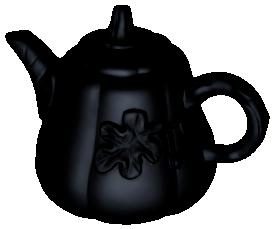}
&
\includegraphics[width=\mywidthx, height=\myheightx, keepaspectratio]{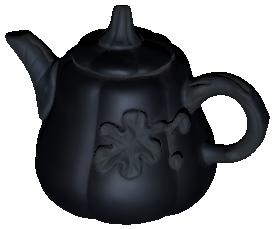}
&
\includegraphics[width=\mywidthx, height=\myheightx, keepaspectratio]{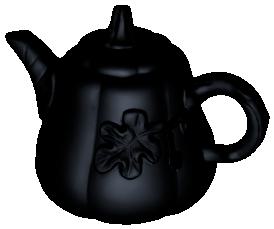}
&
\includegraphics[width=\mywidthx, height=\myheightx, keepaspectratio]{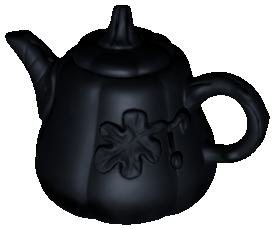}
&
\includegraphics[width=\mywidthx, height=\myheightx, keepaspectratio]{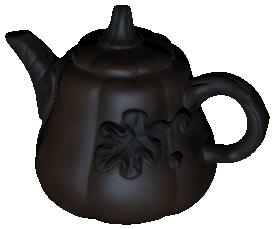}
&
\includegraphics[width=\mywidthx, height=\myheightx, keepaspectratio]{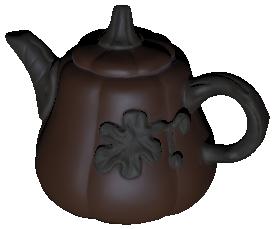}
&
\includegraphics[width=\mywidthx, height=\myheightx, keepaspectratio]{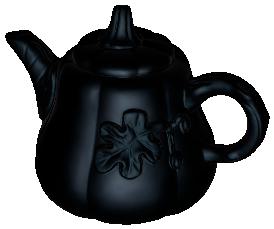}
&
\includegraphics[width=\mywidthx, height=\myheightx, keepaspectratio]{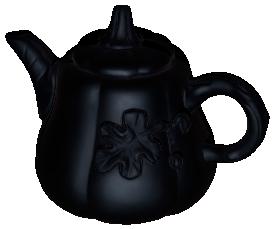}
&
&
\\
&
\rotatebox{90}{\centering \emph{added}} 
&
\includegraphics[width=\mywidthx, height=\myheightx, keepaspectratio]{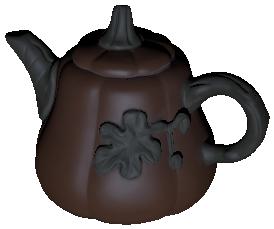}
&
\includegraphics[width=\mywidthx, height=\myheightx, keepaspectratio]{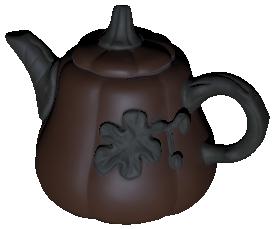}
&
\includegraphics[width=\mywidthx, height=\myheightx, keepaspectratio]{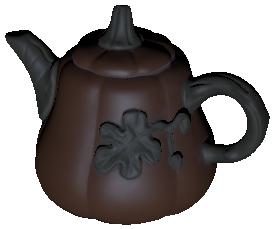}
&
\includegraphics[width=\mywidthx, height=\myheightx, keepaspectratio]{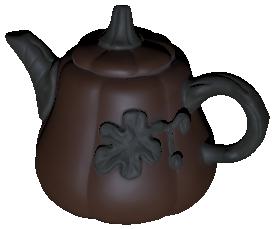}
&
\includegraphics[width=\mywidthx, height=\myheightx, keepaspectratio]{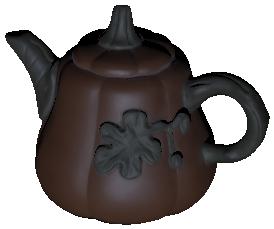}
&
\includegraphics[width=\mywidthx, height=\myheightx, keepaspectratio]{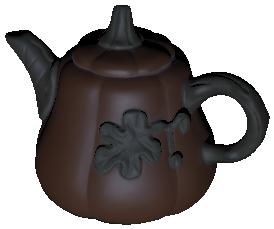}
&
\includegraphics[width=\mywidthx, height=\myheightx, keepaspectratio]{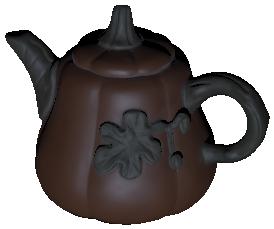}
&
\includegraphics[width=\mywidthx, height=\myheightx, keepaspectratio]{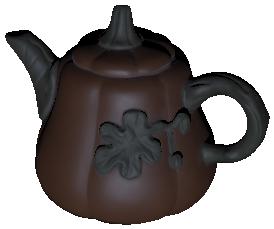}
&
&
\includegraphics[width=\mywidthx, height=\myheightx, keepaspectratio]{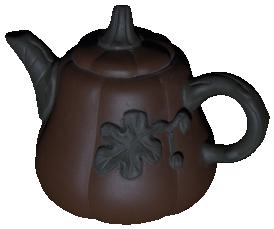}\\
\hline\hline\\
\multirow{2}{*}{\rotatebox{90}{\parbox{3cm}{\centering \diligentc \\ buddha}}} 
&
\rotatebox{90}{\centering \emph{diffuse}} 
&
\includegraphics[width=\mywidthx, height=\myheightx, keepaspectratio]{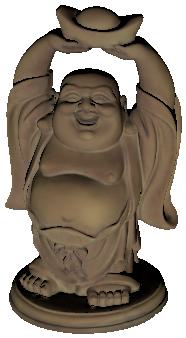}
&
\includegraphics[width=\mywidthx, height=\myheightx, keepaspectratio]{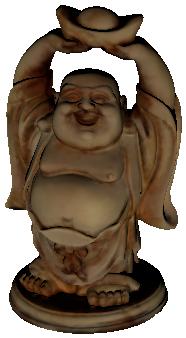}
&
\includegraphics[width=\mywidthx, height=\myheightx, keepaspectratio]{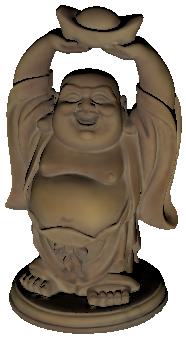}
&
\includegraphics[width=\mywidthx, height=\myheightx, keepaspectratio]{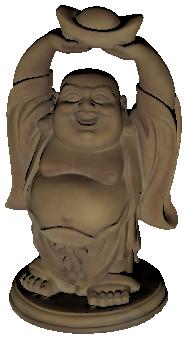}
&
\includegraphics[width=\mywidthx, height=\myheightx, keepaspectratio]{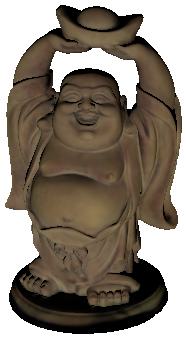}
&
\includegraphics[width=\mywidthx, height=\myheightx, keepaspectratio]{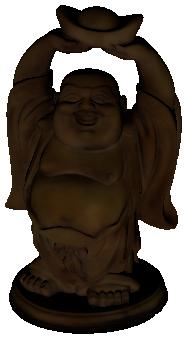}
&
\includegraphics[width=\mywidthx, height=\myheightx, keepaspectratio]{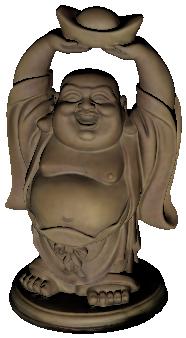}
&
\includegraphics[width=\mywidthx, height=\myheightx, keepaspectratio]{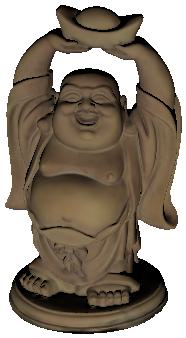}
&
&
\\
&
\rotatebox{90}{\centering \emph{specular}} 
&
\includegraphics[width=\mywidthx, height=\myheightx, keepaspectratio]{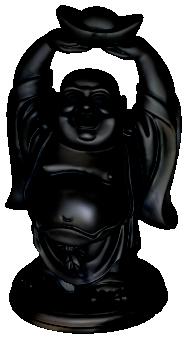}
&
\includegraphics[width=\mywidthx, height=\myheightx, keepaspectratio]{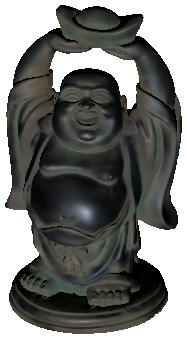}
&
\includegraphics[width=\mywidthx, height=\myheightx, keepaspectratio]{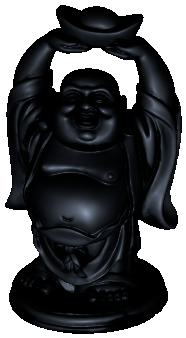}
&
\includegraphics[width=\mywidthx, height=\myheightx, keepaspectratio]{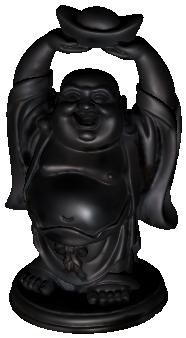}
&
\includegraphics[width=\mywidthx, height=\myheightx, keepaspectratio]{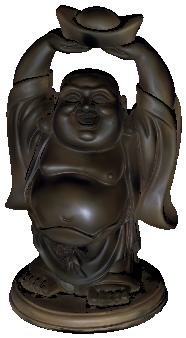}
&
\includegraphics[width=\mywidthx, height=\myheightx, keepaspectratio]{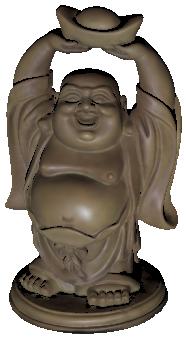}
&
\includegraphics[width=\mywidthx, height=\myheightx, keepaspectratio]{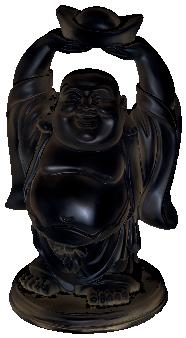}
&
\includegraphics[width=\mywidthx, height=\myheightx, keepaspectratio]{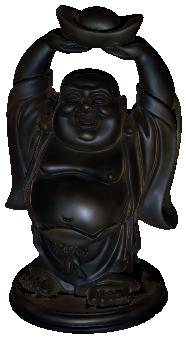}
&
&
\\
&
\rotatebox{90}{\centering \emph{added}} 
&
\includegraphics[width=\mywidthx, height=\myheightx, keepaspectratio]{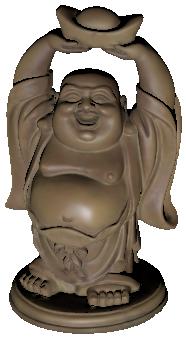}
&
\includegraphics[width=\mywidthx, height=\myheightx, keepaspectratio]{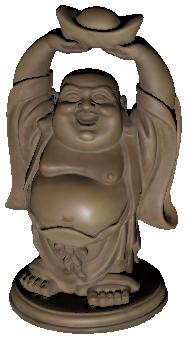}
&
\includegraphics[width=\mywidthx, height=\myheightx, keepaspectratio]{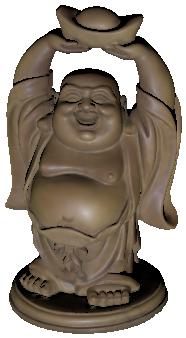}
&
\includegraphics[width=\mywidthx, height=\myheightx, keepaspectratio]{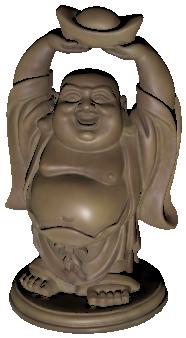}
&
\includegraphics[width=\mywidthx, height=\myheightx, keepaspectratio]{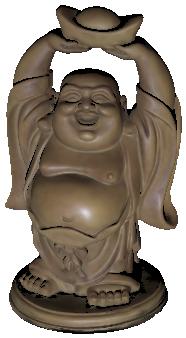}
&
\includegraphics[width=\mywidthx, height=\myheightx, keepaspectratio]{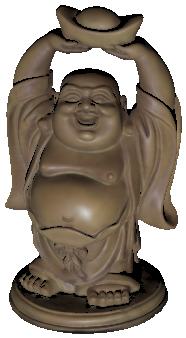}
&
\includegraphics[width=\mywidthx, height=\myheightx, keepaspectratio]{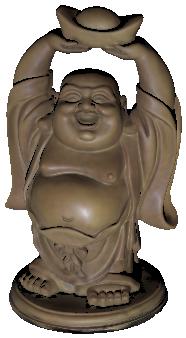}
&
\includegraphics[width=\mywidthx, height=\myheightx, keepaspectratio]{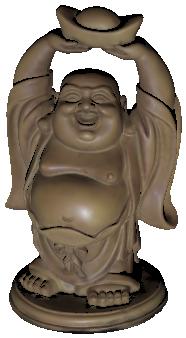}
&
&
\includegraphics[width=\mywidthx, height=\myheightx, keepaspectratio]{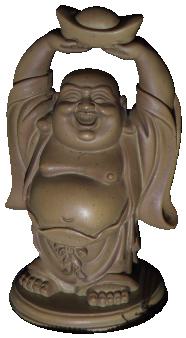}\\
\hline\hline\\[-0.2cm]
&
 & \cellcolor{cellParamBased}\rpc		%
 & \cellcolor{cellParamBased}\tsc		%
 & \cellcolor{cellParamBased}\fmbrdfc		%
 & \cellcolor{cellParamBased}\disneyc		%
 & \cellcolor{celPurelyNeural}Add Sep		%
 & \cellcolor{celPurelyNeural}Add Shared		%
 & \cellcolor{celPurelyNeural}Add Sep (enh.)		%
 & \cellcolor{celPurelyNeural}Add Shared (enh.)		%
& %
 & \gt
  \end{tabular}
\caption{
Renderings of the diffuse and the specular parts separately for all additive models. Note that for the models with the enhanced additive strategy (\emph{enh.}), the diffuse part is already weighted with $\xi$. Also shown are the combined rendering (\emph{added}) and the ground truth image (\emph{GT}). 
All models show a reasonable split into diffuse albedo and specular parts. For the vanilla methods of the purely neural category (Additive Separate and Additive Shared), we observe the tendency to represent more appearance in the specular part, which, as can be seen by the other models, does not seem necessary. Our enhancement for the additive split (\emph{enh.}) as introduced in 
\iftoggle{arxiv}{\cref{sec:enhancingAddSplit}}{Sec.~4.4}
and in particular the regularizers discussed in 
\iftoggle{arxiv}{\cref{sec:enhancingAddSplit}}{Sec.~4.4} and \cref{sec:supp:regularizers_enhanced} seem to improve on the disentanglement of diffuse and specular components.
Note that for the cow object, a colored (RGB) specular component seems necessary to reconstruct its appearance. The FMBRDF model \cite{ichikawa2023fresnel}, which is based on a scalar specular term, shows an unnatural glow. Note that we observed similar behavior for the scalar specular term proposed by NeRFactor \cite{Zhang2021NeRFactor}, as discussed in \cref{sec:supp:changes_nerfactor}.
}
\label{fig:supp_diff_spec_real}
\end{figure*}
\begin{figure*}[t]  %
  \centering  %
  \footnotesize
  \newcommand{\mywidthc}{0.02\textwidth}  %
  \newcommand{\mywidthx}{0.10\textwidth}  %
  \newcommand{\mywidthw}{0.008\textwidth}  %
  \newcommand{\myheightx}{0.15\textwidth}  %
  \newcommand{\mywidtht}{0.04\textwidth}  %
  \newcolumntype{C}{ >{\centering\arraybackslash} m{\mywidthc} } %
  \newcolumntype{X}{ >{\centering\arraybackslash} m{\mywidthx} } %
  \newcolumntype{W}{ >{\centering\arraybackslash} m{\mywidthw} } %
  \newcolumntype{T}{ >{\centering\arraybackslash} m{\mywidtht} } %

  \newcommand{\heightcolorbar}{0.10\textwidth}  %
  \newcommand{\xposOne}{-0.95}
  \newcommand{\yposOne}{0.35}
  \newcommand{\xposTwo}{0.35}
  \newcommand{\yposTwo}{0.8}

  \newcommand{\fontsizePSNR}{\ssmall}
  
  \setlength\tabcolsep{0pt} %

  \setlength{\extrarowheight}{1.25pt}
  
  \def\arraystretch{0.8} %
  \begin{tabular}{TTXXXXXXXXWX}

\multirow{2}{*}{\rotatebox{90}{\parbox{3cm}{\centering \merlc \\ blue acrylic}}} 
&
\rotatebox{90}{\centering \emph{diffuse}} 
&
\includegraphics[width=\mywidthx, height=\myheightx, keepaspectratio]{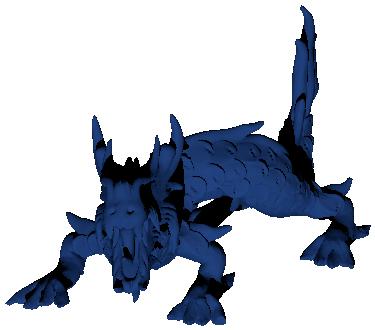}
&
\includegraphics[width=\mywidthx, height=\myheightx, keepaspectratio]{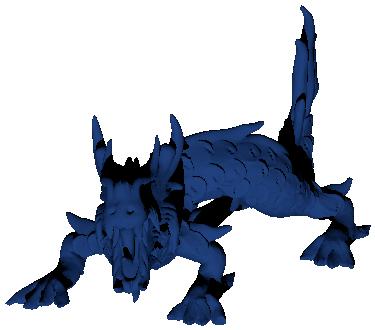}
&
\includegraphics[width=\mywidthx, height=\myheightx, keepaspectratio]{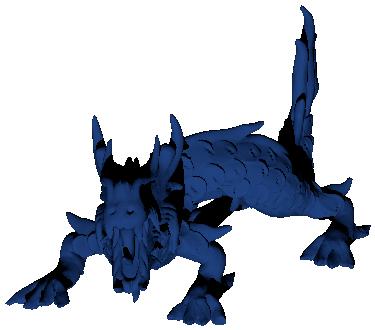}
&
\includegraphics[width=\mywidthx, height=\myheightx, keepaspectratio]{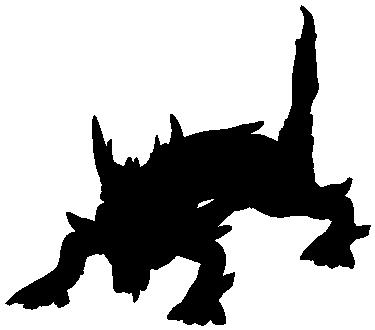}
&
\includegraphics[width=\mywidthx, height=\myheightx, keepaspectratio]{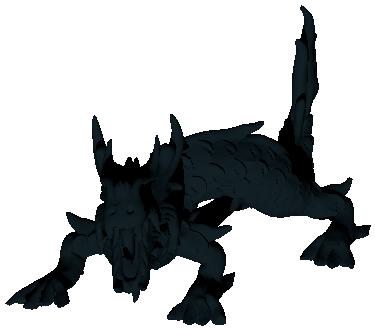}
&
\includegraphics[width=\mywidthx, height=\myheightx, keepaspectratio]{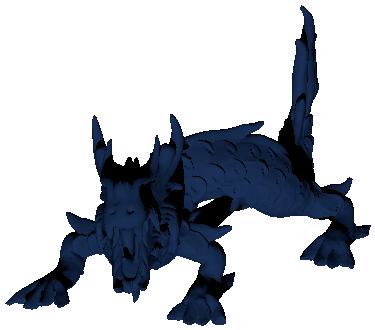}
&
\includegraphics[width=\mywidthx, height=\myheightx, keepaspectratio]{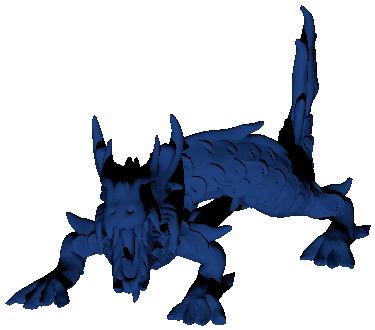}
&
\includegraphics[width=\mywidthx, height=\myheightx, keepaspectratio]{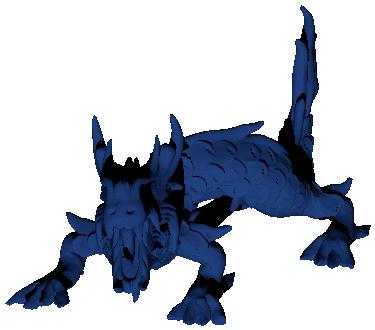}
&
&
\\
&
\rotatebox{90}{\centering \emph{specular}} 
&
\includegraphics[width=\mywidthx, height=\myheightx, keepaspectratio]{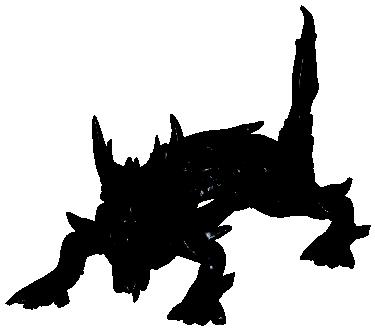}
&
\includegraphics[width=\mywidthx, height=\myheightx, keepaspectratio]{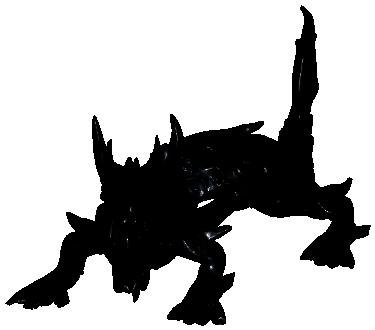}
&
\includegraphics[width=\mywidthx, height=\myheightx, keepaspectratio]{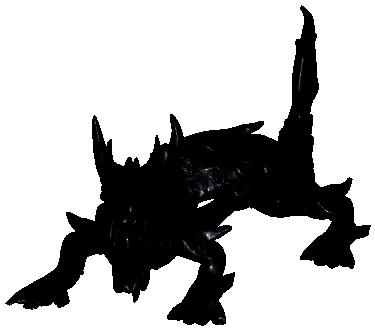}
&
\includegraphics[width=\mywidthx, height=\myheightx, keepaspectratio]{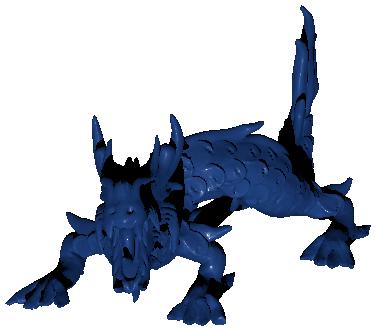}
&
\includegraphics[width=\mywidthx, height=\myheightx, keepaspectratio]{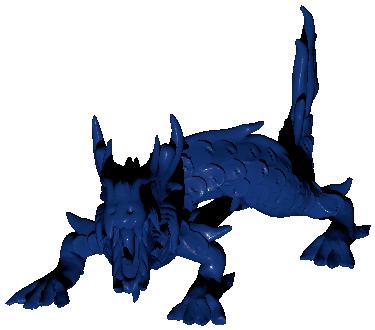}
&
\includegraphics[width=\mywidthx, height=\myheightx, keepaspectratio]{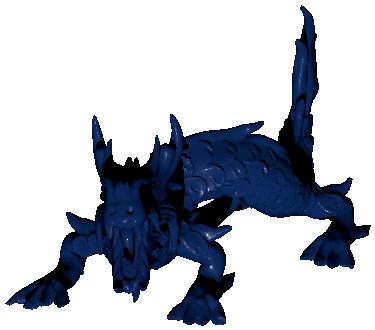}
&
\includegraphics[width=\mywidthx, height=\myheightx, keepaspectratio]{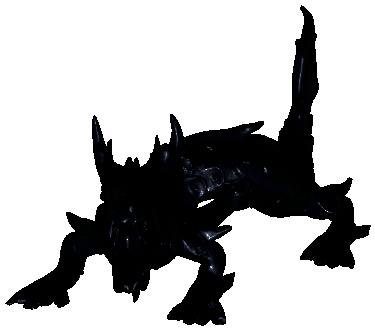}
&
\includegraphics[width=\mywidthx, height=\myheightx, keepaspectratio]{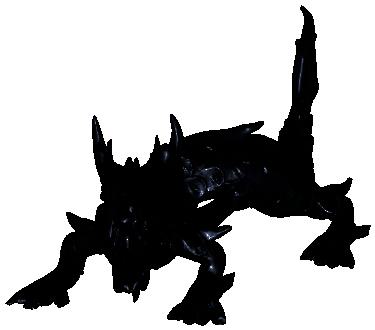}
&
&
\\
&
\rotatebox{90}{\centering \emph{added}} 
&
\includegraphics[width=\mywidthx, height=\myheightx, keepaspectratio]{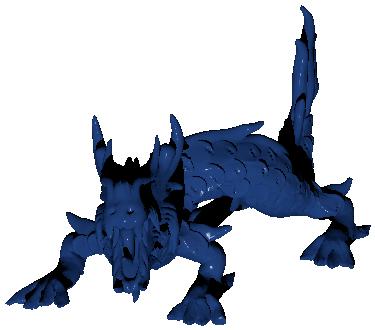}
&
\includegraphics[width=\mywidthx, height=\myheightx, keepaspectratio]{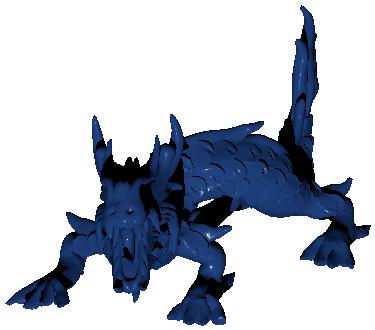}
&
\includegraphics[width=\mywidthx, height=\myheightx, keepaspectratio]{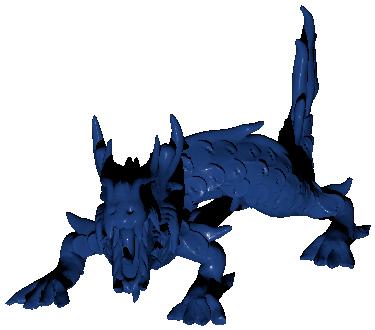}
&
\includegraphics[width=\mywidthx, height=\myheightx, keepaspectratio]{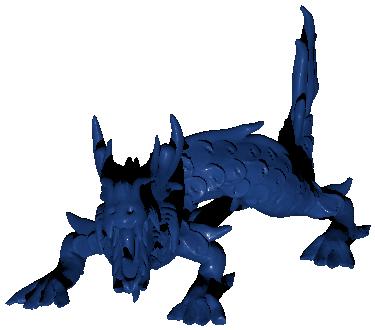}
&
\includegraphics[width=\mywidthx, height=\myheightx, keepaspectratio]{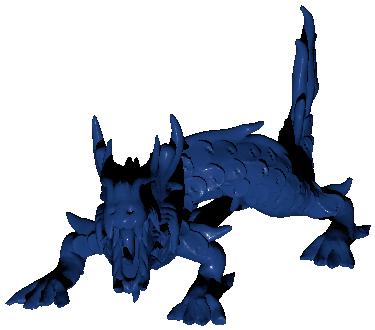}
&
\includegraphics[width=\mywidthx, height=\myheightx, keepaspectratio]{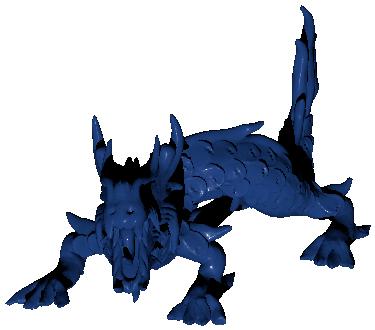}
&
\includegraphics[width=\mywidthx, height=\myheightx, keepaspectratio]{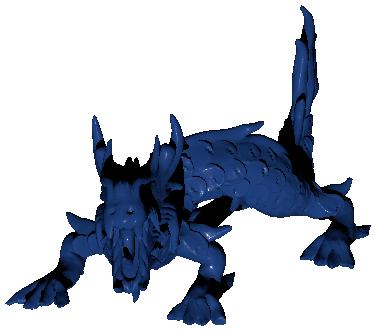}
&
\includegraphics[width=\mywidthx, height=\myheightx, keepaspectratio]{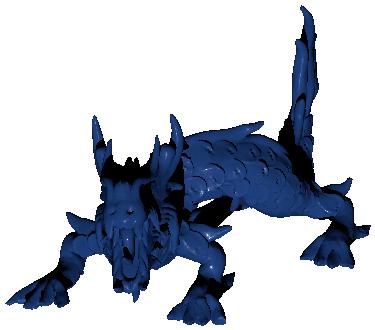}
&
&
\includegraphics[width=\mywidthx, height=\myheightx, keepaspectratio]{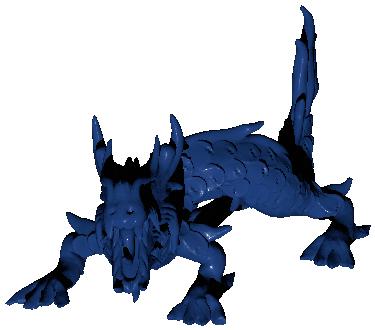}\\
\hline\hline\\
\multirow{2}{*}{\rotatebox{90}{\parbox{3cm}{\centering \merlc \\ alumina oxide}}} 
&
\rotatebox{90}{\centering \emph{diffuse}} 
&
\includegraphics[width=\mywidthx, height=\myheightx, keepaspectratio]{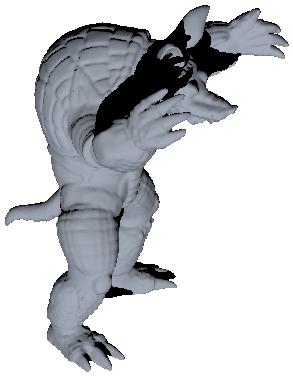}
&
\includegraphics[width=\mywidthx, height=\myheightx, keepaspectratio]{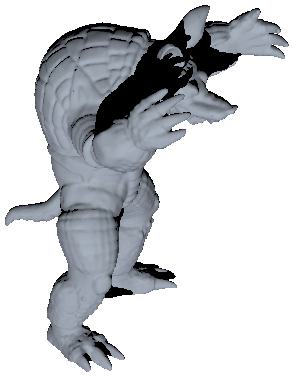}
&
\includegraphics[width=\mywidthx, height=\myheightx, keepaspectratio]{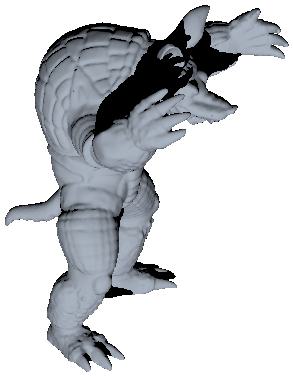}
&
\includegraphics[width=\mywidthx, height=\myheightx, keepaspectratio]{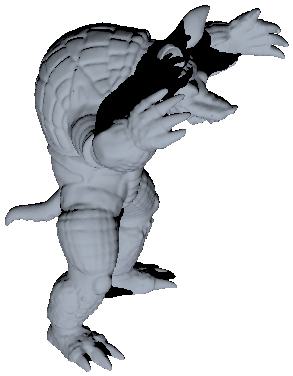}
&
\includegraphics[width=\mywidthx, height=\myheightx, keepaspectratio]{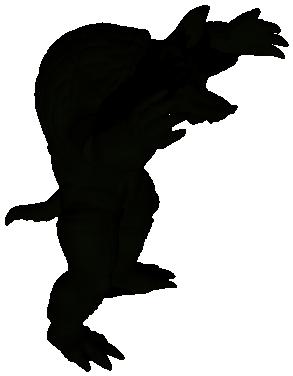}
&
\includegraphics[width=\mywidthx, height=\myheightx, keepaspectratio]{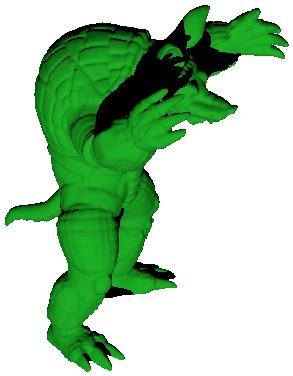}
&
\includegraphics[width=\mywidthx, height=\myheightx, keepaspectratio]{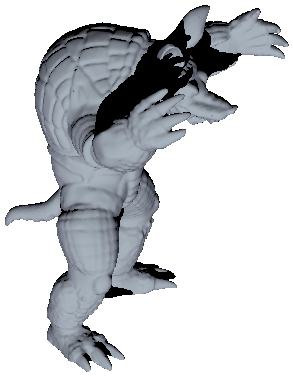}
&
\includegraphics[width=\mywidthx, height=\myheightx, keepaspectratio]{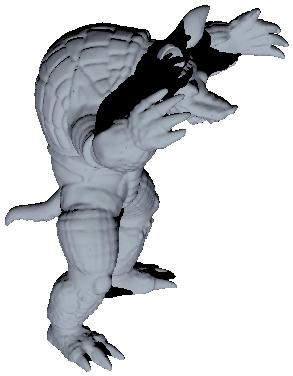}
&
&
\\
&
\rotatebox{90}{\centering \emph{specular}} 
&
\includegraphics[width=\mywidthx, height=\myheightx, keepaspectratio]{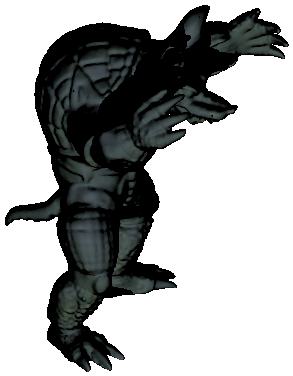}
&
\includegraphics[width=\mywidthx, height=\myheightx, keepaspectratio]{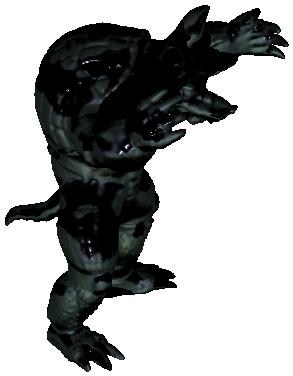}
&
\includegraphics[width=\mywidthx, height=\myheightx, keepaspectratio]{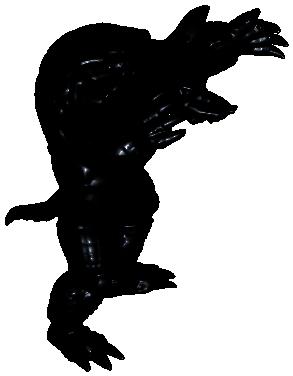}
&
\includegraphics[width=\mywidthx, height=\myheightx, keepaspectratio]{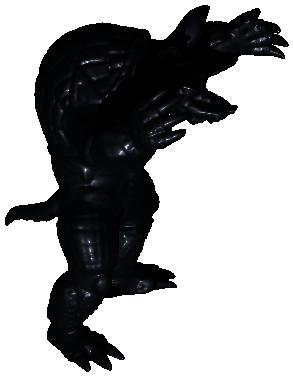}
&
\includegraphics[width=\mywidthx, height=\myheightx, keepaspectratio]{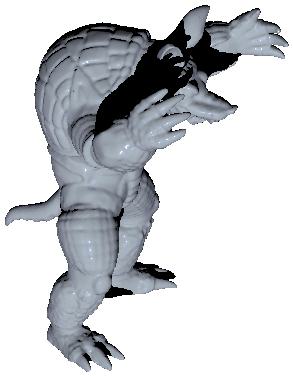}
&
\includegraphics[width=\mywidthx, height=\myheightx, keepaspectratio]{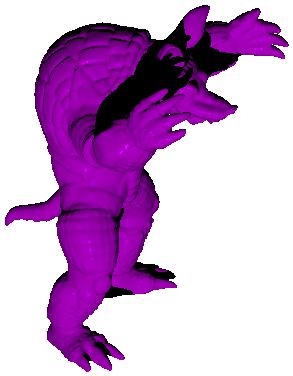}
&
\includegraphics[width=\mywidthx, height=\myheightx, keepaspectratio]{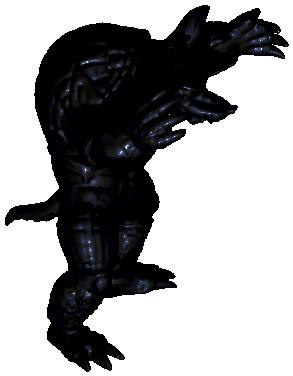}
&
\includegraphics[width=\mywidthx, height=\myheightx, keepaspectratio]{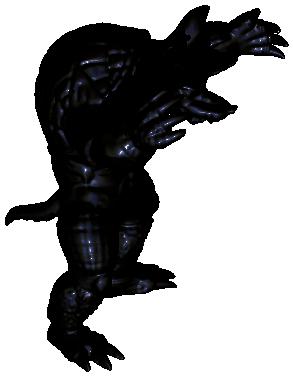}
&
&
\\
&
\rotatebox{90}{\centering \emph{added}} 
&
\includegraphics[width=\mywidthx, height=\myheightx, keepaspectratio]{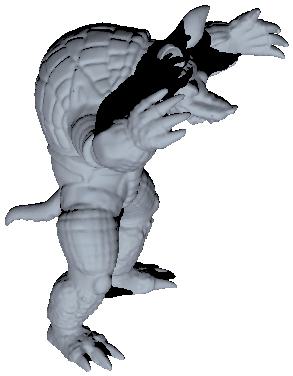}
&
\includegraphics[width=\mywidthx, height=\myheightx, keepaspectratio]{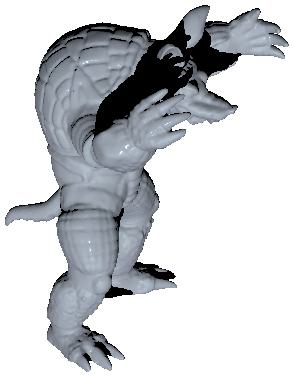}
&
\includegraphics[width=\mywidthx, height=\myheightx, keepaspectratio]{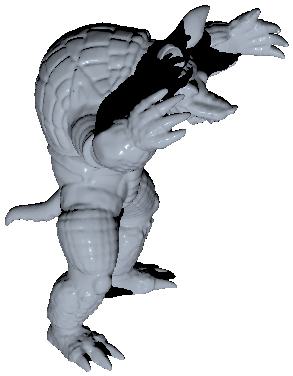}
&
\includegraphics[width=\mywidthx, height=\myheightx, keepaspectratio]{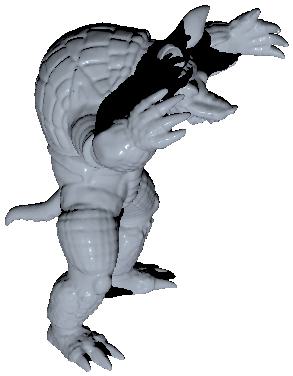}
&
\includegraphics[width=\mywidthx, height=\myheightx, keepaspectratio]{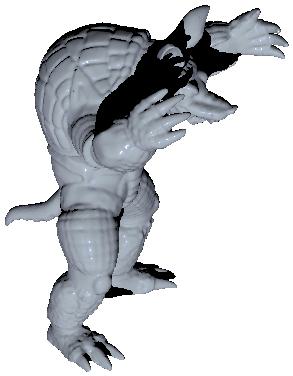}
&
\includegraphics[width=\mywidthx, height=\myheightx, keepaspectratio]{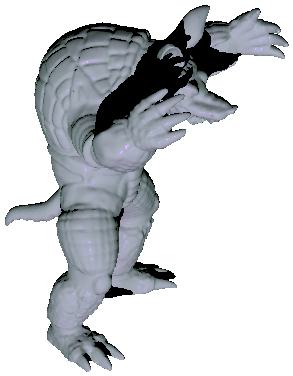}
&
\includegraphics[width=\mywidthx, height=\myheightx, keepaspectratio]{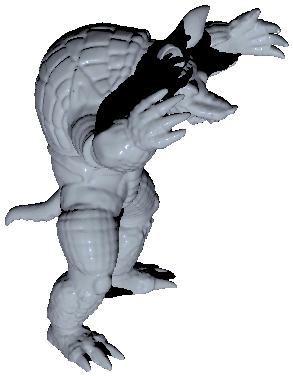}
&
\includegraphics[width=\mywidthx, height=\myheightx, keepaspectratio]{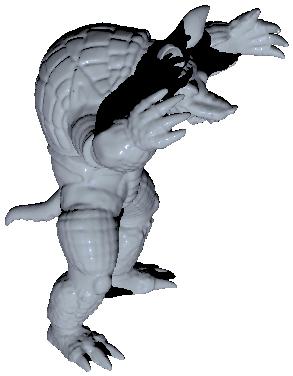}
&
&
\includegraphics[width=\mywidthx, height=\myheightx, keepaspectratio]{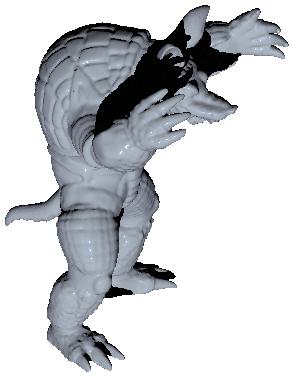}\\
\hline\hline\\
\multirow{2}{*}{\rotatebox{90}{\parbox{3cm}{\centering \merlc \\ green acrylic}}} 
&
\rotatebox{90}{\centering \emph{diffuse}} 
&
\includegraphics[width=\mywidthx, height=\myheightx, keepaspectratio]{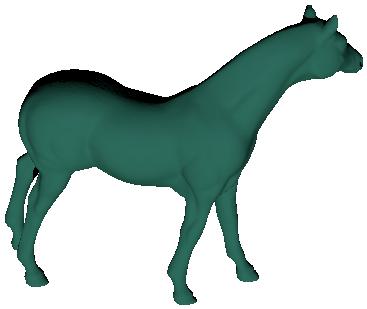}
&
\includegraphics[width=\mywidthx, height=\myheightx, keepaspectratio]{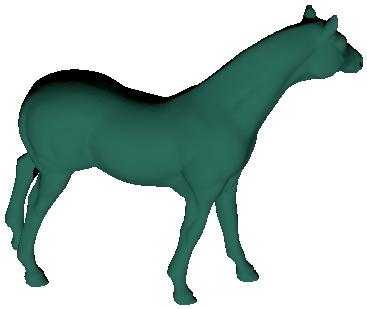}
&
\includegraphics[width=\mywidthx, height=\myheightx, keepaspectratio]{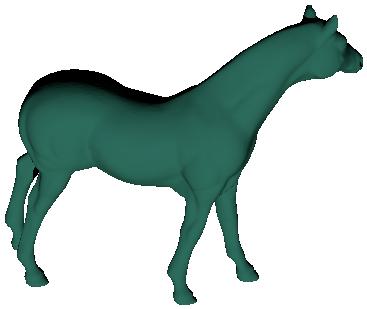}
&
\includegraphics[width=\mywidthx, height=\myheightx, keepaspectratio]{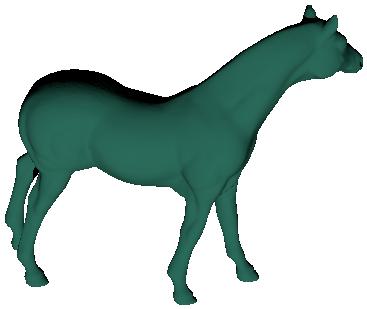}
&
\includegraphics[width=\mywidthx, height=\myheightx, keepaspectratio]{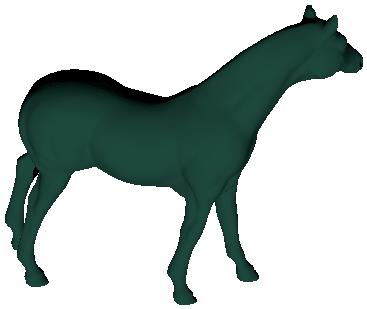}
&
\includegraphics[width=\mywidthx, height=\myheightx, keepaspectratio]{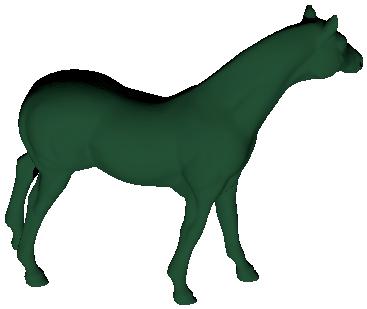}
&
\includegraphics[width=\mywidthx, height=\myheightx, keepaspectratio]{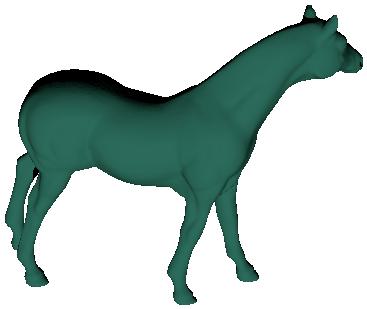}
&
\includegraphics[width=\mywidthx, height=\myheightx, keepaspectratio]{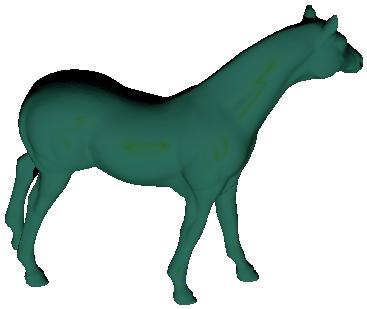}
&
&
\\
&
\rotatebox{90}{\centering \emph{specular}} 
&
\includegraphics[width=\mywidthx, height=\myheightx, keepaspectratio]{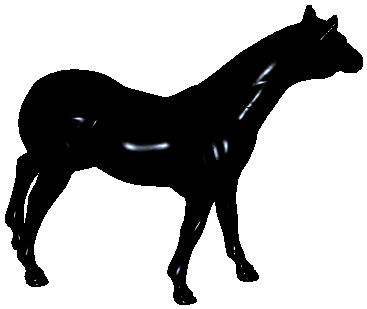}
&
\includegraphics[width=\mywidthx, height=\myheightx, keepaspectratio]{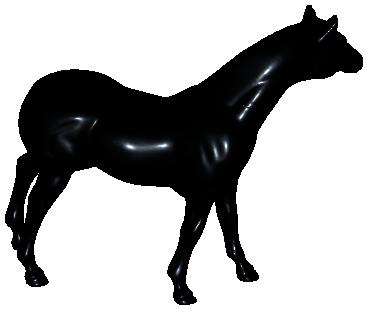}
&
\includegraphics[width=\mywidthx, height=\myheightx, keepaspectratio]{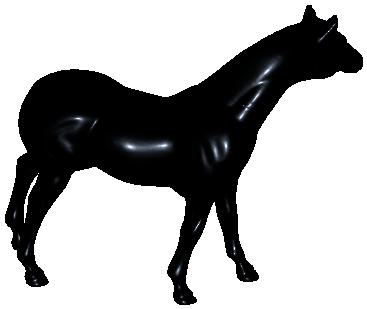}
&
\includegraphics[width=\mywidthx, height=\myheightx, keepaspectratio]{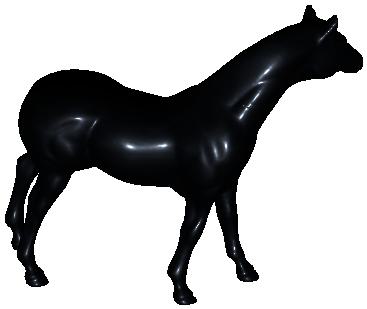}
&
\includegraphics[width=\mywidthx, height=\myheightx, keepaspectratio]{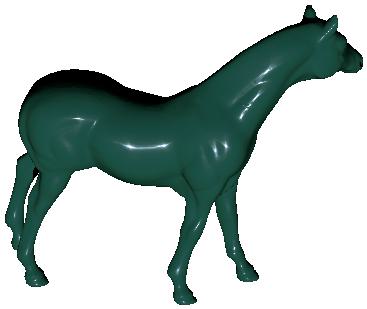}
&
\includegraphics[width=\mywidthx, height=\myheightx, keepaspectratio]{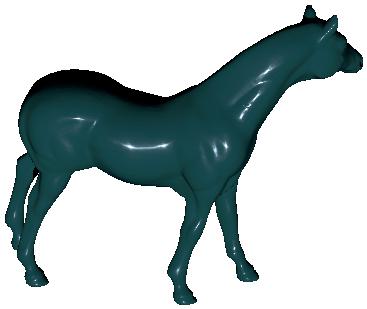}
&
\includegraphics[width=\mywidthx, height=\myheightx, keepaspectratio]{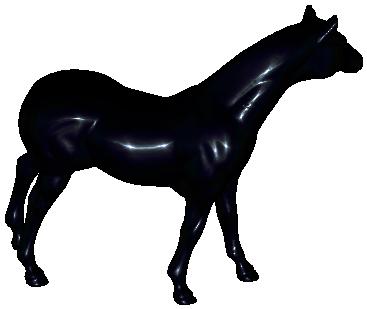}
&
\includegraphics[width=\mywidthx, height=\myheightx, keepaspectratio]{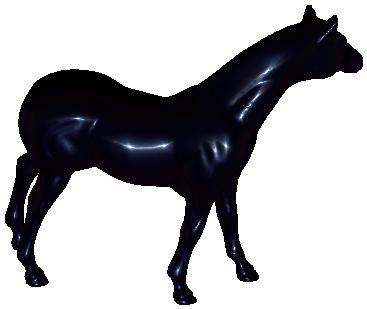}
&
&
\\
&
\rotatebox{90}{\centering \emph{added}} 
&
\includegraphics[width=\mywidthx, height=\myheightx, keepaspectratio]{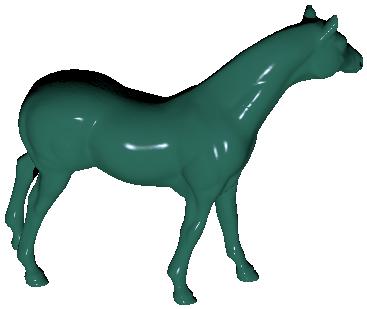}
&
\includegraphics[width=\mywidthx, height=\myheightx, keepaspectratio]{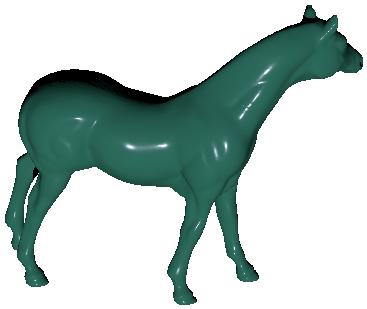}
&
\includegraphics[width=\mywidthx, height=\myheightx, keepaspectratio]{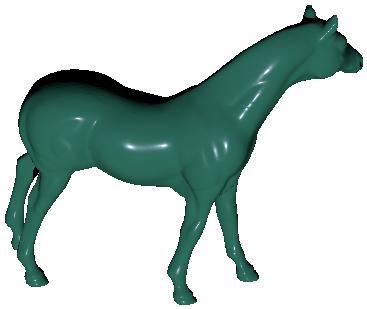}
&
\includegraphics[width=\mywidthx, height=\myheightx, keepaspectratio]{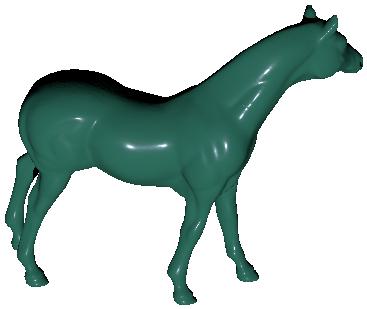}
&
\includegraphics[width=\mywidthx, height=\myheightx, keepaspectratio]{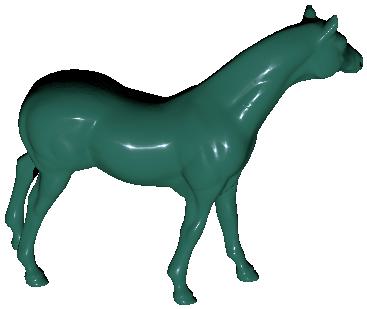}
&
\includegraphics[width=\mywidthx, height=\myheightx, keepaspectratio]{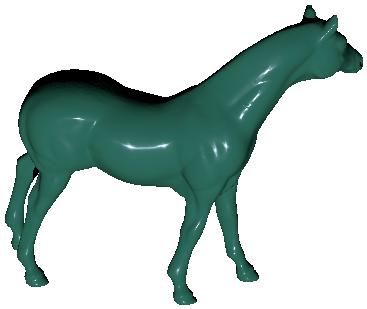}
&
\includegraphics[width=\mywidthx, height=\myheightx, keepaspectratio]{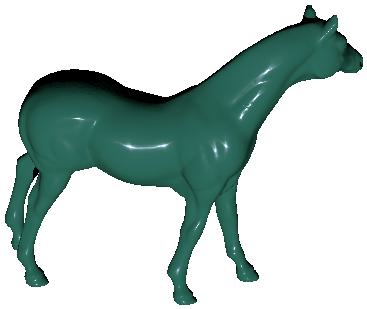}
&
\includegraphics[width=\mywidthx, height=\myheightx, keepaspectratio]{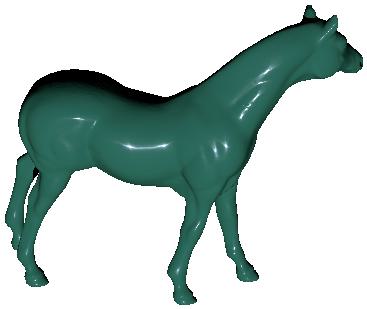}
&
&
\includegraphics[width=\mywidthx, height=\myheightx, keepaspectratio]{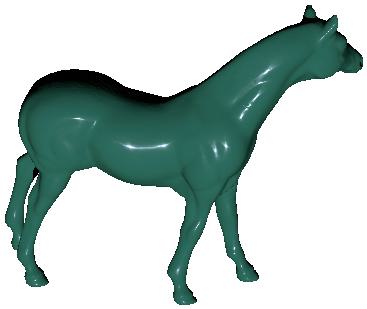}\\
\hline\hline\\[-0.2cm]
&
 & \cellcolor{cellParamBased}\rpc		%
 & \cellcolor{cellParamBased}\tsc		%
 & \cellcolor{cellParamBased}\fmbrdfc		%
 & \cellcolor{cellParamBased}\disneyc		%
 & \cellcolor{celPurelyNeural}Add Sep		%
 & \cellcolor{celPurelyNeural}Add Shared		%
 & \cellcolor{celPurelyNeural}Add Sep (enh.)		%
 & \cellcolor{celPurelyNeural}Add Shared (enh.)		%
& %
 & \gt
  \end{tabular}
\caption{
Renderings of the diffuse and the specular parts separately for all additive models. Note that for the models with the enhanced additive strategy (\emph{enh.}), the diffuse part is already weighted with $\xi$. Also shown are the combined rendering (\emph{added}) and the ground truth image (\emph{GT}). 
The figure shows non-metallic objects. Most models show a reasonable split into diffuse albedo and specular highlights. However, the figure reveals an issue that we observed occasionally for the vanilla purely neural models based on an additive split: For some materials, like the alumina oxide in this case, there seems to be an ambiguity that allows the model to perform an unreasonable ``color-split''. While the added result yields the correct colors for all of our experiments, and we do not observe a reduced reconstruction quality, this ambiguity might cause problems in specific cases. Note that our enhancement for the additive split (\emph{enh.}) as introduced in 
\iftoggle{arxiv}{\cref{sec:enhancingAddSplit}}{Sec.~4.4}
and in particular the regularizers discussed in 
\iftoggle{arxiv}{\cref{sec:enhancingAddSplit}}{Sec.~4.4} and \cref{sec:supp:regularizers_enhanced} eliminate this issue.
}
\label{fig:supp_diff_spec_synth_1}
\end{figure*}
\begin{figure*}[t]  %
  \centering  %
  \footnotesize
  \newcommand{\mywidthc}{0.02\textwidth}  %
  \newcommand{\mywidthx}{0.10\textwidth}  %
  \newcommand{\mywidthw}{0.008\textwidth}  %
  \newcommand{\myheightx}{0.15\textwidth}  %
  \newcommand{\mywidtht}{0.04\textwidth}  %
  \newcolumntype{C}{ >{\centering\arraybackslash} m{\mywidthc} } %
  \newcolumntype{X}{ >{\centering\arraybackslash} m{\mywidthx} } %
  \newcolumntype{W}{ >{\centering\arraybackslash} m{\mywidthw} } %
  \newcolumntype{T}{ >{\centering\arraybackslash} m{\mywidtht} } %

  \newcommand{\heightcolorbar}{0.10\textwidth}  %
  \newcommand{\xposOne}{-0.95}
  \newcommand{\yposOne}{0.35}
  \newcommand{\xposTwo}{0.35}
  \newcommand{\yposTwo}{0.8}

  \newcommand{\fontsizePSNR}{\ssmall}
  
  \setlength\tabcolsep{0pt} %

  \setlength{\extrarowheight}{1.25pt}
  
  \def\arraystretch{0.8} %
  \begin{tabular}{TTXXXXXXXXWX}

\multirow{2}{*}{\rotatebox{90}{\parbox{3cm}{\centering \merlc \\ hematite}}} 
&
\rotatebox{90}{\centering \emph{diffuse}} 
&
\includegraphics[width=\mywidthx, height=\myheightx, keepaspectratio]{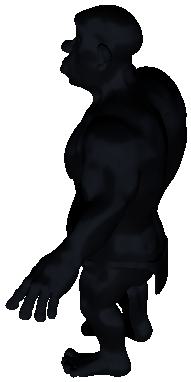}
&
\includegraphics[width=\mywidthx, height=\myheightx, keepaspectratio]{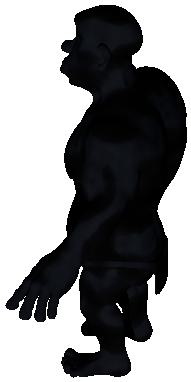}
&
\includegraphics[width=\mywidthx, height=\myheightx, keepaspectratio]{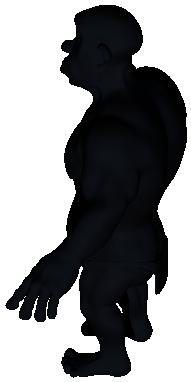}
&
\includegraphics[width=\mywidthx, height=\myheightx, keepaspectratio]{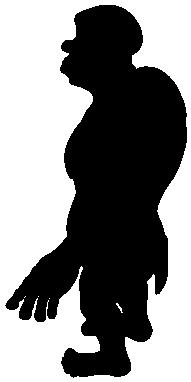}
&
\includegraphics[width=\mywidthx, height=\myheightx, keepaspectratio]{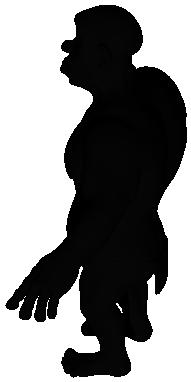}
&
\includegraphics[width=\mywidthx, height=\myheightx, keepaspectratio]{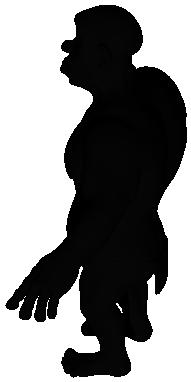}
&
\includegraphics[width=\mywidthx, height=\myheightx, keepaspectratio]{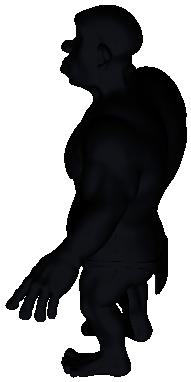}
&
\includegraphics[width=\mywidthx, height=\myheightx, keepaspectratio]{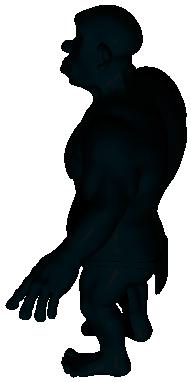}
&
&
\\
&
\rotatebox{90}{\centering \emph{specular}} 
&
\includegraphics[width=\mywidthx, height=\myheightx, keepaspectratio]{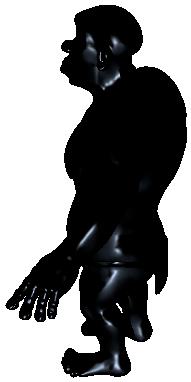}
&
\includegraphics[width=\mywidthx, height=\myheightx, keepaspectratio]{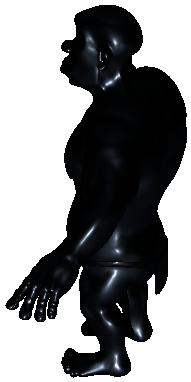}
&
\includegraphics[width=\mywidthx, height=\myheightx, keepaspectratio]{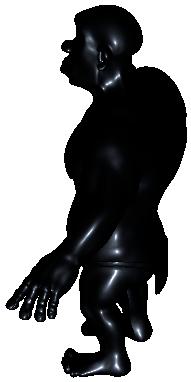}
&
\includegraphics[width=\mywidthx, height=\myheightx, keepaspectratio]{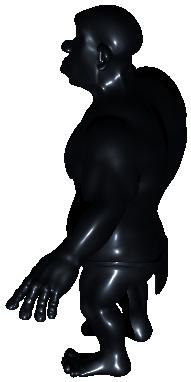}
&
\includegraphics[width=\mywidthx, height=\myheightx, keepaspectratio]{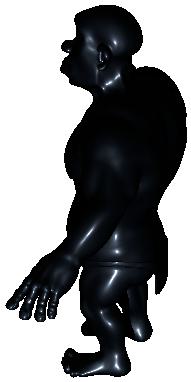}
&
\includegraphics[width=\mywidthx, height=\myheightx, keepaspectratio]{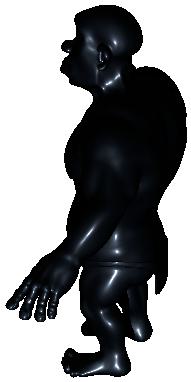}
&
\includegraphics[width=\mywidthx, height=\myheightx, keepaspectratio]{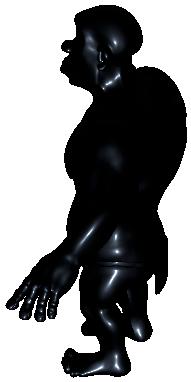}
&
\includegraphics[width=\mywidthx, height=\myheightx, keepaspectratio]{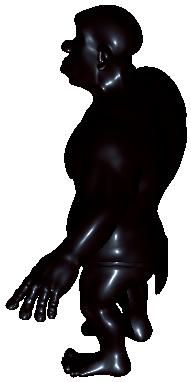}
&
&
\\
&
\rotatebox{90}{\centering \emph{added}} 
&
\includegraphics[width=\mywidthx, height=\myheightx, keepaspectratio]{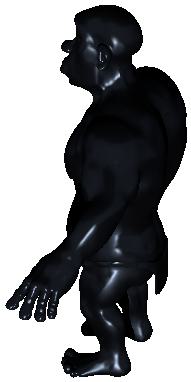}
&
\includegraphics[width=\mywidthx, height=\myheightx, keepaspectratio]{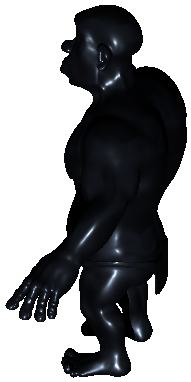}
&
\includegraphics[width=\mywidthx, height=\myheightx, keepaspectratio]{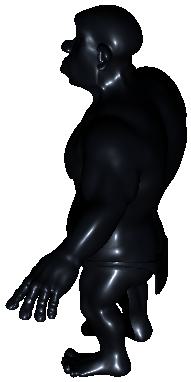}
&
\includegraphics[width=\mywidthx, height=\myheightx, keepaspectratio]{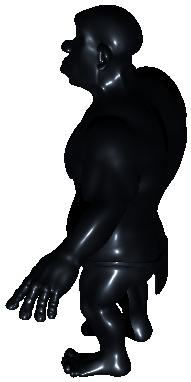}
&
\includegraphics[width=\mywidthx, height=\myheightx, keepaspectratio]{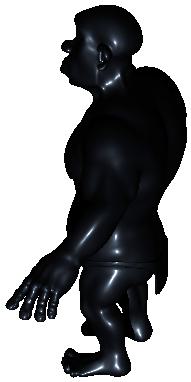}
&
\includegraphics[width=\mywidthx, height=\myheightx, keepaspectratio]{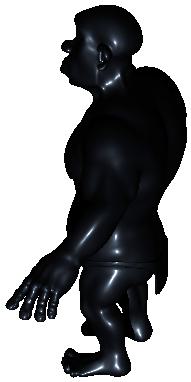}
&
\includegraphics[width=\mywidthx, height=\myheightx, keepaspectratio]{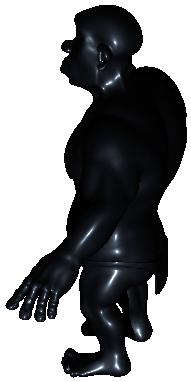}
&
\includegraphics[width=\mywidthx, height=\myheightx, keepaspectratio]{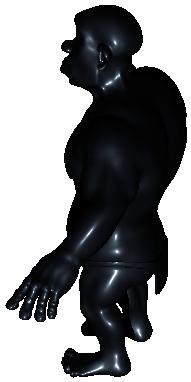}
&
&
\includegraphics[width=\mywidthx, height=\myheightx, keepaspectratio]{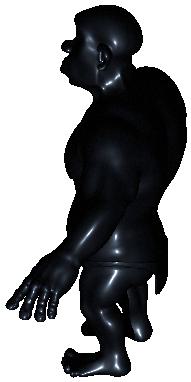}\\
\hline\hline\\
\multirow{2}{*}{\rotatebox{90}{\parbox{3cm}{\centering \merlc \\ green metallic paint2}}} 
&
\rotatebox{90}{\centering \emph{diffuse}} 
&
\includegraphics[width=\mywidthx, height=\myheightx, keepaspectratio]{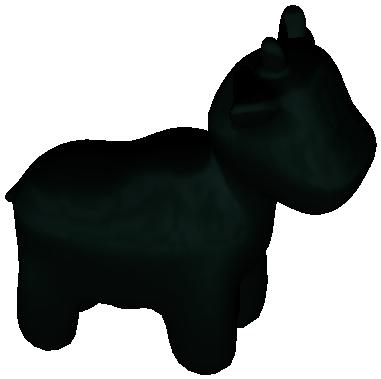}
&
\includegraphics[width=\mywidthx, height=\myheightx, keepaspectratio]{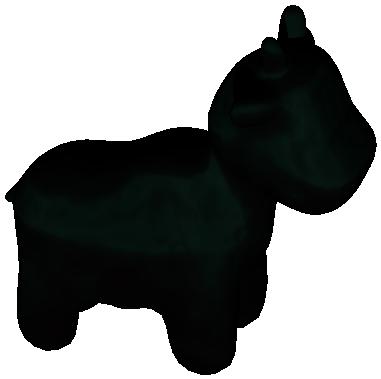}
&
\includegraphics[width=\mywidthx, height=\myheightx, keepaspectratio]{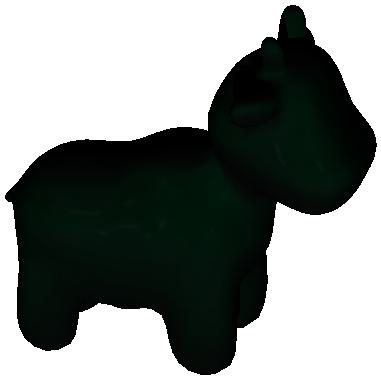}
&
\includegraphics[width=\mywidthx, height=\myheightx, keepaspectratio]{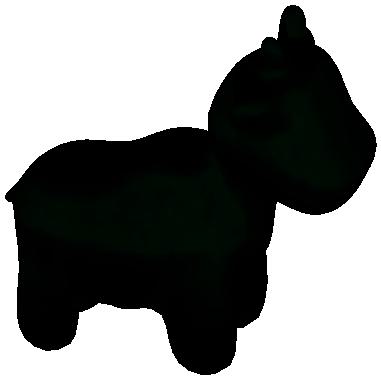}
&
\includegraphics[width=\mywidthx, height=\myheightx, keepaspectratio]{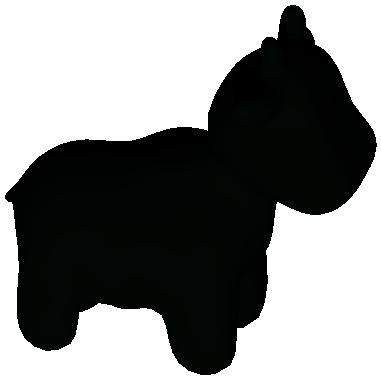}
&
\includegraphics[width=\mywidthx, height=\myheightx, keepaspectratio]{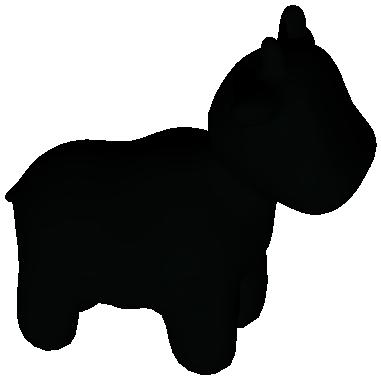}
&
\includegraphics[width=\mywidthx, height=\myheightx, keepaspectratio]{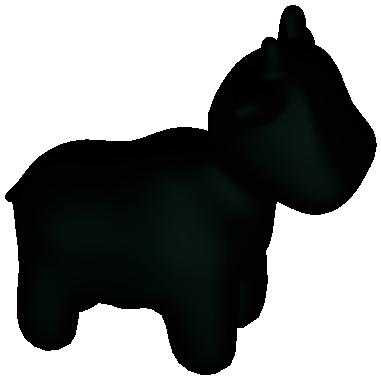}
&
\includegraphics[width=\mywidthx, height=\myheightx, keepaspectratio]{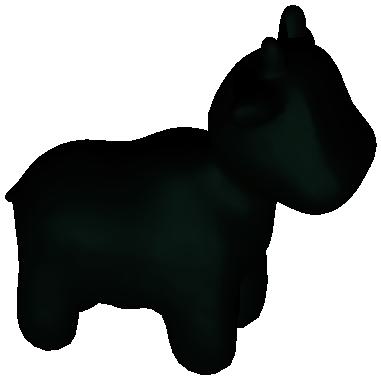}
&
&
\\
&
\rotatebox{90}{\centering \emph{specular}} 
&
\includegraphics[width=\mywidthx, height=\myheightx, keepaspectratio]{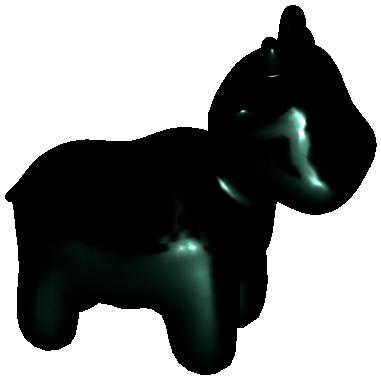}
&
\includegraphics[width=\mywidthx, height=\myheightx, keepaspectratio]{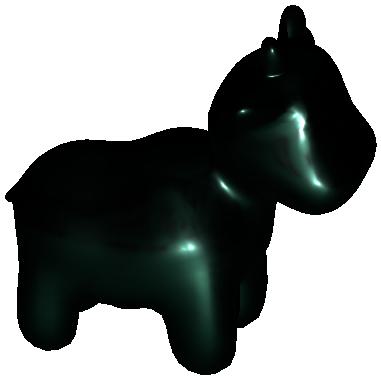}
&
\includegraphics[width=\mywidthx, height=\myheightx, keepaspectratio]{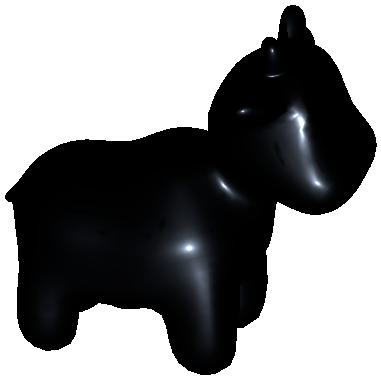}
&
\includegraphics[width=\mywidthx, height=\myheightx, keepaspectratio]{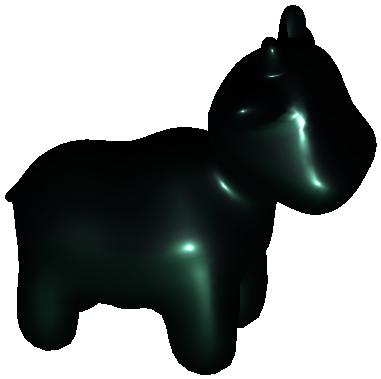}
&
\includegraphics[width=\mywidthx, height=\myheightx, keepaspectratio]{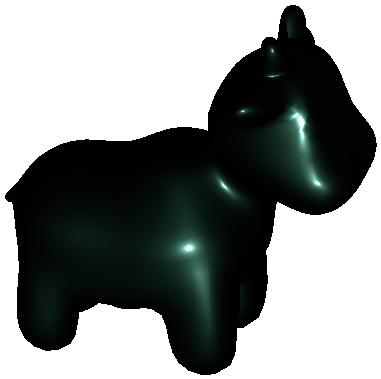}
&
\includegraphics[width=\mywidthx, height=\myheightx, keepaspectratio]{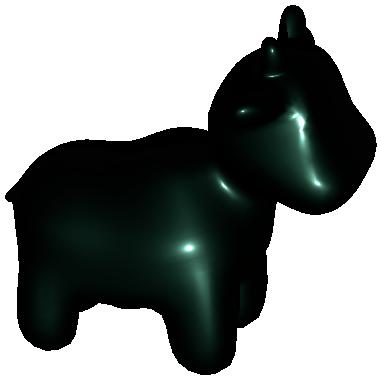}
&
\includegraphics[width=\mywidthx, height=\myheightx, keepaspectratio]{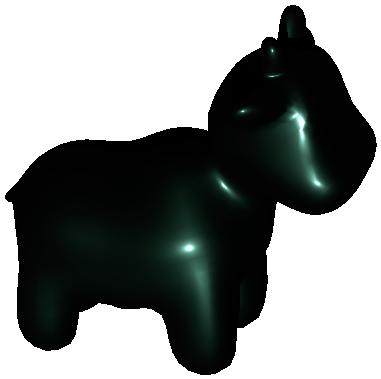}
&
\includegraphics[width=\mywidthx, height=\myheightx, keepaspectratio]{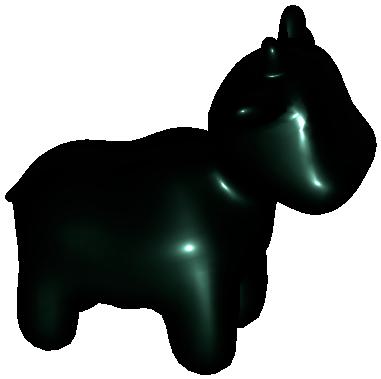}
&
&
\\
&
\rotatebox{90}{\centering \emph{added}} 
&
\includegraphics[width=\mywidthx, height=\myheightx, keepaspectratio]{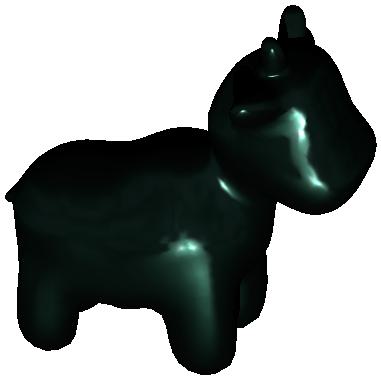}
&
\includegraphics[width=\mywidthx, height=\myheightx, keepaspectratio]{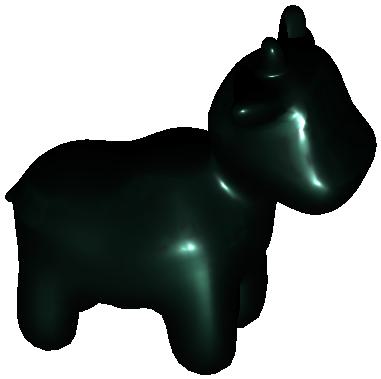}
&
\includegraphics[width=\mywidthx, height=\myheightx, keepaspectratio]{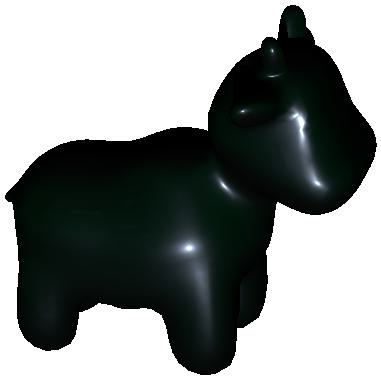}
&
\includegraphics[width=\mywidthx, height=\myheightx, keepaspectratio]{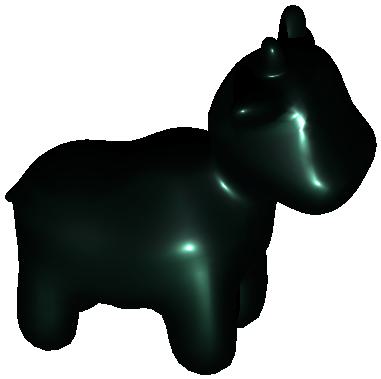}
&
\includegraphics[width=\mywidthx, height=\myheightx, keepaspectratio]{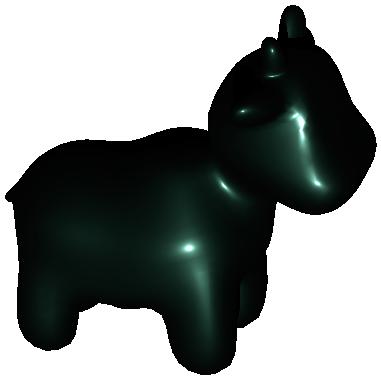}
&
\includegraphics[width=\mywidthx, height=\myheightx, keepaspectratio]{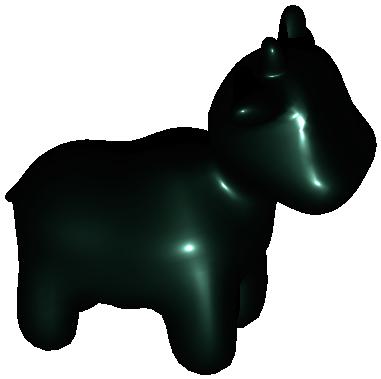}
&
\includegraphics[width=\mywidthx, height=\myheightx, keepaspectratio]{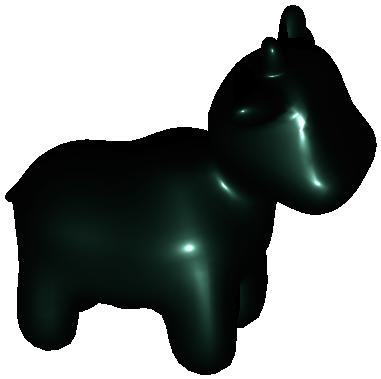}
&
\includegraphics[width=\mywidthx, height=\myheightx, keepaspectratio]{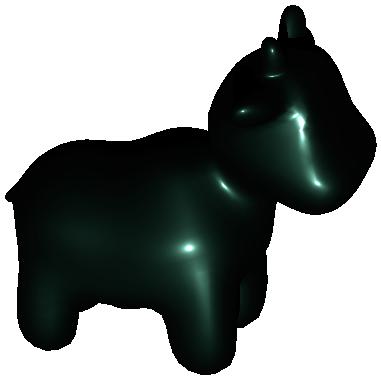}
&
&
\includegraphics[width=\mywidthx, height=\myheightx, keepaspectratio]{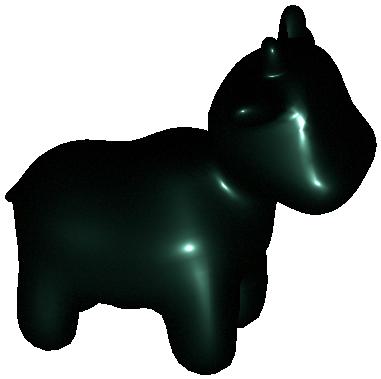}\\
\hline\hline\\
\multirow{2}{*}{\rotatebox{90}{\parbox{3cm}{\centering \merlc \\ chrome steel}}} 
&
\rotatebox{90}{\centering \emph{diffuse}} 
&
\includegraphics[width=\mywidthx, height=\myheightx, keepaspectratio]{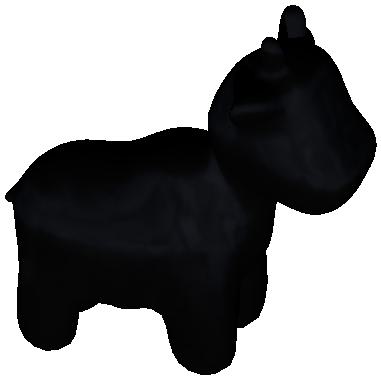}
&
\includegraphics[width=\mywidthx, height=\myheightx, keepaspectratio]{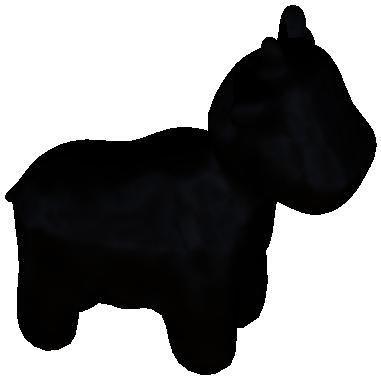}
&
\includegraphics[width=\mywidthx, height=\myheightx, keepaspectratio]{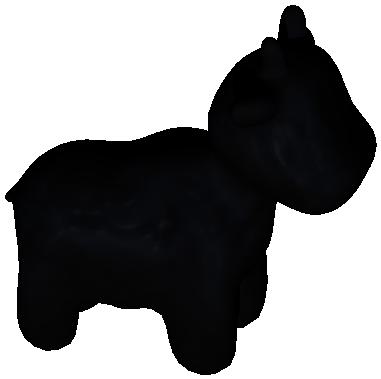}
&
\includegraphics[width=\mywidthx, height=\myheightx, keepaspectratio]{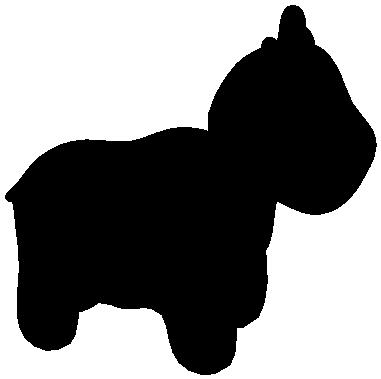}
&
\includegraphics[width=\mywidthx, height=\myheightx, keepaspectratio]{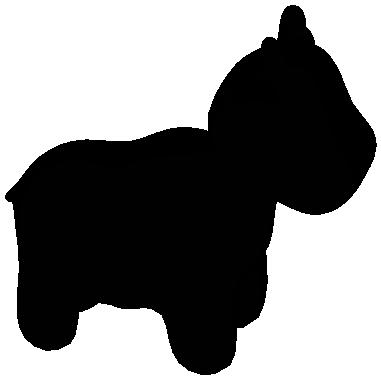}
&
\includegraphics[width=\mywidthx, height=\myheightx, keepaspectratio]{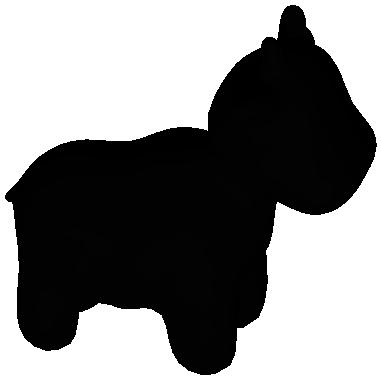}
&
\includegraphics[width=\mywidthx, height=\myheightx, keepaspectratio]{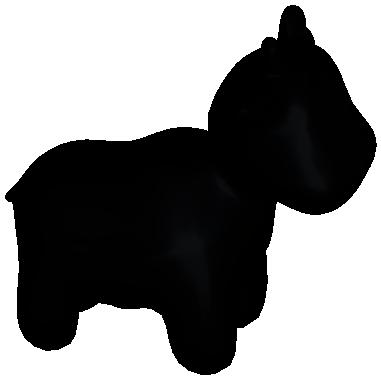}
&
\includegraphics[width=\mywidthx, height=\myheightx, keepaspectratio]{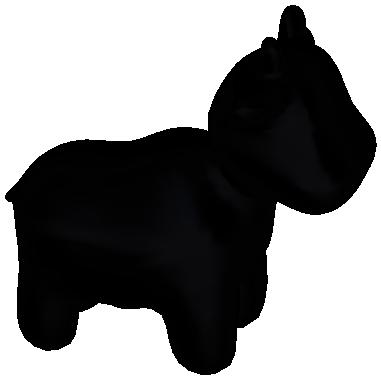}
&
&
\\
&
\rotatebox{90}{\centering \emph{specular}} 
&
\includegraphics[width=\mywidthx, height=\myheightx, keepaspectratio]{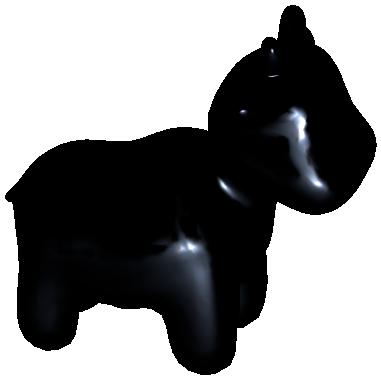}
&
\includegraphics[width=\mywidthx, height=\myheightx, keepaspectratio]{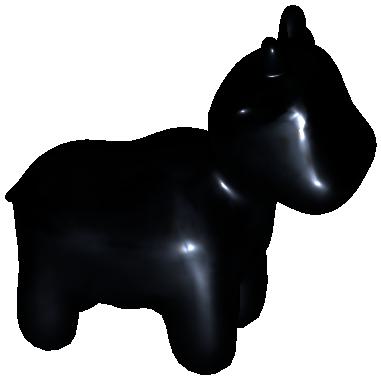}
&
\includegraphics[width=\mywidthx, height=\myheightx, keepaspectratio]{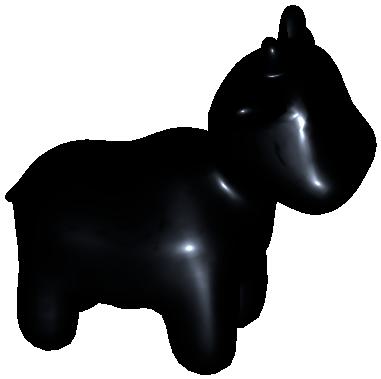}
&
\includegraphics[width=\mywidthx, height=\myheightx, keepaspectratio]{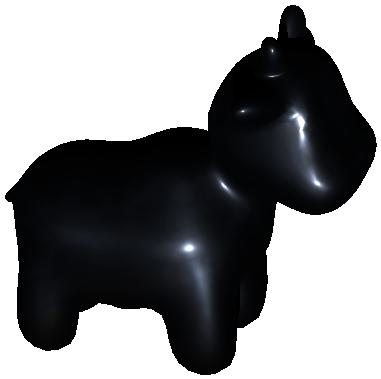}
&
\includegraphics[width=\mywidthx, height=\myheightx, keepaspectratio]{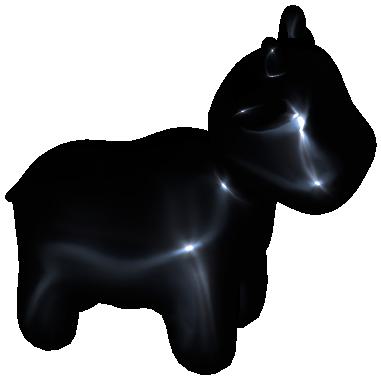}
&
\includegraphics[width=\mywidthx, height=\myheightx, keepaspectratio]{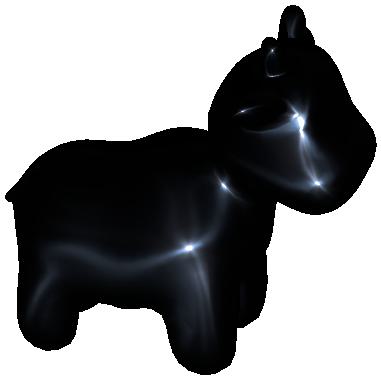}
&
\includegraphics[width=\mywidthx, height=\myheightx, keepaspectratio]{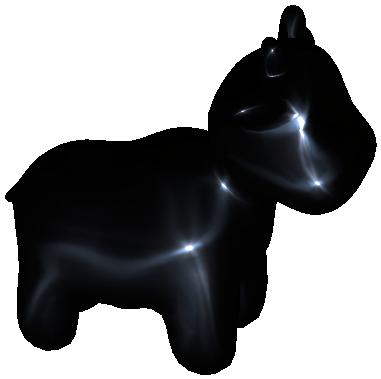}
&
\includegraphics[width=\mywidthx, height=\myheightx, keepaspectratio]{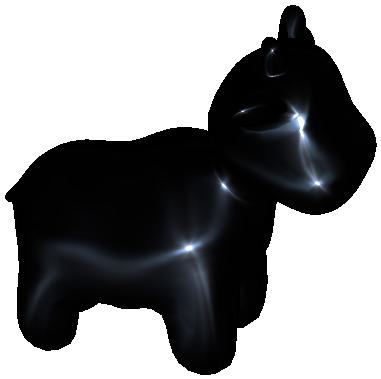}
&
&
\\
&
\rotatebox{90}{\centering \emph{added}} 
&
\includegraphics[width=\mywidthx, height=\myheightx, keepaspectratio]{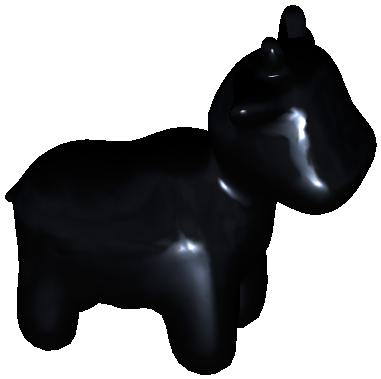}
&
\includegraphics[width=\mywidthx, height=\myheightx, keepaspectratio]{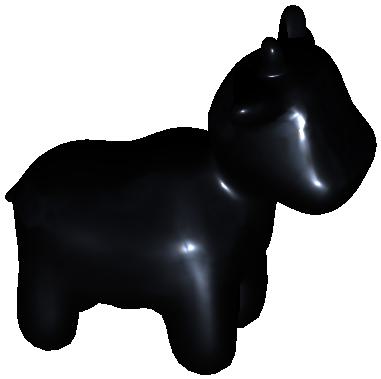}
&
\includegraphics[width=\mywidthx, height=\myheightx, keepaspectratio]{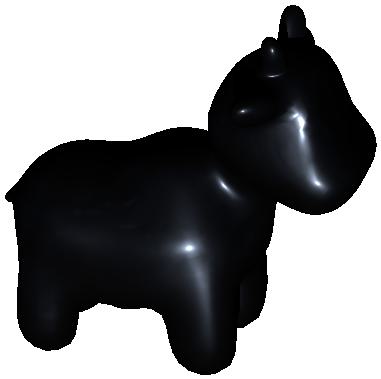}
&
\includegraphics[width=\mywidthx, height=\myheightx, keepaspectratio]{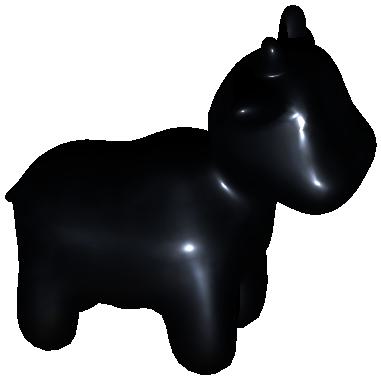}
&
\includegraphics[width=\mywidthx, height=\myheightx, keepaspectratio]{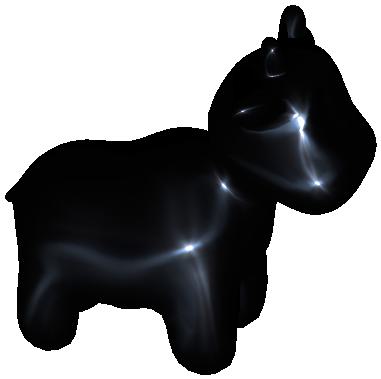}
&
\includegraphics[width=\mywidthx, height=\myheightx, keepaspectratio]{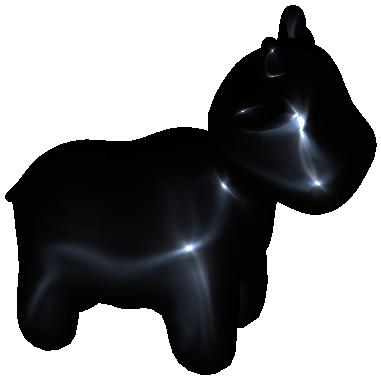}
&
\includegraphics[width=\mywidthx, height=\myheightx, keepaspectratio]{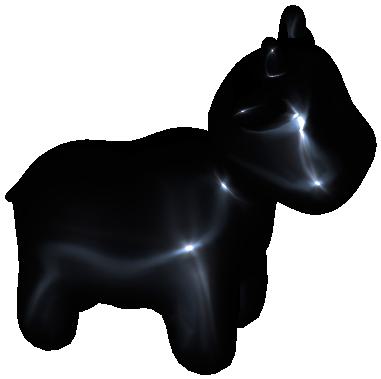}
&
\includegraphics[width=\mywidthx, height=\myheightx, keepaspectratio]{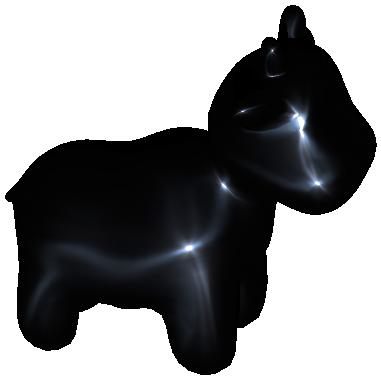}
&
&
\includegraphics[width=\mywidthx, height=\myheightx, keepaspectratio]{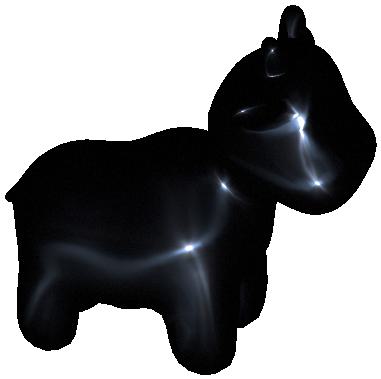}\\
\hline\hline\\[-0.2cm]
&
 & \cellcolor{cellParamBased}\rpc		%
 & \cellcolor{cellParamBased}\tsc		%
 & \cellcolor{cellParamBased}\fmbrdfc		%
 & \cellcolor{cellParamBased}\disneyc		%
 & \cellcolor{celPurelyNeural}Add Sep		%
 & \cellcolor{celPurelyNeural}Add Shared		%
 & \cellcolor{celPurelyNeural}Add Sep (enh.)		%
 & \cellcolor{celPurelyNeural}Add Shared (enh.)		%
& %
 & \gt
  \end{tabular}
\caption{
Renderings of the diffuse and the specular parts separately for all additive models. Note that for the models with the enhanced additive strategy (\emph{enh.}), the diffuse part is already weighted with $\xi$. Also shown are the combined rendering (\emph{added}) and the ground truth image (\emph{GT}). 
The figure shows metallic objects. This type of material shows almost no subsurface scattering, due to the free electrons \cite{akenine2019realTimeRendering}. All models are able to replicate this behavior, as can be seen clearly by the almost non-existent contribution of the diffuse part.
}
\label{fig:supp_diff_spec_synth_2}
\end{figure*}

In \cref{fig:supp_diff_spec_real,fig:supp_diff_spec_synth_1,fig:supp_diff_spec_synth_2}, we present a quantitative analysis of the diffuse and the specular component of all models that split the BRDF into those two contributions. Overall, the methods mostly show a reasonable split. For the purely neural methods, we notice a tendency to represent a larger fraction of the appearance by the specular part, leading to darker diffuse parts. Our enhancements introduced in \iftoggle{arxiv}{\cref{sec:enhancingAddSplit}}{Sec.~4.4}
and in particular the regularizers discussed in \iftoggle{arxiv}{\cref{sec:enhancingAddSplit}}{Sec.~4.4} and \cref{sec:supp:regularizers_enhanced} seem to improve on the disentanglement of diffuse and specular components. \cref{fig:supp_diff_spec_synth_2} reveals that all models can represent the behavior of metallic objects, where almost no subsurface scattering is present, and indeed predict almost no albedo component.

\subsection{Spatial Variation of the Reconstructed BRDFs}
\label{sec:supp:spat_var_recon_brdf}

\begin{figure*}[t]  %
  \centering  %
  \footnotesize
  \newcommand{\mywidthc}{0.02\textwidth}  %
  \newcommand{\mywidthx}{0.11\textwidth}  %
  \newcommand{\mywidthw}{0.03\textwidth}  %
  \newcommand{\myheightx}{0.14\textwidth}  %
  \newcommand{\mywidtht}{0.035\textwidth}  %
  \newcolumntype{C}{ >{\centering\arraybackslash} m{\mywidthc} } %
  \newcolumntype{X}{ >{\centering\arraybackslash} m{\mywidthx} } %
  \newcolumntype{W}{ >{\centering\arraybackslash} m{\mywidthw} } %
  \newcolumntype{T}{ >{\centering\arraybackslash} m{\mywidtht} } %

  \newcommand{\heightcolorbar}{0.10\textwidth}  %
  \newcommand{\xposOne}{-1.05}
  \newcommand{\yposOne}{0.15}
  \newcommand{\xposTwo}{0.0}
  \newcommand{\yposTwo}{0.6}
  \newcommand{\xposThree}{-0.4}
  \newcommand{\yposThree}{0.7}
  \newcommand{\xposFour}{-0.8}
  \newcommand{\yposFour}{1.05}

  \newcommand{\fontsizePSNR}{\ssmall}
  
  \setlength\tabcolsep{0pt} %

  \setlength{\extrarowheight}{1.25pt}
  
  \def\arraystretch{0.8} %
  \begin{tabular}{TXXXXXXXX}

\rotatebox{90}{\parbox{3cm}{\centering \merlc \\ blue acrylic}} 
 & 
 \includegraphics[width=\mywidthx, height=\myheightx, keepaspectratio]{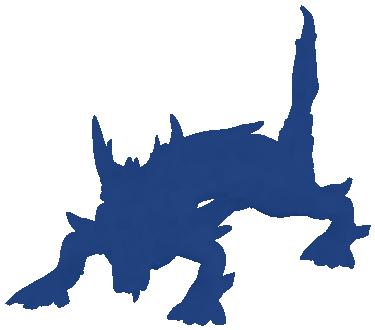}
& 
 \includegraphics[width=\mywidthx, height=\myheightx, keepaspectratio]{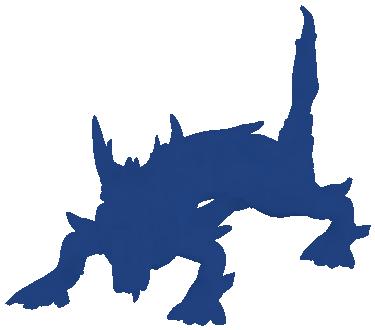}
& 
 \includegraphics[width=\mywidthx, height=\myheightx, keepaspectratio]{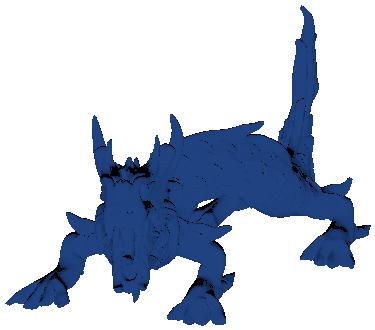}
& 
 \includegraphics[width=\mywidthx, height=\myheightx, keepaspectratio]{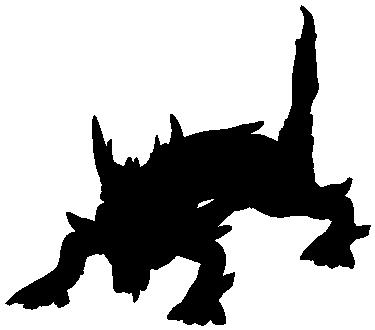}
& 
 \includegraphics[width=\mywidthx, height=\myheightx, keepaspectratio]{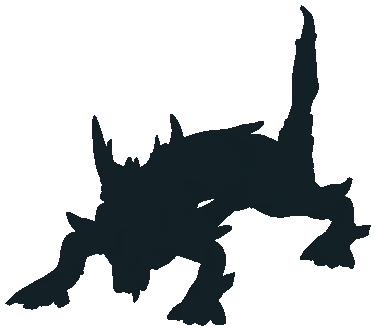}
& 
 \includegraphics[width=\mywidthx, height=\myheightx, keepaspectratio]{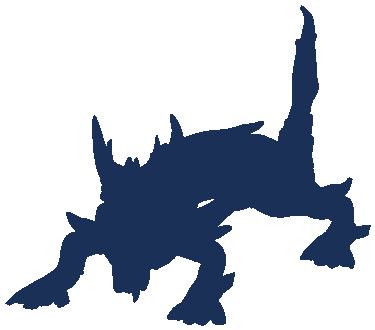}
& 
 \includegraphics[width=\mywidthx, height=\myheightx, keepaspectratio]{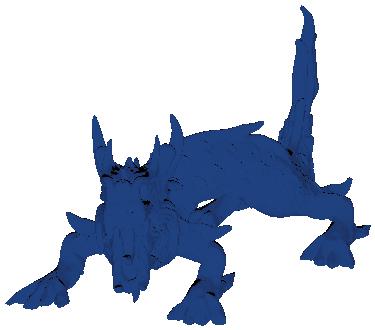}
& 
 \includegraphics[width=\mywidthx, height=\myheightx, keepaspectratio]{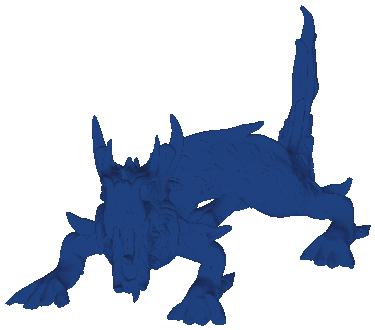}
\\ \hline
\rotatebox{90}{\parbox{3cm}{\centering \merlc \\ ipswich pine 221}} 
 & 
 \includegraphics[width=\mywidthx, height=\myheightx, keepaspectratio]{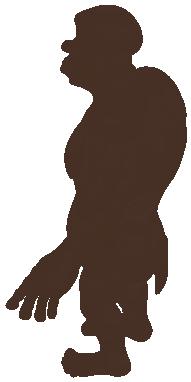}
& 
 \includegraphics[width=\mywidthx, height=\myheightx, keepaspectratio]{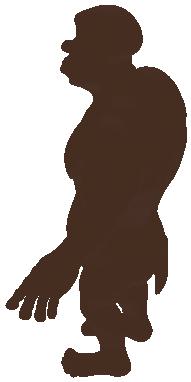}
& 
 \includegraphics[width=\mywidthx, height=\myheightx, keepaspectratio]{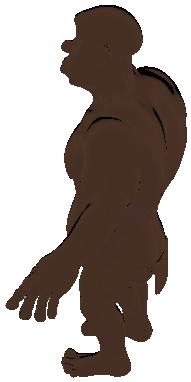}
& 
 \includegraphics[width=\mywidthx, height=\myheightx, keepaspectratio]{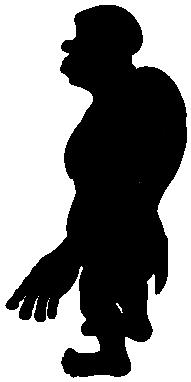}
& 
 \includegraphics[width=\mywidthx, height=\myheightx, keepaspectratio]{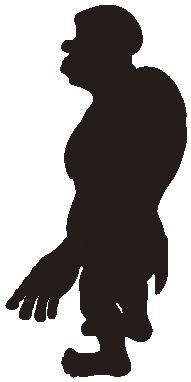}
& 
 \includegraphics[width=\mywidthx, height=\myheightx, keepaspectratio]{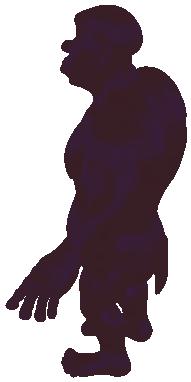}
& 
 \includegraphics[width=\mywidthx, height=\myheightx, keepaspectratio]{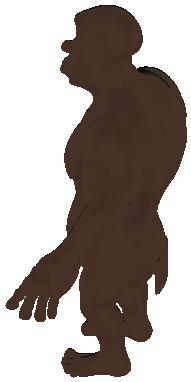}
& 
 \includegraphics[width=\mywidthx, height=\myheightx, keepaspectratio]{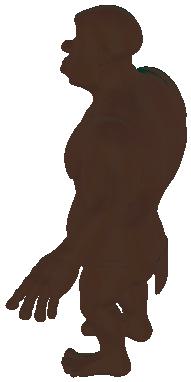}
\\ \hline
\rotatebox{90}{\parbox{3cm}{\centering \merlc \\ white marble}} 
 & 
 \includegraphics[width=\mywidthx, height=\myheightx, keepaspectratio]{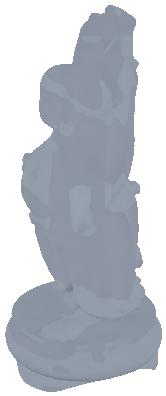}
& 
 \includegraphics[width=\mywidthx, height=\myheightx, keepaspectratio]{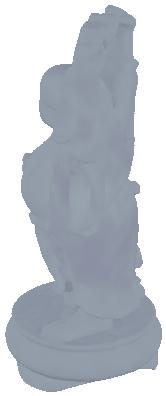}
& 
 \includegraphics[width=\mywidthx, height=\myheightx, keepaspectratio]{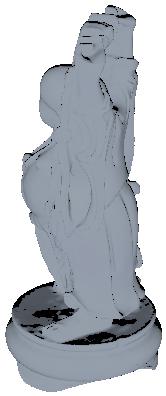}
& 
 \includegraphics[width=\mywidthx, height=\myheightx, keepaspectratio]{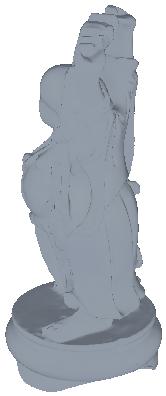}
& 
 \includegraphics[width=\mywidthx, height=\myheightx, keepaspectratio]{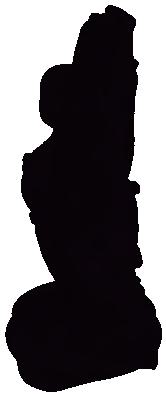}
& 
 \includegraphics[width=\mywidthx, height=\myheightx, keepaspectratio]{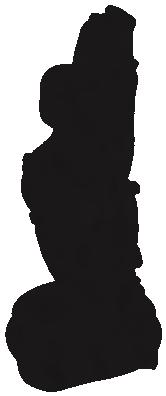}
& 
 \includegraphics[width=\mywidthx, height=\myheightx, keepaspectratio]{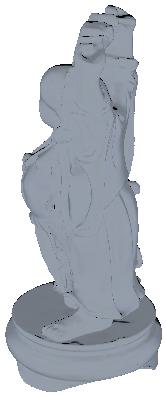}
& 
 \includegraphics[width=\mywidthx, height=\myheightx, keepaspectratio]{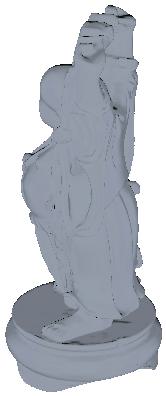}
\\ \hline
\rotatebox{90}{\parbox{3cm}{\centering \merlc \\ maroon plastic}} 
 & 
 \includegraphics[width=\mywidthx, height=\myheightx, keepaspectratio]{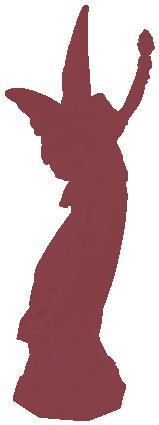}
& 
 \includegraphics[width=\mywidthx, height=\myheightx, keepaspectratio]{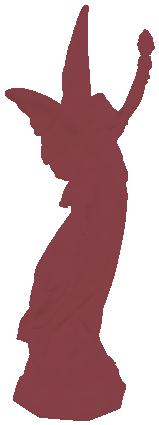}
& 
 \includegraphics[width=\mywidthx, height=\myheightx, keepaspectratio]{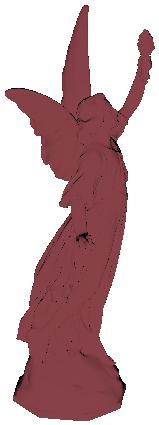}
& 
 \includegraphics[width=\mywidthx, height=\myheightx, keepaspectratio]{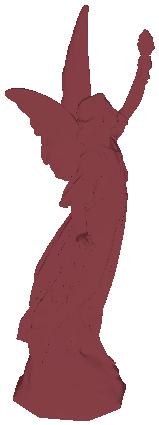}
& 
 \includegraphics[width=\mywidthx, height=\myheightx, keepaspectratio]{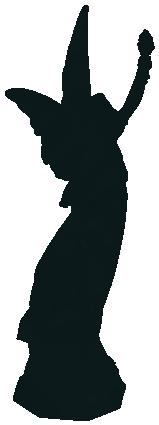}
& 
 \includegraphics[width=\mywidthx, height=\myheightx, keepaspectratio]{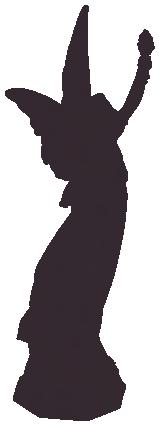}
& 
 \includegraphics[width=\mywidthx, height=\myheightx, keepaspectratio]{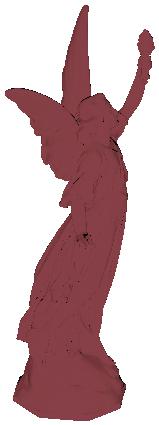}
& 
 \includegraphics[width=\mywidthx, height=\myheightx, keepaspectratio]{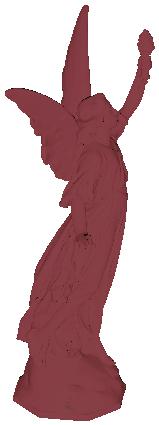}
\\ \hline
\rotatebox{90}{\parbox{3cm}{\centering \merlc \\ green latex}} 
 & 
 \includegraphics[width=\mywidthx, height=\myheightx, keepaspectratio]{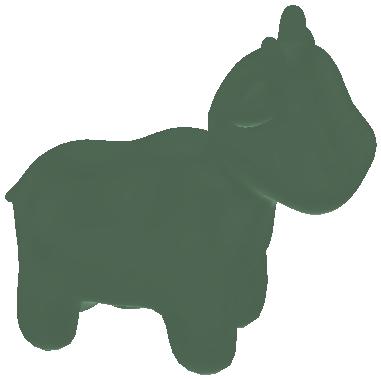}
& 
 \includegraphics[width=\mywidthx, height=\myheightx, keepaspectratio]{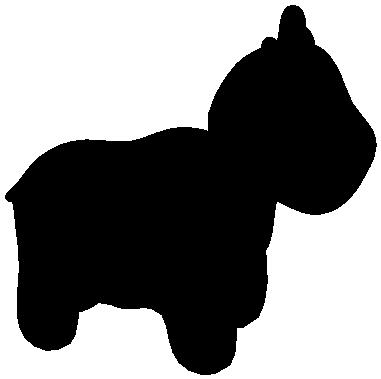}
& 
 \includegraphics[width=\mywidthx, height=\myheightx, keepaspectratio]{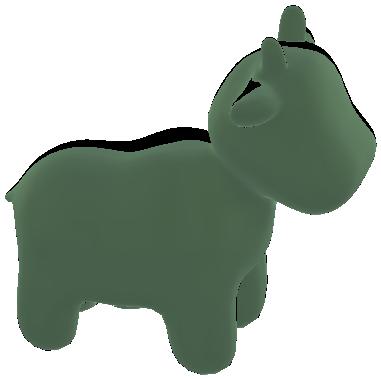}
& 
 \includegraphics[width=\mywidthx, height=\myheightx, keepaspectratio]{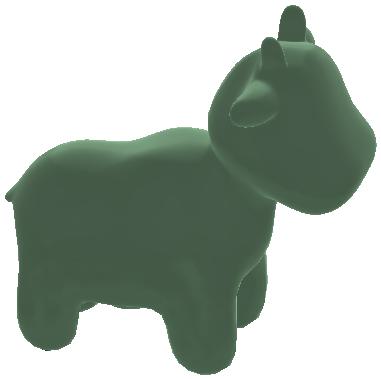}
& 
 \includegraphics[width=\mywidthx, height=\myheightx, keepaspectratio]{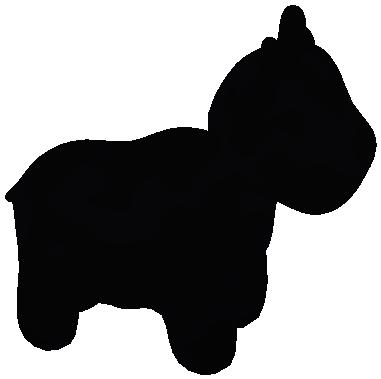}
& 
 \includegraphics[width=\mywidthx, height=\myheightx, keepaspectratio]{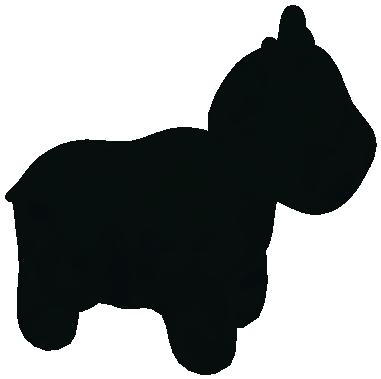}
& 
 \includegraphics[width=\mywidthx, height=\myheightx, keepaspectratio]{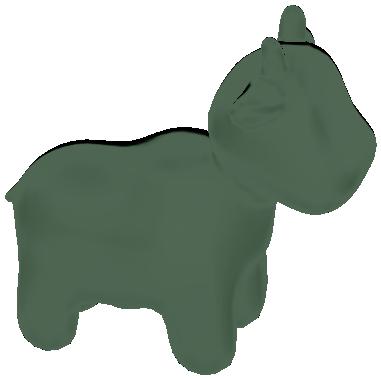}
& 
 \includegraphics[width=\mywidthx, height=\myheightx, keepaspectratio]{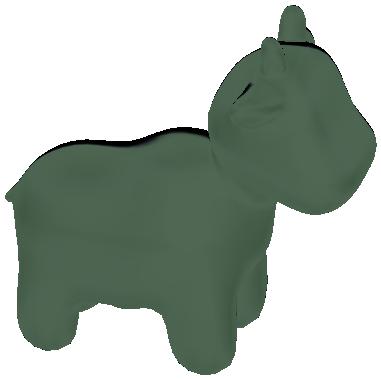}
\\ \hline \\[-0.2cm]
 & \cellcolor{cellParamBased}\rpc		%
 & \cellcolor{cellParamBased}\tsc		%
 & \cellcolor{cellParamBased}\fmbrdfc		%
 & \cellcolor{cellParamBased}\disneyc		%
 & \cellcolor{celPurelyNeural}Add Sep		%
 & \cellcolor{celPurelyNeural}Add Shared		%
 & \cellcolor{celPurelyNeural}Add Sep (enh.)		%
 & \cellcolor{celPurelyNeural}Add Shared (enh.)		%
  \end{tabular}
  \caption{
  Analysis of the spatial variance of the reconstructed BRDFs. Shown are the albedos rendered without the cosine term for five spatially uniform objects of the MERL-based semi-synthetic dataset. We see that overall all models are able to capture the uniformity of the BRDF well and show minimal spatial variation. The results also reveal the ambiguity introduced by an additive splitting strategy, which allows the model to capture the appearance solely by the specular term, leading to almost zero albedo. While this 
  is particularly true for the purely neural models, we also occasionally observe it for other models (\eg green latex for Torrance-Sparrow or Disney for Ipswitch pine).
  }
\label{fig:supp:renderings_diffuse_flat}
\end{figure*}

As described in the main text, the MERL BRDFs are uniform over the respective meshes for the semi-synthetic dataset. To assess how well the models can capture this spatial uniformity, we render the albedo without the cosine term for all models, that employ an additive split. The results for five materials in \cref{fig:supp:renderings_diffuse_flat} reveal very little spatial variation of the albedos, which indicates that all models are able to capture the spatial uniformity quite well.

\subsection{Additional Comparison Results}
\label{sec:supp:additional_experiments_comp}

We report quantitative results for the individual objects of the DiLiGenT-MV dataset in \cref{tab:supp:diligent_quantitative} and qualitative results for both datasets in \cref{fig:supp:renderings_real,fig:supp:renderings_synth_1,fig:supp:renderings_synth_2,fig:supp:renderings_synth_3,fig:supp:renderings_synth_4,fig:supp:renderings_synth_5,fig:supp:renderings_synth_6}.

\paragraph{Real-World Data}
The quantitative evaluation on the individual objects in \cref{tab:supp:diligent_quantitative} confirms that for the DiLiGenT-MV dataset \cite{Li2020DiLiGentMVDataset}, the difference between the approaches based on parametric models and purely neural methods is quite small. We even observe that for individual objects like the bear, some parametric approaches perform slightly better than the purely neural approaches -- in particular, better than approaches with more layers for the directions (\eg Single MLP). This behavior can also be observed qualitatively in \cref{fig:supp:renderings_real}. The reason might be the noise in the real-world data, to which the purely neural methods seem to be slightly more sensitive. The fact that neural approaches with fewer layers for the directions (\eg Additive shared) yield better results supports this claim, in particular in light of the analysis of the number of layers for the directions presented in
\iftoggle{arxiv}{\cref{sec:analysis_brdf_models}}{Sec.~6.2}.

Moreover, \cref{fig:supp:renderings_real} reveals that all methods show errors in similar regions - recesses in particular. We hypothesize un-modelled interreflections as a potential reason. Due to the indicator function in the rendering equation for our scenario
(\iftoggle{arxiv}{\cref{eq:rendering_single_dir_light}}{Eq.~(9)} in the main paper),
shadows in the recesses will be completely black in our renderings. That is, however, a simplification because, in reality, some light reflected off the near surfaces will reach the shaded regions in the recesses. Therefore, larger errors for the image-based metrics in those parts of the mesh are expected.

Finally, \cref{tab:supp:diligent_quantitative} confirms that for the novel additive strategy, which we proposed in
\iftoggle{arxiv}{\cref{sec:enhancingAddSplit}}{Sec.~4.4},
we observe consistent improvements in the extended vanilla additive models.

\paragraph{Semi-Synthetic Data}
\cref{fig:supp:renderings_synth_1,fig:supp:renderings_synth_2,fig:supp:renderings_synth_3,fig:supp:renderings_synth_4,fig:supp:renderings_synth_5,fig:supp:renderings_synth_6} show a systematic advantage of purely neural methods for the challenging materials of the MERL dataset. The error maps reveal that in particular the specular peaks are much better represented, often showing a significant improvement over the parametric methods. While the difference is smaller for more diffuse materials, we still see an advantage of the purely neural methods.

\begin{table*}[t]  %
  \centering  %
  \footnotesize
  
  \setlength\tabcolsep{6pt} %

  \newcolumntype{C}{ >{\centering\arraybackslash} m{0.065\textwidth} } %

\begin{tabular}{l|l||CCCC|CCC|CC}
Datasets & Error metric & \cellcolor{cellParamBased}\rpc		%
 & \cellcolor{cellParamBased}\tsc		%
 & \cellcolor{cellParamBased}\fmbrdf \cite{ichikawa2023fresnel}		%
 & \cellcolor{cellParamBased}\disneyc		%
 & \cellcolor{celPurelyNeural}Single MLP		%
 & \cellcolor{celPurelyNeural}Add Sep		%
 & \cellcolor{celPurelyNeural}Add Shared		%
 & \cellcolor{celPurelyNeural}Add Sep (enh.)		%
 & \cellcolor{celPurelyNeural}Add Shared (enh.)		%
\\ \hline\hline
\multirow{4}{*}{Bear} & \psnrArrow & 43.99 & 44.24 & 44.24 & 44.35 & 44.09 & 43.96 & 44.66 & 44.34 & \textbf{44.74} \\
 & \dssimArrow & 0.476 & 0.464 & 0.464 & 0.454 & 0.489 & 0.477 & 0.445 & 0.463 & \textbf{0.442} \\
 & \lpipsArrow & 0.993 & 0.985 & \textbf{0.973} & 1.022 & 1.124 & 1.115 & 1.030 & 1.062 & 1.022 \\
 & \flipArrow & 2.776 & 2.729 & 2.759 & 2.708 & 2.661 & 2.695 & 2.581 & 2.613 & \textbf{2.550} \\ \hline
\multirow{4}{*}{Buddha} & \psnrArrow & 36.30 & 36.41 & 36.50 & 36.35 & 35.50 & 35.96 & \textbf{36.77} & 36.07 & 36.71 \\
 & \dssimArrow & 1.298 & 1.275 & 1.257 & 1.281 & 1.387 & 1.299 & 1.228 & 1.283 & \textbf{1.224} \\
 & \lpipsArrow & 2.278 & 2.272 & 2.240 & 2.376 & 2.489 & 2.257 & 2.211 & 2.229 & \textbf{2.169} \\
 & \flipArrow & 4.212 & 4.187 & 4.184 & 4.243 & 4.156 & 4.051 & 3.945 & 4.014 & \textbf{3.928} \\ \hline
\multirow{4}{*}{Cow} & \psnrArrow & 46.02 & 46.57 & 43.84 & 46.57 & 46.22 & 46.37 & 47.08 & 46.49 & \textbf{47.12} \\
 & \dssimArrow & 0.395 & 0.373 & 0.422 & 0.372 & 0.391 & 0.381 & 0.358 & 0.374 & \textbf{0.354} \\
 & \lpipsArrow & 1.682 & 1.686 & 1.885 & 1.701 & 1.635 & 1.593 & 1.595 & 1.570 & \textbf{1.568} \\
 & \flipArrow & 2.270 & 2.133 & 3.066 & 2.159 & 1.987 & 1.969 & 1.933 & 1.961 & \textbf{1.919} \\ \hline
\multirow{4}{*}{Pot2} & \psnrArrow & 46.13 & 46.35 & 46.47 & 46.49 & 46.64 & 46.63 & \textbf{46.92} & 46.66 & 46.92 \\
 & \dssimArrow & 0.524 & 0.502 & 0.502 & 0.490 & 0.492 & 0.492 & 0.474 & 0.486 & \textbf{0.472} \\
 & \lpipsArrow & 1.117 & 1.109 & 1.101 & 1.132 & 1.068 & 1.067 & 1.073 & \textbf{1.054} & 1.073 \\
 & \flipArrow & 2.841 & 2.774 & 2.748 & 2.759 & 2.606 & 2.622 & 2.576 & 2.608 & \textbf{2.569} \\ \hline
\multirow{4}{*}{Reading} & \psnrArrow & 35.51 & 35.58 & 36.05 & 35.72 & 35.77 & 35.94 & 36.31 & 35.84 & \textbf{36.38} \\
 & \dssimArrow & 1.184 & 1.173 & 1.114 & 1.155 & 1.185 & 1.139 & 1.066 & 1.116 & \textbf{1.052} \\
 & \lpipsArrow & 2.786 & 2.765 & 2.708 & 2.709 & 2.506 & 2.476 & 2.447 & \textbf{2.403} & 2.466 \\
 & \flipArrow & 3.497 & 3.446 & 3.383 & 3.411 & 3.397 & 3.281 & 3.243 & 3.218 & \textbf{3.172} \\ \hline

\end{tabular}
\caption{
    Quantitative comparison of the BRDF models on all individual objects of the real-world data \cite{Li2020DiLiGentMVDataset}. \psnr, DSSIM and \lpips are computed for the sRGB renderings. All quantities are first averaged over one object and then averaged over all objects in the respective dataset. DSSIM, LPIPS and \FLIP are scaled by 100. We see that for this dataset, the approaches based on parametric models (\mysquare[cellParamBased]) yield results that are comparable to the purely neural approaches (\mysquare[celPurelyNeural]). For individual objects, some parametric approaches even show slightly better results than the purely neural ones, in particular better than approaches with more layers for the directions (\eg Single MLP). Again, a potential reason might be noise in the real-world data to which the purely neural models seem to be a little more sensitive; especially with more layers for the directions. This is supported by the observation that among the purely neural approaches, we see again the tendency that fewer layers for the directions (\eg Additive Separate) is better than more (\eg Single MLP).
    Our enhancement for the additive split (\emph{enh.}) as introduced in 
    \iftoggle{arxiv}{\cref{sec:enhancingAddSplit}}{Sec.~4.4}
    shows consistent improvements of the respective vanilla additive model.
}
\label{tab:supp:diligent_quantitative}
\end{table*}
\input{figures/supp_renderings_real}
\input{figures/supp_renderings_synth_1}
\input{figures/supp_renderings_synth_2}
\input{figures/supp_renderings_synth_3}
\input{figures/supp_renderings_synth_4}
\input{figures/supp_renderings_synth_5}
\input{figures/supp_renderings_synth_6}

\end{document}